\documentclass[12pt,a4paper]{report}
\let\openright=\clearpage

\usepackage[utf8]{inputenc}

\usepackage{amsmath}        
\usepackage{amsfonts}       
\usepackage{amsthm}         
\usepackage{bbding}         
\usepackage{bm}             
\usepackage{graphicx}       
\usepackage{fancyvrb}       
\usepackage{natbib}         
\usepackage[nottoc]{tocbibind} 
\usepackage{dcolumn}        
\usepackage{booktabs}       
\usepackage{paralist}       
\usepackage[usenames]{xcolor}  
\usepackage{pdflscape}		
\usepackage{algorithm}
\usepackage{algpseudocode}
\usepackage[page,toc,titletoc,title]{appendix}
\usepackage{multirow}
\usepackage{makecell}
\usepackage[figuresleft]{rotating}
\usepackage{fixmath} 
\usepackage[font=small]{caption} 
\usepackage{subcaption} 
\usepackage[section]{placeins} 
\usepackage{tocloft}

\def\ThesisTitle{Using reinforcement learning to learn how to play text-based games}

\def\ThesisAuthor{Bc. Mikuláš Zelinka}

\def\YearSubmitted{2017}

\def\Department{Department of Theoretical Computer Science and Mathematical Logic}

\def\DeptType{Department}

\def\Supervisor{Mgr. Rudolf Kadlec, Ph.D.}

\def\SupervisorsDepartment{Department of Theoretical Computer Science and Mathematical Logic}

\def\StudyProgramme{Informatics}
\def\StudyBranch{Artificial Intelligence}

\def\Dedication{%
Firstly, I am very grateful to the Interactive Fiction community for their helpful and kind responses to my questions about IF games.

Secondly, I would like to thank my supervisor, Rudolf Kadlec, for his kindliness and for broadening my horizons by offering inspirational and valuable insights.

Finally, my sincerest gratitude belongs to my friends (especially to the very nitpicky friend for his thorough comments), and to my whole family for being so incredibly supportive, patient and encouraging.
}

\def\Abstract{%
The ability to learn optimal control policies in systems where action space is defined by sentences in natural language would allow many interesting real-world applications such as automatic optimisation of dialogue systems. Text-based games with multiple endings and rewards are a promising platform for this task, since their feedback allows us to employ reinforcement learning techniques to jointly learn text representations and control policies. We present a general text game playing agent, testing its generalisation and transfer learning performance and showing its ability to play multiple games at once. We also present pyfiction, an open-source library for universal access to different text games that could, together with our agent that implements its interface, serve as a baseline for future research.
}

\def\Keywords{%
{reinforcement learning}, {text games}, {neural networks}
}

\usepackage[pdftex,unicode,backref=page]{hyperref}   
\hypersetup{breaklinks=true}
\hypersetup{pdftitle={\ThesisTitle}}
\hypersetup{pdfauthor={\ThesisAuthor}}
\hypersetup{pdfkeywords=\Keywords}
\hypersetup{urlcolor=blue}

\makeatletter
\def\@makechapterhead#1{
  {\parindent \z@ \raggedright \normalfont
   \Huge\bfseries \thechapter. #1
   \par\nobreak
   \vskip 20\p@
}}
\def\@makeschapterhead#1{
  {\parindent \z@ \raggedright \normalfont
   \Huge\bfseries #1
   \par\nobreak
   \vskip 20\p@
}}
\makeatother

\def\chapwithtoc#1{
\chapter*{#1}
\addcontentsline{toc}{chapter}{#1}
}

\theoremstyle{plain}

\theoremstyle{plain}

\theoremstyle{remark}

\DefineVerbatimEnvironment{code}{Verbatim}{fontsize=\small, frame=single}

\newcommand{\cd}[1]{\texttt{#1}}

\renewcommand*{\backref}[1]{}
\renewcommand*{\backrefalt}[4]{{\footnotesize [%
    \ifcase #1 Not cited.%
	\or Cited on page~#2%
	\else Cited on pages #2%
	\fi%
]}}

\let\realref=\ref
\def\ref{\unskip~\realref}
\let\realpageref=\pageref
\def\pageref{\unskip~\realpageref}

\begin{document}


\pagestyle{empty}
\hypersetup{pageanchor=false}
\begin{center}

\centerline{\mbox{\includegraphics[width=166mm]{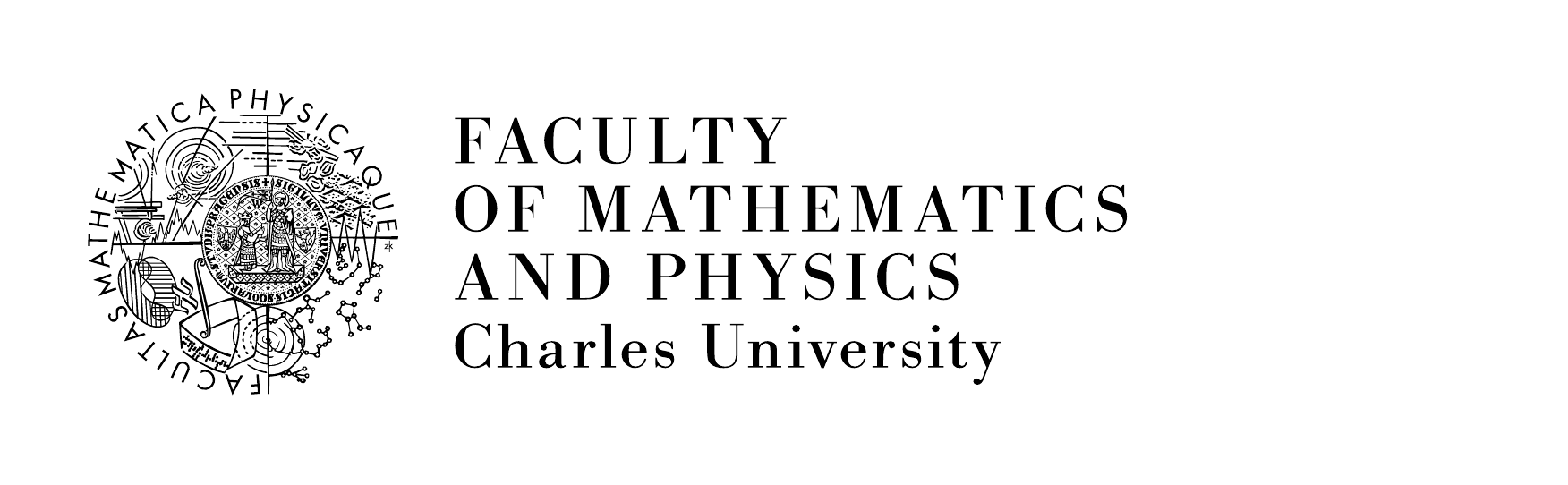}}}

\vspace{-8mm}
\vfill

{\bf\Large MASTER THESIS}

\vfill

{\LARGE\ThesisAuthor}

\vspace{15mm}

{\LARGE\bfseries\ThesisTitle}

\vfill

\Department

\vfill

\begin{tabular}{rl}
Supervisor of the master thesis: & \Supervisor \hphantom{centr} \\
\noalign{\vspace{2mm}}
Study programme: & \StudyProgramme \\
\noalign{\vspace{2mm}}
Study branch: & \StudyBranch \\
\end{tabular}

\vfill

Prague \YearSubmitted

\end{center}

\newpage



\openright
\hypersetup{pageanchor=true}
\pagestyle{plain}
\pagenumbering{roman}
\vglue 0pt plus 1fill

\noindent
I declare that I carried out this master thesis independently, and only with the cited
sources, literature and other professional sources.

\medskip\noindent
I understand that my work relates to the rights and obligations under the Act No.~121/2000 Sb.,
the Copyright Act, as amended, in particular the fact that the Charles
University has the right to conclude a license agreement on the use of this
work as a school work pursuant to Section 60 subsection 1 of the Copyright Act.

\vspace{10mm}

\hbox{\hbox to 0.5\hsize{%
In ........ date ............	
\hss}\hbox to 0.5\hsize{%
signature of the author
\hss}}

\vspace{20mm}
\newpage


\openright

\vbox to 0.5\vsize{
\setlength\parindent{0mm}
\setlength\parskip{5mm}

Title:
\ThesisTitle

Author:
\ThesisAuthor

\DeptType:
\Department

Supervisor:
\Supervisor, \SupervisorsDepartment

Abstract:
\Abstract


\vss}

\newpage


\openright

\noindent
\Dedication

\newpage

\openright
\pagestyle{plain}
\pagenumbering{arabic}
\setcounter{page}{1}

\renewcommand\cftfigfont{\footnotesize}
\renewcommand\cftfigpagefont{\footnotesize}
\renewcommand\cfttabfont{\footnotesize}
\renewcommand\cfttabpagefont{\footnotesize}
\tableofcontents

\chapter*{Introduction}
\addcontentsline{toc}{chapter}{Introduction}

The process of learning to understand and reason in natural language has always been near the centre of attention in Artificial Intelligence (AI) research.
One of the recently explored tasks in language understanding whose successful solving could have a big impact on learning to comprehend and respond in dialogue-like environments (\cite
{DBLP:books/lib/JurafskyM09}) is the task of playing text-based games.

\bigskip
In text games, which are also known as Interactive Fiction (IF, \cite{Montfort:2005}), the player is given a description of the game state in natural language and then chooses one of the actions which are also given by textual descriptions.
The executed action results in a change of the game state, producing new state description and waiting for the player's input again.
This process repeats until the game is over.

More formally speaking, text games are sequential decision making tasks with both state and action spaces given in natural language.
In fact, a text game can be seen as a dialogue between the game and the player and the task is to find an optimal policy (a mapping from states to actions) that would maximise the player's total reward.
Consequently, the task of learning to play text games is very similar to learning how to correctly answer in a dialogue.

Usually, text games have multiple endings and countless paths that the player can take.
It is often clear that some of the paths or endings are better than others and different rewards can be assigned to them.
Availability of these feedback signals makes text games an interesting platform for using reinforcement learning (RL), where one can make use of the feedback provided by the game in order to try and infer the optimal strategy of responding in the given game, and potentially, in a dialogue. 


\bigskip

Recently, there have been successful attempts (\cite{DBLP:journals/corr/HeCHGLDO15}, \cite{DBLP:journals/corr/NarasimhanKB15}) to play IF games using RL agents.
However, the selected games were quite limited in terms of their difficulty and even more importantly, the resulting models had mostly not been tested on games that had not been seen during learning.

While being able to learn to play a text game is undoubtedly a success in itself, we should keep in mind that in order for the resulting model to be useful, it must generalise well to previously unseen data.
In other words, we can merely hypothesise that a successful IF game agent can at least partly understand the underlying game state and potentially transfer the knowledge to other, previously unseen, games, or even natural dialogues.
And for the most part, it remains to be seen how the RL agents presented in \cite{DBLP:journals/corr/HeCHGLDO15} and \cite{DBLP:journals/corr/NarasimhanKB15} perform in terms of generalisation in the domain of IF games.

\bigskip

To summarise, IF games provide a very large and interesting platform for language research, especially when using reinforcement learning.
Some attempts have been made at learning to play them using RL techniques, however, there has not been enough evidence that the resulting models do indeed understand the text and that they can generalise to new games or dialogues, which is what would make the models useful in real-life applications.

\newpage

In this work, we mainly focus on the following:
\begin{itemize}
\item We present \textbf{pyfiction}\footnote{See appendix \ref{app:pyfiction} or \url{https://github.com/MikulasZelinka/pyfiction}.}, an open-source library that enables researchers to universally access different IF games and integrates into OpenAI Gym\footnote{A platform for evaluating RL agents: \url{https://gym.openai.com/}, \cite{DBLP:journals/corr/BrockmanCPSSTZ16}.}.
\item We employ reinforcement learning algorithms to create a general agent capable of learning to play IF games. The agent is a part of the pyfiction library and can consequently be used as a baseline for future research.
\item We explore what properties the resulting models have and how they perform in terms of generalisation.
\end{itemize}

\bigskip
\bigskip
\bigskip

The structure of the thesis is as follows.

\bigskip

In chapter 1, we describe the domain of text-based games and their variants, emphasising the influence of different game attributes on the learning task difficulty.

In chapter 2, we formally define the text-game learning task and review the methods commonly used for similar problems, including reinforcement learning, neural networks and techniques for language representation.

In chapter 3, we introduce our text-game playing agent with a general architecture and compare the agent to the ones presented in \cite{DBLP:journals/corr/HeCHGLDO15} and \cite{DBLP:journals/corr/NarasimhanKB15}.

In chapter 4, we test the agent on single-game and multiple-games learning tasks and conduct experiments focused on its transfer learning and generalisation abilities.

\chapter{Text-based games}
\label{chp:text-games}

In this chapter, we introduce the domain of text-based games, commonly known as \textbf{interactive fiction} (IF). We describe the properties of IF games that influence the difficulty of the learning task.

We also present \textbf{pyfiction}, a library that allows convenient access to different IF games for research purposes.

\section{Game structure}

IF games typically use a basic input-output loop of providing a text description of the game state, waiting for player's action, updating the game state according to the chosen action and responding with a new text description of the new game state.
This then continues until a final state is reached.

\begin{figure}[!ht]\centering
\includegraphics[]{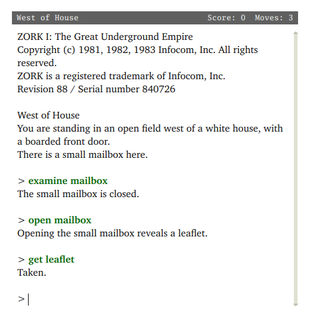}
\caption{\emph{Zork I}, a classic IF game\protect\footnotemark.}
\label{fig:zork}
\end{figure}

\footnotetext{Source: \url{https://en.wikipedia.org/wiki/File:Zork_I_screenshot_video_game_Gargoyle_interpreter_on_Ubuntu_Linux.png}.}

There are exceptions to the basic loop in several games such as visual-based puzzles or logical minigames.
In this work, though, we focus on simpler games without similar elements.

Nevertheless, such games will likely present interesting challenges in~the~future.

Additionally, we simplify the task by ignoring the occasional extra resources such as images, sounds or other complementary game assets that enhance the player experience but are not necessary for the artificial agent's learning process.

\newpage

\section{Genres and variants}
\label{sec:variants}

\begin{figure}[!ht]\centering
\makebox[\textwidth]{\includegraphics[width=\textwidth]{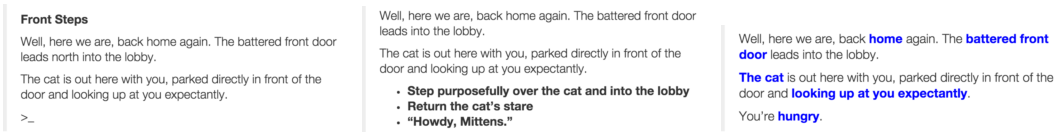}}

\caption{Parser, choice and hypertext-based games (\cite{DBLP:journals/corr/HeCHGLDO15}).}
\label{fig:if-types}
\end{figure}

While the output of the game simulator is almost always a text description, the form of input that the game expects does vary.
One of the most common criteria for classifying IF games is based on their accepted types of text input (for examples, see figure \ref{fig:if-types}):

\begin{itemize}

\item \emph{parser-based}, where the player types in any text input freely,

\item \emph{choice-based}, where multiple actions to choose from are typically available, \emph{in~addition} to the state description,

\item \emph{hypertext-based}, where multiple actions are present as clickable links \emph{inside} the state description.
\end{itemize}

This classification is mentioned mainly because it has significant implications for our task. There is a plethora of other criteria that can be used for classification of different types of IF games.
The largest source of information on IF games and their variants is the Interactive Fiction Database\footnote{IFDB: \url{http://ifdb.tads.org/}\label{ftn:ifdb}.} which features useful filters, tags and lists for finding specific kinds of games.
For our purposes, though, the differences in game genre or storytelling elements do not largely matter.

Text games come in virtually all genres, often with different minigames or puzzles incorporated into them.
As mentioned earlier, we will mostly deal with games without any additional special parts that do not fit into the general and simple input-output loop described above.
\\\\
In this work, we only deal with \emph{choice-based} and \emph{hypertext-based} games but both variants are referred to as \emph{choice-based}, as the hyperlinks are simply considered additional choices.

In fact, note that even all parser-based games with finite number of actions can be converted to choice-based games by simply enumerating all the actions accepted by the interpreter at different time steps.
This trick was used by \cite{DBLP:journals/corr/NarasimhanKB15} where a subset of all possible action combinations was presented as the possible choices to the agent.

\newpage

\section{Rewards}
\label{sec:rewards}

IF games do often have multiple endings which are based on the player's choices. These games can be found on the IFDB\textsuperscript{\ref{ftn:ifdb}} under the \cd{multiple endings} or \cd{cyoa} (choose your own adventure) tags.
\\\\
In some games, explicit numerical rewards are present to reflect the player's performance.
One example of such game is \emph{Six}\footnote{Six: \url{http://ifdb.tads.org/viewgame?id=kbd990q5fp7pythi}.} by Wade Clarke in which the player is rewarded after ``tagging'' one of the six children during the Tag playground game.
\\\\
In other games, the feedback comes in purely textual form, saying how well the player did. The most extreme cases of distinct endings are usually the classic \emph{``You live happily every after.''} and \emph{``You died''}.
There are games with almost any number of endings, but importantly, if we define the ending as the last received state description, there are even games with thousands or more endings.

One example of such game is \emph{Star Court}\footnote{Star Court: \url{http://ifdb.tads.org/viewgame?id=u1v4q16f7gujdb2g}.} by Anna Anthropy, where in a specific ending, the player is sentenced to spend a variable --- randomly generated --- number of years in prison.

Similarly, in \emph{Machine of Death}\footnote{Machine of Death: \url{http://ifdb.tads.org/viewgame?id=u212jed2a7ljg6hl}.} by Hulk Handsome, one ending includes a recapitulation of player's recent choices, resulting in an exponential number of endings in the number of mentioned choices.

\subsubsection{Assigning numerical rewards}

In order to make use of the amount of feedback present in various text games, \cite{DBLP:journals/corr/HeCHGLDO15} manually assigned numerical rewards to various endings of the games \emph{Saving John} and \emph{Machine of Death}, trying to quantify how well the player did to reach each specific ending.

We use the same rewards for these two games that are also used in our experiments. Additionally, we repeat the process of annotating game endings for different games (see appendix \ref{app:text-games} for details).

\subsubsection{Reward distribution}

Note that we have only mentioned annotating different \emph{endings}, and not states in general.

This is because in \emph{Saving John}, \emph{Machine of Death} as well as in other games we used, it corresponds to the nature of the game. 

For all other non-terminating game states, the player is given a small negative rewards in order to encourage the agent to learn to play efficiently.
Altogether, the rewards built into the games as a feedback enable us to use reinforcement learning effectively for this task.
\\\\
In general, though, we hypothesise that games like \emph{Six} with more densely distributed rewards should be easier to learn thanks to their presence throughout the whole game.

\newpage

\section{Game properties and their influence on task difficulty}
\label{sec:game-properties}

IF games obviously differ immensely in terms of both their content and presentation. These differences are hard to capture and it is also unclear how they impact the learning process. Here, we focus on some of the quantifiable properties of IF games that contribute to the overall difficulty of the learning task.

\subsection{Vocabulary size}
\label{sec:vocabulary}

By vocabulary size we mean the number of unique words --- tokens --- used throughout the game.
In natural language processing (NLP) tasks, the tokens are commonly converted to word indices and consequently, a single sentence is represented as a vector of these indices.
Vocabulary size then corresponds to the number of different word indices or to the maximum word index.

Unsurprisingly, larger vocabulary results in a more difficult task. In order to simplify the problem, techniques that reduce the vocabulary size such as \emph{stemming} or \emph{lemmatisation} are commonly used in NLP tasks.

The core idea behind these algorithms is to identify and unify semantically identical or similar words with different syntactic expressions, in particular synonyms.
\\\\
In our case, the individual IF games are relatively small and usually only contain thousands of unique tokens at most. Since the amount of data does not present a problem, we only employ few simple rules to help reduce the vocabulary size (see section \ref{sec:preprocessing} for more details).

\subsection{Action input type}

The types of input accepted by IF games are described in section \ref{sec:variants}.

Their impact on the task difficulty is straightforward; parser-based games require the player to generate their input, resulting in unbounded action space.

On the other hand, choice-based and hypertext-based do not require any text generation and it is only necessary to evaluate the present actions to learn to play the games.

While most parser-based games accept only a finite number of sentences and consequently can be made much easier by enumerating all actions accepted by the game interpreter, their original version presents a great challenge for the future.

A different direction of research might be learning from IF manuals or experienced players, similar to how \cite{DBLP:journals/corr/BranavanSB14} learned to play the strategy game Civilisation using game manuals.

\subsection{Game size}

The sheer number of underlying game states the player can visit and the number of actions they can take has an increasing impact on the task difficulty.

In particular, the branching factor (the average number of actions resulting in different states) is a very important metric.
Consider an agent that is randomly exploring the state space.
Now if the agent has thirty actions to take of which only one leads to a positive reward, it is highly unlikely that the exploration policy will find the reward.
This problem can be further highlighted if these states with a high number of possible actions are themselves located further into the game, meaning even reaching the difficult decision state is unlikely.
\\\\
Moreover, if rewards are only given to the player at the end of each game episode, as is the case in the games we utilise for training the agents, longer paths to the endings with more states result in much higher difficulty due to how sparse the rewards are.

\subsection{Cycles}
\label{sec:cycles}

\begin{figure}[!ht]\centering
\makebox[\textwidth]{\includegraphics[width=\textwidth]{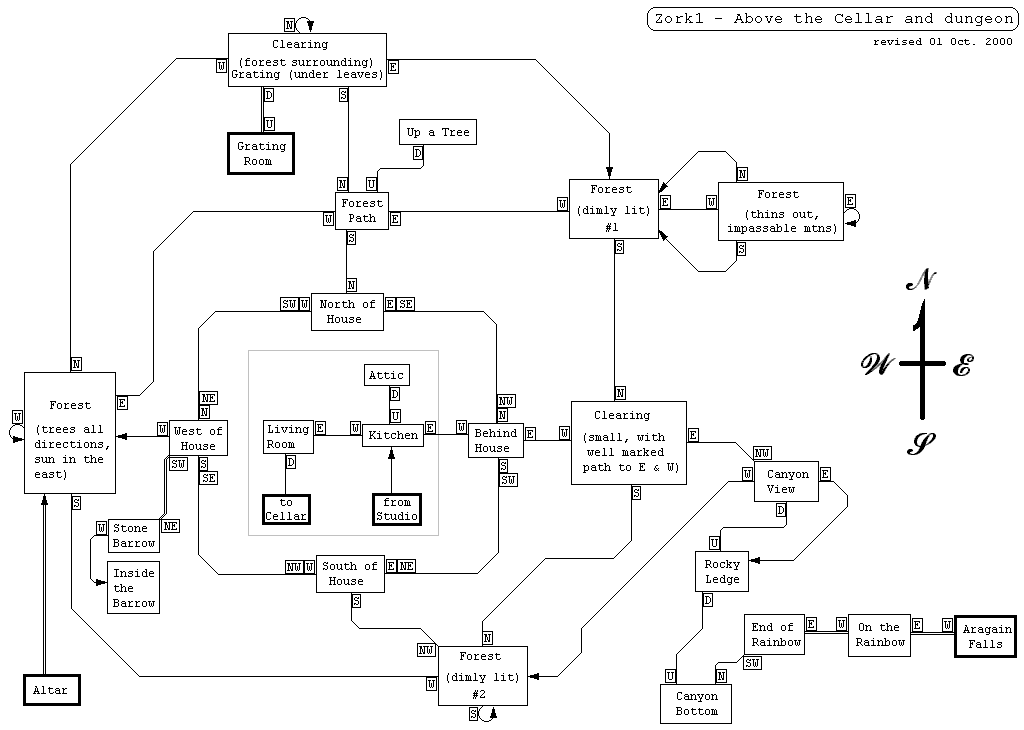}}
\caption{Part of \emph{Zork I}'s game map\protect\footnotemark.}
\label{fig:zork-map}
\end{figure}

If we imagine an IF game as an oriented graph with nodes corresponding to the game states and actions representing the edges between the nodes (see figure~\ref{fig:zork-map}), we can identify some interesting properties of the graph relevant to the task difficulty.

\footnotetext{Source: \url{http://www.lafn.org/webconnect/mentor/zork/zorkText.htm}.}

The most important feature is the presence or absence of cycles in the graph. In other words, if we have an agent with a deterministic policy, is it possible that it will never finish the game?

Interestingly, in most IF games, the answer to this question is yes, ergo there is a number of cycles in which a reflexive agent (an agent without memory) can get stuck.
The most common example of such cycle is a transition between two rooms with a bidirectional connection, where the agent would choose \cd{go north} and \cd{go south} respectively, resulting in an infinite loop.
\\\\
There are other important metrics, such as the average number of actions per state or perhaps the ratio of meaningful actions (actions that do further progress the story).
These features do impact the task difficulty, although not as severely as cycle presence.
While cycles can make it impossible for a simple agent to learn or at least completely explore the state space, the number of actions and the ratio of ``good'' and ``bad'' actions should only impact the speed of the learning process.

\subsection{Completeness and hidden game states}
\label{sec:hidden-states}

In simple games, the text descriptions of states given to the player at any time contain all necessary information and thus provide complete game state to the player.

However, it is fairly common, especially in more complex games, that the game is making use of an underlying engine that produces descriptions that describe the state of the game world only partly.
We call the underlying and complete game state the \emph{hidden state} and the description the player can see is referred to as the \emph{visible state}.
\\\\
There are two common principles that result in the presence of hidden states.

Firstly, the game engine may be keeping track of important statistics such as player's health and their inventory, or of the state and location of various game objects that the player can potentially interact with.
In order to find out what is in the player's inventory, for example, the player has to select an action resulting in inventory description that is completely separate from the world description received in the last step. We say that the game provides \emph{incomplete} state descriptions.

To play these games successfully, an agent with internal memory or at least an agent that takes into account a history of the received game descriptions is required.
\\\\
Secondly, games often make use of random text descriptions for the same underlying state. For example, in \emph{Star Court}, the player is told that they took a specific job, but the job description varies based on few randomly selected options.
For the game engine, however, all the jobs essentially represent the same game state and the point of the random descriptions is to add some flavour to the game and increase its replay value.

Implications of non-deterministic descriptions on task difficulty are discussed in section \ref{sec:random-descriptions}.
\\\\
As we will see in the next chapter, the additional complexity coming from the incompleteness of state descriptions is rather significant, as it basically changes the underlying task from solving simple Markov decision processes (MDP, see section \ref{sec:mdp}) to solving partially observable MDPs, which are typically intractable (\cite{monahan1982state}).

\subsection{Deterministic and non-deterministic properties}
\label{sec:determinism}

There is a varying degree of randomness in IF games. Some, like \emph{Saving John}, have no random elements and are purely deterministic.
These games can be solved by simple graph searching techniques such as breadth-first or depth-first search.
In fact, even an agent that behaves randomly, only remembers its best path and iteratively plays the game will eventually find an optimal solution to such game.

Other games, for example \emph{Machine of Death}, can be random, resulting in a much more difficult task. We identify two possible random elements to the IF games.

\subsubsection{State and action descriptions}
\label{sec:random-descriptions}

In some games, a single room, a single game state or a single action is always described using the exact same words; in others, the description can be generated randomly, resulting in different descriptions for the same inner states expressing the same meaning.
For example, a state could be randomly described as either ``\emph{There is a glass on the table.}'' or ``\emph{On the table, you notice an empty glass.}''
\\\\
The most obvious and a very important implication of such functionality is that the agent cannot easily tell if two states are the same.
As a consequence, these games can no longer be solved by simple graph search algorithms as one cannot determine if two nodes are really the same.

In particular, the agent now only observes visible states, different from the underlying hidden states.
In section \ref{sec:hidden-states}, we mentioned that hidden states, where the complete description of the game world consists of \emph{multiple and separate} descriptions, present a challenge.
However, in this case we only have a \emph{single} hidden state expressed in different ways.

Consequently, in the case of random descriptions, no memorisation is explicitly required.
Even an agent that does not account for these hidden states can theoretically learn to play games with random descriptions, as it can potentially either learn to handle a higher number of varying state descriptions (that represent a single hidden state capturing the whole game world) or even learn --- from context --- that a set of states really represents the same underlying state.

%
%
\subsubsection{State-action transitions}

Randomness can also lie in the way transitions from states using specific actions work. Selecting an action $a$ in state $s$ might either always result in the same state (deterministic behaviour); or in different states depending on the non-deterministic game engine (non-deterministic behaviour).

It appears that most games, or more precisely most state-actions tuples in most games, are deterministic. In other words, even in games where some transitions are random (e.g. \emph{Machine of Death} or \emph{Star Court}), most transitions are deterministic.
\\\\
Non-deterministic behaviour of the modelled environment is implicitly accounted for by the theory of MDPs and does not significantly increase the theoretical learning task difficulty --- or more precisely, the tractability of the problem.
Nonetheless, the games we use for training our agents do differ in terms of the presence of random transitions and this additional complexity is very noticeable as the convergence of the learning algorithms is much slower on the non-deterministic games in practice (see section \ref{sec:individual-games} for the results of learning individual games).

\section{Related problems and human performance}

Even though humans would typically find it easier to play text games than some of the classic visual-based games, the opposite is interestingly true for RL agents.
Most notably, \cite{DBLP:journals/corr/MnihKSGAWR13} and \cite{DBLP:journals/nature/MnihKSRVBGRFOPB15} reached human-like or even superhuman performance on classic, visual-based games from the Atari platform, whereas \cite{DBLP:journals/corr/HeCHGLDO15} showed that humans significantly outperform their agent in both presented text games.

These results are not surprising as there has undoubtedly been less research towards RL for text-game playing agents than towards agents playing visual-based games.

However, we hypothesise that the main factor behind this gap is that the challenge in the majority of Atari games comes mostly from having to react fast, or more precisely, the lack of the ability to plan can be compensated for by very fast reactions in a lot of games, which is obviously very easy for artificial agents.

In text games, on the other hand, reaction speed is not a factor and the main challenge, besides planning in more complex IF games, lies in understanding natural language.
\\\\
Additionally, in \cite{DBLP:journals/corr/HeCHGLDO15}, not only did human players outperform the virtual IF agents significantly but their sample efficiency was also incomparably higher. While the human players managed to learn to play the games in only few tries, the agents took at least thousands of iterations to reach acceptable performance.

\newpage

\section{Unifying access to various text games}

There are different formats, engines and interpreters used to create and play IF games.
To our knowledge, there is currently no universal interface to the various text-game types, making it difficult to try learning to play multiple games or conduct experiments related to generalisation that would require larger sets of games.
\cite{DBLP:journals/corr/NarasimhanKB15} and \cite{DBLP:journals/corr/HeCHGLDO15} also used different games based on different engines for their experiments, making it troublesome to compare the results.

It is important for natural language processing (NLP) tasks, including the task of learning to play IF games, to be able to work with large amounts of data and cover as much of the language space as possible due to the complexity and the unbounded nature of human language.
\\\\
To help address this issue, we present \textbf{pyfiction}\footnote{\label{ftn:pyfiction}See \url{https://github.com/MikulasZelinka/pyfiction}.}, a Python library whose aim is to enable researchers to simplify access to IF games of various types or formats.
Among others, the library currently supports games \emph{Saving John} and \emph{Machine of Death} used in \cite{DBLP:journals/corr/HeCHGLDO15}.
See appendix \ref{app:text-games} for a detailed description of supported games used in this thesis and available for general purposes.
\\\\
The library is easily extensible and also features support of any games written in \textbf{Inform 7}\footnote{A system for designing IF games, see \url{http://inform7.com}.}. and runnable by the \textbf{Glulxe}\footnote{Interpreter of Z-machine game files, see \url{https://github.com/erkyrath/glulxe}.} interpreter by providing a Python interface to the interpreter.

Since most tools for creating IF games also support exporting to the web-based HTML format, pyfiction features a Python-based HTML simulator with several example games.
\\\\
All IF game simulators provided by pyfiction share the same I/O interface, making it very simple to test an agent across a number of games based on different engines.
\\\\
Additionally, in order to provide even more universal interface, we plan to integrate the games supported by pyfiction into OpenAI Gym (\cite{DBLP:journals/corr/BrockmanCPSSTZ16}), a general framework for evaluating RL agents on games of various genres.
\\\\
For more details about pyfiction and its game interface, refer to appendix \ref{app:pyfiction} or to the library website\textsuperscript{\ref{ftn:pyfiction}}.

\chapter{Background}

In this chapter, we define the task of playing text-based games and briefly review the methods and tools we utilise for building our agent.

\section{Core definitions}

Now that we have a clear idea about what properties text-based games have (see~chapter \ref{chp:text-games} for their detailed description), we can define the concept and subsequently the task of solving the games more formally.

\subsection{Text games}

\newtheoremstyle{break}
  {9pt}
  {9pt}
  {\itshape}
  {}
  {\bfseries}
  {.}
  {\newline}
  {}

\theoremstyle{break}
\newtheorem{definition}{Definition}[section]

Text game is a sequential decision-making task with both input and output spaces given in natural language.

\begin{definition}[text game]
\label{def:text-game}
Let us define a text game as a tuple $G = \langle H, H_t, S, A, \mathcal{D}, \mathcal{T}, \mathcal{R} \rangle$, where
\begin{itemize}
\item $H$ is a set of game states, $H_t$ is a set of terminating game states, $H_t \subseteq H$,
\item $S$ is a set of possible state descriptions,
\item $A$ is a set of possible action descriptions,
\item $\mathcal{D}$ is a function generating text descriptions, $\mathcal{D}: H \to (S \times 2^A)$,
\item $\mathcal{T}$ is a transition function, $\mathcal{T}: (H \times A) \to H$,
\item $\mathcal{R}$ is a reward function, $\mathcal{R}: S \to \mathbb{R}$.
\end{itemize}
\end{definition}

Generally speaking, both the transition function $\mathcal{T}$ and the description function $\mathcal{D}$ may be stochastic and as mentioned in section \ref{sec:game-properties}, the properties of the game and its functions have a great impact on the task difficulty.

In particular, if both $\mathcal{D}$ and $\mathcal{T}$ are deterministic, the \emph{whole game} is deterministic.
Consequently, as we discussed in section \ref{sec:determinism}, the problem is then reduced to a simple graph search problem that can be solved by graph search techniques and there is no need --- from the perspective of finding optimal rewards --- to employ reinforcement learning methods.

However, from the generalisation perspective, it still might be useful to attempt to learn simple games using RL techniques as the agents could potentially be able to generalise or transfer their knowledge to other, more difficult problems --- which certainly does not apply to the graph search algorithms.

\newpage

\subsection{Markov decision process}
\label{sec:mdp}

We formalise the problem of playing text games as the task of solving a \emph{Markov decision process} (MDP), where the parameters of the MDP correspond to the parameters of a text game as defined in \ref{def:text-game}.

\begin{definition}[Markov decision process]
MDP is a stochastic process defined as a tuple $\mathcal{M} = \langle S, A, \mathcal{P}, \mathcal{R}, \gamma \rangle$, where

\begin{itemize}
\item $S$ is a set of states,
\item $A$ is a set of actions,
\item $\mathcal{P}(s_{t+1} | s_{t}, a_{t})$ is the probability that choosing an action $a_{t}$ in state $s_{t}$ will result in state $s_{t+1}$,
\item $\mathcal{R} (s_t, a_t, s_{t+1})$ is the immediate reward function that assigns a~numerical reward to a~state transition,
\item $\gamma \in [0,1]$ is the discount factor determining preference of immediate or future rewards.
\end{itemize}
\end{definition}

The transition function $\mathcal{P}$ has the \emph{Markov property}, meaning that the result of the state update only depends upon the present state and not upon the state history.

In other words, $\mathcal{P}$ is a function of the last state only, as also seen in the definition.

\begin{definition}[Markov property]
Function $\mathcal{P}: (S \times A) \to S$ is said to possess the \emph{Markov property} if:

$\mathcal{P}(s_{t+1} | s_t, a_t, s_{t-1}, \ldots, s_0) = \mathcal{P}(s_{t+1} | s_t, a_t)$.

\end{definition}

In our case, a text-game MDP is simply an MDP where the states of the MDP correspond to the game state descriptions and the same applies for actions and rewards.

The probability function $\mathcal{P}(s' | s, a)$ is then realised by the game transition function $\mathcal{T}$ and crucially, it is unknown to the agent.

The terminal states of a text game can be implicitly encoded in its MDP as states with a deterministic transition to themselves and a zero reward.
\\\\
The $\gamma$ parameter is commonly called a \emph{discount factor} and it determines the weights of rewards at different time steps.
Lower values lead to preference of immediate rewards whereas higher values result in taking the future into account more.
This parameter plays an important role in determining the optimal behaviour of the agent through influencing the definition of the cumulative reward as seen in equation \ref{eq:cumulative-reward}.
\\\\
Notice that not all IF games necessarily have the Markov property, i.e. in text games, there can be long-term dependencies.
In fact, it is even difficult to determine whether a text game is or is not Markovian.

However, even problems with non-Markovian characteristics are commonly represented and modelled as MDPs while still giving good results (\cite{DBLP:books/lib/SuttonB98}).

\subsection{Learning task}
\label{sec:task}

We define the task of learning to play a text game as the process of finding an optimal policy for the respective text game MDP.

Policy is a function, usually realised by a set of decision-making rules, that takes a state as an input and produces an action as a result. We denote that applying a policy $\pi$ in state $s$ results in action $a$ by writing:

\begin{equation}
a \gets \pi(s).
\end{equation}

The \textbf{discounted cumulative reward} that the agent receives is defined as

\begin{equation}
\label{eq:cumulative-reward}
\sum^{\infty}_{t=0} {\gamma^t \mathcal{R} (s_t, a_t, s_{t+1})}
\end{equation}

and thus the \textbf{optimal policy $\pi^*$}  maximises the following term:

\begin{equation}
\label{eq:optimal-policy}
\sum^{\infty}_{t=0} {\gamma^t \mathcal{R} (s_t, a_t, s_{t+1})}\text{, where } a_t = \pi^{*}(s_t).
\end{equation}

Note that in order to avoid infinite loops, we limit the maximum number of steps the agent can take. The game episodes are consequently always finite and the rewards do not need to be discounted.

If the maximum number of steps is $n$, the agent's reward is then equal to~$\sum^{n}_{t=0} \mathcal{R} (s_t, a_t, s_{t+1})$.
\\\\
For finding the optimal policy in the text-game MDP, we employ reinforcement learning, whose solution techniques, as well as our approach to this specific instance of the problem, are described next.

\newpage

\section{Reinforcement learning}

Reinforcement learning (RL) is an area of machine learning methods and problems based on the notion of learning from numerical rewards obtained by an agent through interacting with an environment \cite{DBLP:books/lib/SuttonB98}.

The RL agent does not have any supervisor and does not know the dynamics of its environment. Consequently, the agent has to both explore its environment and exploit its knowledge thereof.
\\\\
In any given step, the agent observes a \emph{state} of the environment and receives a~\emph{reward signal}.
Based on the current state and the agent's behaviour function --- the \emph{policy} --- the~agent chooses an~\emph{action} to take.

\begin{figure}[!ht]\centering
\includegraphics[]{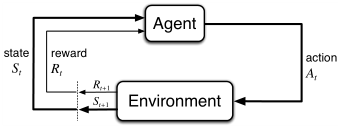}
\caption{Interaction between the agent and the environment (\cite{DBLP:books/lib/SuttonB98}).}
\label{fig:rl-loop}
\end{figure}

The action is then sent to the environment which is updated and the loop repeats. See figure \ref{fig:rl-loop} for illustration of the agent-environment interface.

Reinforcement learning is commonly employed for solving Markov decision process problems in which the dynamics of the environment are unknown to the agent.

The core idea behind solving an MDP problem using RL is iterative learning through combined exploration (finding out which states and actions lead to which rewards) and exploitation (updating the agent's policy as to improve its expected long-time reward).

We follow the notation established in section \ref{sec:task} and denote the agent's learned policy as $\pi(s)$.


\subsection{Updating the policy}
\label{sec:policy-updating}

There are different approaches to learning the optimal policy. Some are based on learning the dynamics of the environment (model-based methods) while others solve the problem without explicitly modelling its transition function (model-free).

Here, we focus on model-free methods that have been shown to perform well on a variety of game-related tasks (\cite{DBLP:journals/cacm/Tesauro95}, \cite{DBLP:journals/nature/MnihKSRVBGRFOPB15}).
\\\\
In order to be able to update the policy, it is useful to work with the concept of \emph{value functions}.
A value function evaluates how good a given state under a given policy is, that is, what reward we can expect to obtain in the long run if we follow the policy from the specific state.
\\\\
The \textbf{state-value function} $v_\pi(s)$ for policy $\pi$ can be formally defined as (\cite{DBLP:books/lib/SuttonB98})

\begin{equation}
\label{eq:state-value-function}
v_\pi(s) = \mathbb{E}_{\pi} \left[ \sum_{t=0}^{\infty} \gamma^t r_{t+1}\ \middle|\ s \right],
\end{equation}

where $r_t$ corresponds to~$\mathcal{R} (s_t, a_t, s_{t+1})$.

Similarly, an \textbf{action-value function} that determines the value of taking action $a$ in state $s$ using policy $\pi$ is defined as

\begin{equation}
\label{eq:action-value-function}
Q^\pi(s, a) = \mathbb{E}_{\pi} \left[ \sum_{t=0}^{\infty} \gamma^t r_{t+1}\ \middle|\ s, a \right].
\end{equation}

Now the optimal policy, denoted $\pi^*$, can be characterised by the optimal state-value function or the action-value function. The latter is defined as

\begin{equation}
\label{eq:optimal-action-value}
Q^*(s, a) = \underset{\pi}{\max}\ Q^\pi(s,a).
\end{equation}

In other words, if we know the optimal action-value function $Q^*$, we can obtain the optimal policy $\pi^*$ by simply choosing the actions with maximum Q-values:

\begin{equation}
\label{eq:optimal-q-policy}
\pi^{*}(s) = \underset{a}{\max}\ Q^{*}(s,a).
\end{equation}

Using this approach, we have now formally reduced the task of learning to play the text-based games to estimating an optimal action-value function in a text-game MDP.
The Q-learning algorithm for solving the problem of optimal action-value function estimation is described next.

\subsection{Q-learning}
\label{sec:q-learning}

Q-learning (\cite{DBLP:journals/ml/WatkinsD92}) is an off-policy method for model-free control based on estimating the action-value function --- the Q-function.

The~Q-learning algorithm was shown to work well across a variety of different game-playing tasks (\cite{DBLP:journals/nature/MnihKSRVBGRFOPB15}, \cite{DBLP:journals/corr/NarasimhanKB15}, \cite{DBLP:journals/corr/HeCHGLDO15}).
\\\\
Q-learning is an off-policy algorithm, meaning the agent can follow an exploration policy while improving its estimates of the optimal policy by updating the Q-values.

Typically, the exploration policy is $\epsilon$-greedy with relation to the Q-function, where $0 \leq \epsilon \leq 1$ is a parameter that corresponds to the probability of choosing a random action.
\\\\

Q-learning attempts to find the optimal Q-values which obey the Bellman equation (\cite{bellman2013dynamic}):

\begin{equation} \label{eq:bellman}
Q^*(s_{t},a_{t}) = \mathbb{E}_{s_{t+1}} \left[ r_{t} + \gamma \cdot \max_{a_{t+1}}Q^{*}(s_{t+1}, a_{t+1})\ \middle|\ s_t, a_t \right].
\end{equation}

Consequently, the update rule for improving the estimate of the Q-function is as follows (\cite{DBLP:books/lib/SuttonB98}):

\begin{equation} \label{eq:q}
Q(s_{t},a_{t}) \gets Q(s_{t},a_{t}) + \\
\alpha_{t} \cdot \left( r_{t+1} + \gamma \cdot \max_{a_{t+1}}Q(s_{t+1}, a_{t+1}) - Q(s_{t},a_{t}) \right),
\end{equation}

where $0 \leq \alpha_t \leq 1$ is the learning rate.
\\\\
The Q-learning algorithm is guaranteed to converge towards the optimal solution (\cite{DBLP:books/lib/SuttonB98}).
\\\\
In simpler problems and by default, it is assumed that the Q-values for all state-action pairs are stored in a table.

This approach is, however, not feasible for more complex problems such as the Atari platform games task or our text-game task, where the state and action spaces are simply too large to store.
In text games, for example, the spaces are infinite.
\\\\
We deal with this problem by approximating the optimal Q-function by a function approximator in the form of a neural network. The Q-function is parametrised as

\begin{equation}
\label{eq:nn-q}
Q^*(s_t, a_t) \approx Q(s_t, a_t, \theta_t) = \theta_t(s_t, a_t),
\end{equation}

where the $\theta$ function is realised by a neural network (see section \ref{sec:nn} for an introduction and section \ref{sec:slsn} for architecture description).
\\\\
The advantage of this non-tabular approach is that even in infinite spaces, the neural network can generalise to previously unobserved inputs and consequently cover the whole search space with reasonable accuracy.

In contrast to linear function approximators, though, non-linear approximators such as neural networks do not guarantee convergence in this context.

\newpage

\section{Neural networks}
\label{sec:nn}

Artificial neural network (ANN or NN) is a computing system realised by interconnected groups of nodes inspired by biological neurons.
\\\\
The original units, inspired by properties of biological neurons and capable of binary classification of linearly separable data, were called perceptrons (\cite{Rosenblatt58theperceptron:}).

A perceptron has a weight vector $\mathbold{w}$, a scalar bias value $b$ and realises a simple binary function on its input vector $\mathbold{x}$:

\begin{equation}
f(\mathbold{x}) = \begin{cases}1 & \text{if }\mathbold{w} \cdot \mathbold{x} + b > 0\\0 & \text{otherwise}\end{cases},
\end{equation}

where $\mathbold{w} \cdot \mathbold{x}$ is the dot product $\sum_{i} \mathbold{w}_i \mathbold{x}_i$.
\\\\

First simple ANNs, called multilayer perceptrons (MLP), then consisted of organised layers of individual perceptron units connected together in a feed-forward fashion, forming a directed acyclic graph.

MLPs have one input and one output layer and an arbitrary number of hidden layers in between.
Each layer consists of a set of perceptrons that additionally apply a non-linear \emph{activation function} $\phi$ to their original output value:

\begin{equation}
f(\mathbold{x}) = \phi (\mathbold{w} \cdot \mathbold{x} + b).
\end{equation}

The presence of multiple neurons in a single layer coupled with the nonlinear activations means that, in contrast to perceptrons, MLP classifiers --- even those with only one hidden layer --- are theoretically capable of classifying data that is not linearly separable.

In fact, a neural network with one hidden layer is a universal function approximator (\cite{DBLP:journals/nn/Hornik91}).
\\\\
MLPs are now commonly referred to as \emph{feedforward neural networks} and the simple perceptron layers are called \emph{fully-connected} or \emph{dense} layers.
\\\\
Neural networks are mostly employed in supervised learning classification problems across a variety of domains.
In this context, the network is given training data $(\mathbold{x}_0, \mathbold{y}_0), \ldots, (\mathbold{x}_n, \mathbold{y}_n)$, where $\mathbold{x}_i$ are the \emph{input vectors} and $\mathbold{y}_i$ are the \emph{target vectors} and the task is to adapt the weights of the network so that $\theta(\mathbold{x}_i) = \mathbold{y}_i$, where $\theta(\mathbold{x})$ is the network function.

\subsection{Gradient descent}

The task of learning in neural networks can be formalised as minimising a defined \textbf{loss function}.

Loss function of an input vector $\mathbold{x}$ and a desired target vector $\mathbold{y}$ is defined to quantify the error produced by the network.
\\\\
One example of a loss function, denoted $\mathcal{L}$, is the mean squared error (MSE) function defined for a neural network $\theta$ and $\mathbold{x}$, $\mathbold{y}$ of length $n$ as

\begin{equation}
\label{eq:mse}
\mathrm{MSE}(\mathbold{x}, \mathbold{y}, \theta) = \frac{1}{n}\sum_{i=1}^n(\theta(\mathbold{x})_i - \mathbold{y}_i)^2.
\end{equation}

The optimisation problem then lies in minimising the given loss function.
\\\\
There are different approaches to the optimisation problem in neural networks and here, we focus on the \textbf{backpropagation} algorithm (\cite{werbos1974beyond}, \cite{hecht1988theory}) for gradient computation.
\\\\
Backpropagation is based on first computing the error in the output layer and then propagating it backwards through the network.

More specifically, we compute the partial derivatives of the loss function with respect to the network weights and update the individual weights between neurons $i$ and $j$ in the direction of the gradient $\frac{\partial \mathcal{L}}{\partial \mathbold{w}_{ij}}$.

The error is then propagated to all weights in the previous layer and this is repeated for all layers up to the first hidden layer.
\\\\
\indent Using the backpropagated errors, the learning is commonly done using stochastic gradient descent (SGD):

\begin{equation}
\mathbold{w} \gets \mathbold{w} - \alpha \nabla \mathcal{L}(\mathbold{w}),
\end{equation}

where $\alpha$ is the learning rate and $\mathcal{L}$ is the loss function.
\\\\
There are alternative optimisers for gradient descent in neural networks such as Adam (\cite{DBLP:journals/corr/KingmaB14}) or RMSProp (\cite{rmsprop}) based on the idea of adaptively changing the learning rate according to the gradient fluctuations that often give better results in practice.

Additionally, the weight updates are commonly done in batches, meaning a number of data samples is propagated through the network at once instead of separately for each sample, resulting in a faster training process and better convergence properties (\cite{Goodfellow-et-al-2016}).






\subsection{Recurrent networks}

\begin{figure}[ht!]\centering
\includegraphics[scale=0.5]{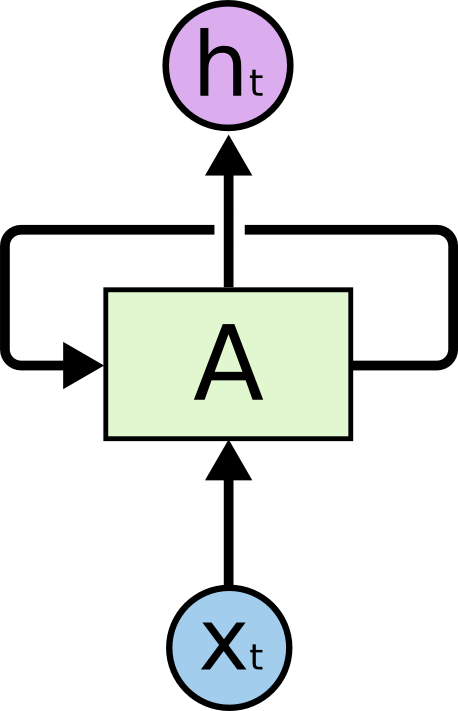}
\captionsetup{justification=centering}
\caption[A simple RNN (\cite{olah2015understanding}).]{A simple RNN (\cite{olah2015understanding}).\\The cell~\textbf{A} receives its last state as an additional input.}
\label{fig:rnn}
\end{figure}

Recurrent neural networks (RNNs) represent a slightly different paradigm of computation, in which there are additional --- recurrent --- connections in the neural cells.
\\\\
In this paradigm, instead of being presented at once, the input is supplied gradually in successive \emph{timesteps}. And in addition to the ``normal'' input at a given timestep, a simple recurrent cell also receives its own last state as an input (see figure \ref{fig:rnn} for illustration).

Consequently, recurrent units, layers and networks provide additional computational power as they are able to maintain information about the past and, theoretically, capture long-term dependencies.

For learning in RNNs, a variant of backpropagation called backpropagation through time (BPTT) is used.
BPTT is based on unfolding the network in time and gradually propagating the error to earlier timesteps (\cite{werbos1990backpropagation}).
\\\\
In practice, however, RNNs have great difficulty remembering the relevant information for long enough and tend to fail to learn important dependencies if there are too many timesteps between them (\cite{hochreiter1991untersuchungen}, \cite{DBLP:journals/tnn/BengioSF94}).

\subsubsection{Long short-term memory}
\label{sec:lstm}

To address the issue of long-term dependencies, \cite{DBLP:journals/neco/HochreiterS97} introduced a special and a more complex variant of RNN cells, the long short-term memory (LSTM).
\\\\
The basic idea of LSTMs is to work with multiple information pathways inside the cell.

On the main pathway, the information from the previous state flows unchanged and additional information can be added to the previous cell state though different gates realised by simple neural network layers with the sigmoid activation function.

For a thorough explanation, see \cite{olah2015understanding}.
\\\\
In practice, LSTM-based networks have displayed incredible performance, obtaining state-of-the-art results in a number of disciplines including language modelling (\cite{DBLP:conf/interspeech/SundermeyerSN12}) or speech recognition (\cite{DBLP:journals/corr/abs-1303-5778}).
\\\\
Thanks to the emphasis that is put on preserving long-term relations in LSTMs, they are suited particularly well for NLP-related tasks, where long-term dependencies in the input data often form important and useful features.

\newpage

\section{Language representation}

One of the critical aspects of NLP tasks is learning a good representation of the input data through creating meaningful features that capture the word semantics.
\\\\
In this section, we first briefly talk about text preprocessing as a tool to reduce the vocabulary size and consequently make the task somewhat easier.

Second, we describe the process of converting the input textual data into a powerful numerical vector representation.

\subsection{Text preprocessing}
\label{sec:preprocessing}

As we mentioned in section \ref{sec:vocabulary} and as should be clear from the description~of word embeddings (see section \ref{sec:embeddings}), the difficulty of the representation task as well as the text-game task grows with the increasing number of unique words.
\\\\
In our task, we currently only deal with vocabulary sizes of thousands of words, which is why we employ only few basic text preprocessing rules based on the syntactic content and not on semantics.

In other words, we are not interested in matching synonyms or more generally, sets of words that would increase the vocabulary size linearly.
Instead, we primarily deal with strings whose number can potentially explode exponentially., i.e. with tokens that do bear a meaning by themselves but are usually not separated from other tokens by whitespace characters.
\\\\
For preprocessing, we follow this procedure:

\begin{enumerate}
\item convert the text to lower case,
\item remove all special characters (only preserve alphanumerical and whitespace characters, quotes and hyphens),
\item split numbers into digits,
\item insert a space between \emph{X's} expressions, i.e. \cd{X's} becomes \cd{X 's},
\item expand unambiguous contracted expressions such as 'll or 've.
\end{enumerate}

Finally, split the text data by white spaces and assign a unique \textbf{token} to each non-empty unique string.

\subsection{Embeddings}
\label{sec:embeddings}

In order to be able to work with high-dimensional complex data such as text in neural networks, we have to first convert the text data into numerical vectors.

Generally, this process is called embedding and in our case, \textbf{word embeddings} are the real-valued vectors corresponding to the actual words in the input.
\\\\
NLP techniques enable us to both map complex textual data into vectors of numbers and, importantly, to also find a relevant mapping function that results in a powerful useful representation.

The process of creating word embeddings from text is described next.
\\\\
The first step is to identify possible unique tokens (see section \ref{sec:preprocessing} for an example in our context).
This is usually done by scanning for all possible words in a dataset, however, it is possible to theoretically add new tokens on the fly at the cost of computational expenses.
In text games, for example, it is convenient to parse the source code or sample the game randomly for thousands of episodes to obtain a good estimate of its complete vocabulary.

After obtaining the vocabulary of tokens, they are ordered and an index is assigned to each token, meaning each sentence is now represented as a vector of token indices.
\\\\
Finally, each token has a set of parameters to it; this is the actual word embedding.
Word embeddings are high-dimensional vectors (the dimension is typically 50 or 100) that --- we hope and hypothesise --- represent semantic features of their corresponding words.
\\\\
There are different approaches to learning good and useful word embeddings.

Recently, RNNs and RNNs with LSTM units have been successfully applied to a number of related tasks in representation generation (\cite{DBLP:word2vec}, \cite{DBLP:journals/taslp/PalangiDSGHCSW16}).

Specifically, we follow the approach used in \cite{DBLP:journals/corr/NarasimhanKB15}, where the representation generator module was trained as a part of a neural network for estimating a Q-value, learning the representation and the control policy jointly.

\section{Summary}
\label{sec:summary-model}

We briefly reviewed the basic methods of reinforcement learning, neural networks and language representation that we utilise for building our agent. Below, we summarise their usage in the context of the text game learning task.
\\\\
We utilise reinforcement learning to define the agent's objective and consequently learn the control policy.
We use the model-free Q-learning control algorithm that estimates the agent's action-value function (Q-function).

The Q-function is approximated using supervised learning in the form of a neural network.
The network makes use of embedding and LSTM layers for representation generation and of dense layers followed by an interaction function between states and actions to finally estimate the Q-value.

The learning targets that determine the value of the network's loss function are then based on the basic Q-learning update rule (see equation \ref{eq:q}).

The network is trained in an end-to-end manner, implicitly building important feature representations of the input text in each layer.

The resulting architecture and its parameters are described in more detail in chapter \ref{chp:agent}.
\chapter{Agent architecture}
\label{chp:agent}

We have now introduced the domain of IF games and the methods commonly used for solving similar sequential decision-making tasks.
\\\\
In this chapter, we present the architecture of our agent, denoted SSAQN, capable of playing choice-based text-games using reinforcement learning.

After presenting recent relevant models and explaining our motivation behind choosing a slightly different architecture, the structure of the chapter closely follows agent's data flow --- we start with complete text descriptions of states and actions and gradually work our way towards a condensed vector representation used for determining the compatibility of the original state-action pairs.

\section{Related models}

We briefly describe two architectures that were recently used to play text games.

\subsection{LSTM-DQN}

\cite{DBLP:journals/corr/NarasimhanKB15} presented an LSTM-DQN agent that used an LSTM network for representation generation, followed by a variant of a Deep Q-Network (DQN, \cite{DBLP:journals/nature/MnihKSRVBGRFOPB15}) used for scoring the generated state and action vectors.

The underlying task of playing parser-based games was effectively reduced to playing choice-based games by presenting a subset of all possible verb-object tuples as available actions to the agent.

In the framework, a single action consists of a verb and an object (e.g. \emph{eat apple}) and the model computes Q-values for objects --- $Q(s, o)$ --- and actions --- $Q(s, a)$ --- separately.

The final Q-value is obtained by averaging these two values.

\subsection{DRRN}
\cite{DBLP:journals/corr/HeCHGLDO15} introduced a Deep Reinforcement Relevance Network (DRRN) for playing hypertext-based games.
The agent learned to play \emph{Saving John} and \emph{Machine of Death} and used a simple bag-of-words (BOW) representation of the input texts.

The learning algorithm is also a variant of DQN; DRRN refers to the network that estimates the Q-value, using separate embeddings for states and actions. DRRN then uses a variable number of hidden layers and makes use of softmax action-selection.

The final Q-value is obtained by computing an inner product of the inner representation vectors of states and actions.

\newpage

\section{Motivation}

Our goal is to introduce a minimal architecture serving as a proof of concept, with the ability to capture important sentence-level features and ideally capable of reasonable generalisation to previously unseen data.

First, we highlight some of the aspects of the LSTM-DQN and DRRN models that could be improved upon in terms of these requirements.
\\\\
The main drawback of DRRN is its use of BOW for representation generation.
Consequently, the model is incapable of properly handling state aliasing and differentiating simple, yet important, nuances in the input, such as in \emph{``There is a treasure chest to your left and a dragon to your right.''} and \emph{``There is a treasure chest to your right and a dragon to your left.''}.

Moreover, \cite{DBLP:journals/corr/HeCHGLDO15} claim that separate embeddings for state and action spaces lead to faster convergence and to a better solution. However, since both state and action spaces really contain the same data --- at least in most games and especially in hypertext games where actions are a subset of states --- we aim to employ a joint embedding representation of states and actions.

We also believe that a joint representation of states and actions should eventually lead to stronger generalisation capabilities of the model, since such model should be able to transfer knowledge between state and action descriptions as their representation would be shared.
\\\\
The LSTM-DQN agent, on the other hand, utilises an LSTM network that can theoretically capture more complex sentence properties such as the word order.
However, its architecture only accepts actions consisting of two words.

Additionally, the two action Q-values are finally averaged, which would arguably be problematic if the verbs and objects were not independent.
For example, the value of the verb ``\emph{drink}'' varies highly based on the object; consider the difference between the values of ``\emph{drink}'' when followed by either ``\emph{water}'' or ``\emph{poison}'' objects.
\\\\
We thus aim to utilise a minimalistic architecture that should:
\begin{itemize}
\item be able to capture dependencies on sentence level such as word order,
\item accept text input of any length for both states and action descriptions,
\item accept and evaluate any number of actions,
\item use a powerful interaction function between states and actions.
\end{itemize}

Next, we present our chosen model.

\newpage

\section{SSAQN}
\label{sec:slsn}

Our neural network model is inspired by both LSTM-DQN and DRRN. For the sake of clarity, it is referred to as \textbf{SSAQN} (Siamese State-Action Q-Network).

Similarly to \cite{DBLP:journals/corr/NarasimhanKB15} and \cite{DBLP:journals/corr/HeCHGLDO15}, we employ a variant of DQN (\cite{DBLP:journals/nature/MnihKSRVBGRFOPB15}) with experience replay and prioritised sampling which uses a neural network (SSAQN) for estimating the DQN's Q-function.
\\\\

SSAQN uses a siamese network architecture (\cite{DBLP:conf/nips/BromleyGLSS93}), where two branches --- a state branch and an action branch --- share most of the layers that effectively generate useful representation features.

\begin{figure}[htb!]
\includegraphics[width=\textwidth]{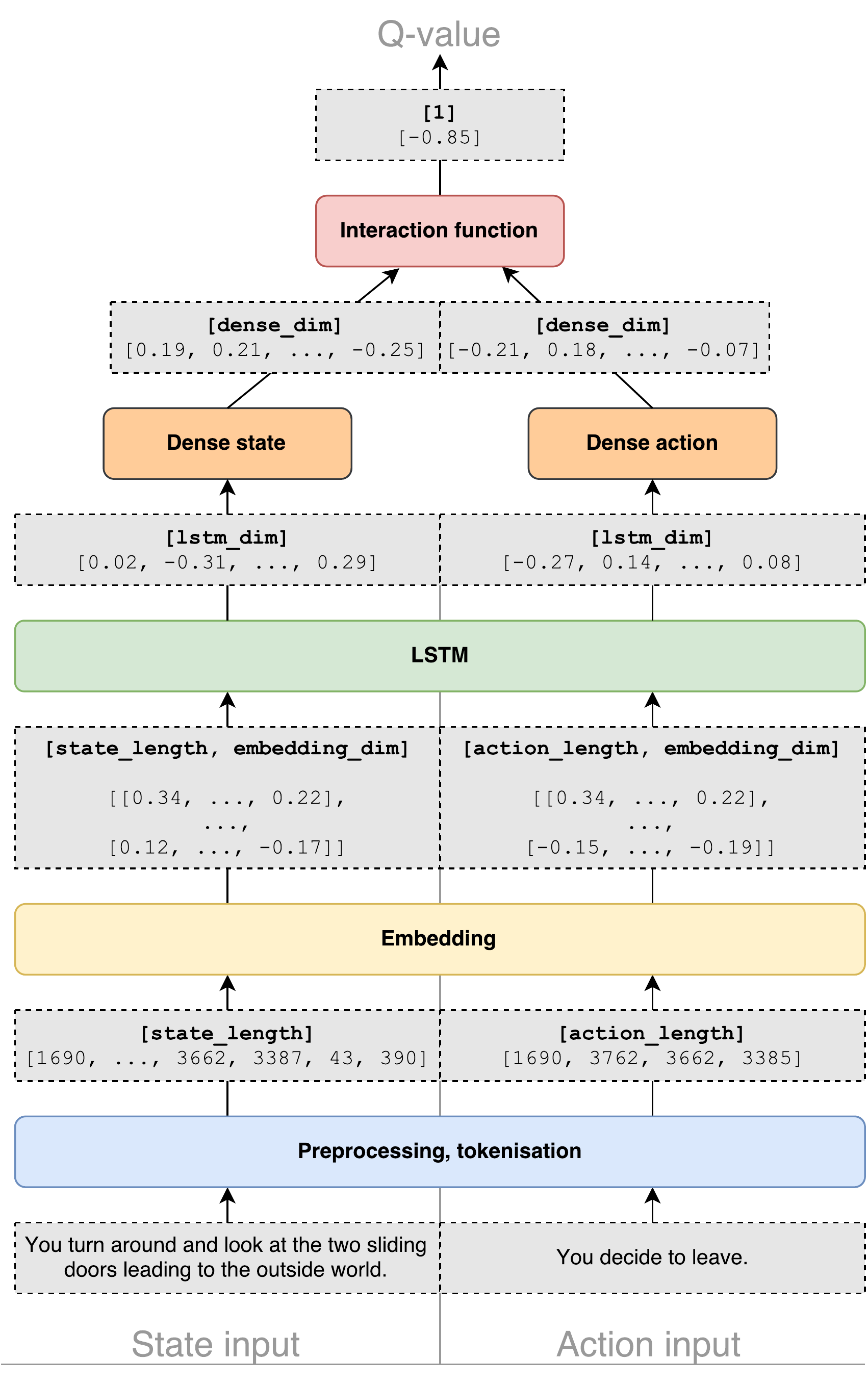}
\caption[Architecture of the SSAQN model and its data flow.]{
Architecture of the SSAQN model and its data flow. Grey boxes represent data values in different layers; the bold text corresponds to the shape of the data~tensors.}
\label{fig:architecture}
\end{figure}

This is best illustrated by visualising the network's computational graph; see figure \ref{fig:architecture} on the next page.
\\\\


As we are using a twin architecture, the weights of the embedding and LSTM layers are shared between state and action data passing through.

States and actions are only differentiated in the dense layers whose outputs are then fed into the similarity interaction function.
\\\\

The most important differences to LSTM-DQN and DRRN are:

\begin{itemize}
\item the network accepts two text descriptions (state and action) of arbitrary length as an input,
\item the embedding and LSTM layers are the same for states and action, i.e. their weights are shared,
\item the interaction function of inner vectors of states and actions is realised by a normalised dot product, commonly called cosine similarity (see sec. \ref{sec:interaction-function}).
\end{itemize}

The output of the SSAQN is the estimated Q-value for the given state-action pair, i.e. the network with parameters $\theta$ realises a function:

\begin{equation}
\theta(s, a) =  Q(s, a, \theta) \approx Q^*(s, a),
\tag{\ref{eq:nn-q}}
\end{equation}
where the input variables $s$ and $a$ contain preprocessed text as described in section~\ref{sec:preprocessing}.

To compute the $Q(s, a^i)$ for different $i$ --- for multiple actions --- we simply run the forward pass multiple times.
\\\\
Next, the SSAQN architecture is described layer-by-layer in more detail. Additional technical details including the network parameters are discussed in section~\ref{sec:technical}.

\newpage

\FloatBarrier
\subsection{Word embeddings}
\label{sec:emb-layer}

After preprocessing the text (see section \ref{sec:preprocessing}), we convert the words to their vector representation using word embeddings (see section \ref{sec:embeddings}).
\\\\
Since our dataset is comparatively small, we use a relatively low dimensionality for the word representations and work with \cd{embedding\_dim} of~$16$.

The weights of the word embeddings are initialised to small random values and trained as a part of the gradient descent algorithm.

Naturally, the weights of the state and action branches are shared, meaning that a word appearing in a state description and in a action descriptions is converted to the same vector in both branches.
\\\\
Importantly, when training, instead of updating single state-action tuples, we perform the updates in batches.
Consequently, we pad the input sequences of words for each batch so that they are all aligned to the same length.

The padding is realised by prepending a special token with an index of~$0$ to the padded sentence.
\\\\
In principle, it is not necessary to pad the sequences or perform the updates in batches, but this leads to a faster training process.
Still, note that the length of the output of the embedding layer is variable in length.

\subsection{LSTM layer}
\label{sec:lstm-layer}

The inputs of the LSTM layer (explained in section \ref{sec:lstm}) are the word embedding vectors of variable length.
\\\\
Similarly to embeddings, the weights of the LSTM units are also initialised to small random values and their weights are shared between states and actions.
\\\\
The role of the LSTM layer is to successively look at the input words, changing its inner state in the process and thus detecting time-based features in its input.

It is also at this layer that we go from having a data shape of arbitrary length to having a fixed-length output vector.

The output size is equal to the number of LSTM units in the layer and in our experiments, we use \cd{lstm\_dim} of~$32$.

\subsection{Dense layer}
\label{sec:dense-layer}

Following the shared LSTM layer, we now have two dense layers (also commonly called fully-connected), one for states and one for actions.
\\\\
Again, we initialise the weights randomly and we also apply the hyperbolic tangent activation function:

\begin{equation}
\label{eq:tanh}
\mathrm{tanh}(x) = \frac{1 - e^{-2x}} {1 + e^{-2x}}.
\end{equation}

As the dense layers for states and actions are the only layers to not necessarily share weights between the two network branches, they do play an important role in building differentiated representations for state and action input data.
\\\\
Note that as the interaction layer uses a dot product, we require both outputs of the dense layers to be of the same dimension and we set \cd{dense\_dim} of both branches to~$8$.

However, theoretically, it would be interesting to use different layer dimensions for states and actions at this level, as usually, the original state text descriptions carry more information than action descriptions in IF games.

Thus, a possible extension of the network would be to use two or more hidden dense layers in the  state branch of the network and to only reduce the dimension to the action dimension in the last hidden dense state layer.

\subsection{Interaction function}
\label{sec:interaction-function}

Lastly, we apply the cosine similarity interaction function to the state and action dense activations, resulting in the final Q-value.

If the input are two vectors $\mathbold{x}$ and $\mathbold{y}$ of~$\cd{dense\_dim} = n$ elements, we define their cosine similarity as a dot product of their L2-normalised (and consequently unit-sized) equivalents:

\begin{equation}
\label{eq:cos}
\mathrm{cs}(\mathbold{x}, \mathbold{y}) = {\frac{\mathbold{x} \cdot \mathbold{y}} {\|\mathbold{x}\|_2 \|\mathbold{y}\|_2}} = \frac{ \sum\limits_{i=1}^{n}{\mathbold{x}_i  \mathbold{y}_i} }{ \sqrt{\sum\limits_{i=1}^{n}{\mathbold{x}_i^2}}  \sqrt{\sum\limits_{i=1}^{n}{\mathbold{y}_i^2}} },
\end{equation}

which corresponds to the cosine of the angle between the input vectors.
\\\\
Cosine similarity is commonly used for determining document similarity (\cite{huang2008similarity}).

Here, we apply it to the two hidden vectors of dense layer values that should meaningfully represent the condensed information that was originally received as a text input by the network and we interpret the resulting value as an indicator of mutual compatibility of the original state-action pair.
\\\\
Obviously, the range of values of the $\cos$ function is $[-1, 1]$, whereas the original rewards that we aim to estimate have arbitrary values.
Therefore, we need to scale the approximated Q-values as discussed in section \ref{sec:scaling-q}.

\subsection{Loss function and gradient descent}

Recall the Q-learning rule (see equation \ref{eq:q}). We define the loss function at time~$t$ as

\begin{equation}
\label{eq:loss}
\mathcal{L}_t = \left( r_{t} + \gamma \cdot \max_{a}Q(s_{t+1}, a) - Q(s_{t},a_{t}) \right)^2,
\end{equation}

which is simply a mean squared error (see equation \ref{eq:mse}) of the last estimated Q-value and the target Q-value.
\\\\
For gradient descent, we make use of the RMSProp optimiser (\cite{rmsprop}) that has been shown to perform well in numerous practical applications, especially when applied in LSTM networks (\cite{DBLP:journals/corr/DauphinVCB15}).

\section{Action selection}
\label{sec:action-selection}

Given an SSAQN $\theta$, where $Q(s,a) \gets \theta(s, a)$, the agent selects an action by following the $\epsilon$-greedy policy $\pi_\epsilon(s)$ (\cite{DBLP:books/lib/SuttonB98}) realised by the following algorithm:

\begin{algorithm}
\caption{Action selection}\label{alg:action-selection}
\begin{algorithmic}[1]
\Statex 
\Statex   $\epsilon$ .. probability of choosing a random action
\Statex   $h(s,a)$ .. number of times the agent selected $a$ in $s$ in the current run
\Statex 
\Function{Act}{$s$, $actions$, $\epsilon$, $h$}
  \If{$random() < \epsilon$}
    \Return {random action} \EndIf
  \State $q\_values = \theta(s, a_i)$ \textbf{for} $a_i$ in $actions$ \label{row:4}
  \State $q\_values = (q\_values + 1) / 2$ \Comment{scale Q-values from $[-1, 1]$ to~$[0, 1]$}
  \State $q_i = q_i^{h(s, a_i) + 1}$ \textbf{for} $q_i$ in $q\_values$ \Comment{apply the history function} \label{row:history}
  \State $q\_values = (q\_values \cdot 2) - 1$ \Comment{scale Q-values from $[0, 1]$ to~$[-1, 1]$}
  \State \textbf{return} $a_i$ with $\max q_i$ \label{row:8} \Comment{\footnotemark}
  
\EndFunction
\end{algorithmic}
\end{algorithm}

\footnotetext{Lines \ref{row:4} to \ref{row:8} can be formally written as:\\\phantom{XX} \textbf{return} $\underset{a_i \in actions}{\mathrm{argmax}_{a_i}} {((((\theta(s, a_i) + 1)/2) ^ {h(s, a_i)}) \cdot 2) - 1 }$}

The $\epsilon$ is the exploration parameter as described in section \ref{sec:q-learning}.

For sampling in training phase (see algorithm \ref{alg:dqn}), $\epsilon$ is gradually decayed from the starting value of~$1$, i.e. at first, the agent's policy is completely random.

In testing phase, the agent is greedy, i.e. $\epsilon$ is set to~$0$ and the agent always chooses the action with the maximum Q-value for the given state.

\subsubsection{History function}

The only important difference between the standard $\epsilon$-greedy control algorithm and our action selection policy is that we additionally employ a \emph{history function}, $h(s, a)$.

The scope of the history function is a single run of the agent on a single game, i.e. it is reset every time a game ends.

$h(s,a)$ returns a value equal to the number of times the agent selected action $a$ in state $s$ in the current run.
That is, if the agent never selects an action twice in the same state during a run, the history function has no impact on action selection.

To be precise, the history function penalises the already visited state-action pairs, as seen on line \ref{row:history} of algorithm \ref{alg:action-selection}.
\\\\
The history function serves as a very simple form of intrinsic motivation (\cite{DBLP:conf/nips/SinghBC04}, \cite{DBLP:journals/tamd/SinghLBS10}).
It is similar to optimistic initialisation (\cite{DBLP:books/lib/SuttonB98}) in that it leads the agent to select previously unexplored state-action tuples.

Additionally, note that the history function is not Markovian in the sense that it takes the whole game episode into account.
In practice, the history function greatly helps the agent to avoid infinite loops, since for many games, it is likely to get stuck in an infinite loop when following a random deterministic Markovian policy (see section \ref{sec:cycles}).

\section{Training loop}

Putting together all the parts introduced above, we can now formally describe the agent's learning algorithm.

We use a variant of DQN (\cite{DBLP:journals/nature/MnihKSRVBGRFOPB15}) with experience replay and prioritised sampling of experience tuples with positive rewards (\cite{DBLP:journals/corr/SchaulQAS15}).

Note that the agent supports playing --- and learning on --- multiple games at once.

\begin{algorithm}
\caption{Training algorithm (a variant of DQN)}\label{alg:dqn}
\begin{algorithmic}[1]
\Statex 
\Statex   $episodes$ .. number of episodes, $b$ .. batch size, $p$ .. prioritised fraction
\Statex   $\epsilon$ .. exploration parameter, $\epsilon\_decay$ .. rate at which $\epsilon$ decreases
\Statex 
\Function{Train}{$episodes$, $b$, $p$, $\epsilon = 1$, $\epsilon\_decay$}

  \State Initialise experience replay memory $\mathcal{D}$
  \State Initialise the neural network $\theta$ with small random weights
  \State Initialise all game simulators and load the vocabulary

  \For {$e \in {0, \ldots, episodes - 1}$}
    \State Sample each game once using $\pi_\epsilon$, store experiences into~$\mathcal{D}$
    \State $batch$ $\gets$ $b$ tuples $(s_t, a_t, r_t, s_{t+1}, a_{t+1})$ from $\mathcal{D}$, where \phantom. \phantom. \phantom. \phantom. \phantom. \phantom. \phantom. \phantom. \phantom. \phantom. \phantom. \phantom. \phantom. \phantom. \phantom. \phantom. \phantom. \phantom.\phantom. \phantom. \phantom. a fraction of~$p$ have $r_t > 0$
   
    \For {$i, (s_t^i, a_t^i, r_t^i, s_{t+1}^i, a_{t+1}^i)$ in $batch$}
      \State $target_i \gets r_t^i$
      \If {$a_{t+1}^i$} \Comment{$s_{t+1}^i$ is not terminal}
        \State $ target_i \mathrel{{+}{=}} \gamma  \cdot \mathrm{max}_{a_{t+1}^i}\theta(s_{t+1}^i, a_{t+1}^i)$ \Comment{$Q(s,a, \theta) = \theta(s,a)$}
      \EndIf
    \EndFor
    
    \State Define loss as $\mathcal{L}_e(\theta) \gets (target_i - \theta(s_t^i, a_t^i))^2$ \Comment{see equation \ref{eq:loss}}
    \State Perform gradient descent on $\mathcal{L}_e(\theta)$
    \State $\epsilon \gets \epsilon \cdot \epsilon\_decay$ \Comment{decrease the exploration parameter (algorithm \ref{alg:action-selection})}
    
  \EndFor
\EndFunction  
\end{algorithmic}
\end{algorithm}

\section{Technical details}
\label{sec:technical}

The architecture described above as well as its optimisation phase was implemented in \textbf{Keras} (version 2.0.4, \cite{chollet2015keras}), using the \textbf{Tensorflow} backend (version 1.1.0, \cite{tensorflow2015-whitepaper}).

For complete reference, the source code of the agent is freely available as part of \textbf{pyfiction} (see appendix \ref{app:pyfiction}).
Additionally, visualisation of the implemented architecture as defined in figure \ref{fig:architecture} is available on page \pageref{fig:architecture-keras}.

\subsection{Parameters}
\label{sec:params}

Both the training algorithm and the neural network use a set of parameters that can dramatically change the course of the learning process.

The employed dimensions of the neural network layers as defined in figure \ref{fig:architecture} are described above, in sections \ref{sec:emb-layer}, \ref{sec:lstm-layer} and \ref{sec:dense-layer}, as they were identical in all our experiments.

All parameters of the training algorithm are summarised in section \ref{sec:exp-params} or given in the sections of the individual experiments.

\subsection{Scaling the Q-values}
\label{sec:scaling-q}

In text games, the player can be given rewards of arbitrary values.

However, note that the Q-value as approximated by SSAQN is a result of the cosine similarity operation (see equation \ref{eq:cos}), thus the approximated value is in $[-1, 1]$.

Since we are trying to approximate the actual game reward, we use this to our advantage and scale the Q-value --- with the knowledge of maximum possible reward in game environment --- to~$[-r_{\mathrm{max}}, r_{\mathrm{max}}]$, where $r_{\mathrm{max}}$ is the largest possible cumulative reward, ergo $r_{\mathrm{max}}$ is a parameter of the game environment.
\\\\
This approach has the additional benefit of the Q-values getting normalised between games with different reward scales.

\subsection{Estimating multiple Q-values effectively}

In section \ref{sec:slsn}, we said that in order to determine Q-values in a single state for different actions, multiple forward passes of the network function $\theta(s,a)$ are needed.

Obviously, there is overhead associated with such approach, as we are evaluating the state branch repeatedly using the same state input text.
\\\\
To compute multiple $Q(s,a_i)$ values effectively, we instead perform a single partial state forward pass and cache the value of the state dense layer activations.

Then, we perform a partial action forward pass for each action and compute the respective Q-value using the cached values of the state dense layer, resulting in up to twice as effective computation when compared to the naive approach.

\section{Summary}

We presented SSAQN, a minimalistic neural network with siamese architecture for efficient estimation of Q-values in the context of the text-game learning task.

SSAQN is powerful in the sense that it accepts and handles input of arbitrary length and is, at least theoretically, capable of capturing long-term language dependencies thanks to the LSTM layer that operates on word embeddings.

Moreover, the employed similarity interaction function applied on inner state and action representations should be able to give good results even for mutually semantically dependent state-action pairs.
\\\\
To see how the model performs in practice, we conduct experiments on several text games in chapter \ref{chp:experiments}.
\chapter{Experiments}
\label{chp:experiments}

\let\oldfig=\thefigure
\renewcommand{\thefigure}{\arabic{chapter}.\arabic{section}.\arabic{figure}} 

In this chapter, we conduct learning experiments using the SSAQN agent described in chapter \ref{chp:agent}.
\\\\
In each experiment, we first explain our motivation, then describe the methods and employed parameters and after presenting the results, we discuss the experiment's implications.
\\\\
We evaluate the agent on a combination of six IF games, namely \emph{Saving John}, \emph{Machine of Death}, \emph{Cat Simulator 2016}, \emph{Star Court}, \emph{The Red Hair} and \emph{Transit} (see section \ref{sec:games} for their description).
\\\\
For more details about the selected games including relevant statistics and ending annotations, refer to appendix \ref{app:text-games} which also includes a table summarising all game parameters as well as the experiment results (see page \pageref{tab:summary}).

\setcounter{figure}{0}
\section{Setup}

In this section, we briefly introduce the IF games we used for the learning tasks and describe the experiment evaluation process.

We also give an overview of methods and parameters common for all experiments.

\subsection{Games}
\label{sec:games}

When selecting IF games for the learning task, we searched for hypertext-based or choice-based games with multiple endings, preferably with a high rating count on the IFDB.
\\\\
The game simulators for the selected games, \emph{Saving John} (SJ), \emph{Machine of Death} (MoD), \emph{Cat Simulator 2016} (CS), \emph{Star Court} (SC), \emph{The Red Hair} (TRH) and \emph{Transit} (TT) are all implemented in pyfiction (see appendix \ref{app:pyfiction}) and share the same agent-environment interface.

For \emph{Saving John} and \emph{Machine of Death}, we wrapped the functionality of game simulators as presented in \cite{DBLP:journals/corr/HeCHGLDO15}.

For all other games, we obtained their publicly available web-based versions and their simulators internally use a web browser as an interface.

\begin{table}[!htb]
\centering
\small
\begin{tabular}{lrrrrrr}
\toprule                                                                     

                                      & SJ & MoD & CS & SC & TRH & TT \\
\midrule                                                                     
\# tokens                             &  1119  &   2055  &   364 &  3929  &  155   &  575  \\
\# states                             & 70   &  $\geq$ 200   &  37  & $\geq$ 420  &  18   &  76  \\
\# endings                            &  5  &  $\geq$ 14  &  8  & $\geq$ 17   &    7 &  10  \\
Avg. words/description                &  73.9  & 71.9    & 74.4   &  66.7  &  28.7   & 87.0   \\
Deterministic transitions                      &  Yes  & No    & Yes   & No   &   Yes  & Yes    \\
Deterministic descriptions                     &   Yes & Yes    &  Yes  &  No  &  Yes   & Yes  \\
Optimal reward                        &   19.4 &   $\approx$ 21.4  &  19.4  &  ?  &  19.3   & 19.5  \\
\bottomrule
\end{tabular}
\caption{Summary of game statistics.}
\label{tab:game-stats}
\end{table}

Table \ref{tab:game-stats} summarises game statistics relevant to the learning task difficulty as discussed in section \ref{sec:game-properties}, including the presence of non-deterministic transitions and descriptions.

\FloatBarrier

Next, we discuss the consequences of the specific non-deterministic properties of the games on the learning task difficulty.

\subsubsection{Machine of Death}
\label{sec:gmod}

In \emph{Machine of Death}, the player is randomly given a prophecy of one of three possible fates very early on in the story when interacting with the Machine of~Death.

After that, there are three completely independent stories based on the randomly selected fate, meaning \emph{Machine of Death} essentially contains three different games.

The game thus does have a few non-deterministic transitions but all its descriptions are deterministic.

As mentioned in section \ref{sec:rewards}, an additional complexity in \emph{Machine of Death} comes from the fact that one of the endings enumerates a number of player's recent actions, meaning there are effectively hundreds of possible final states.
\\\\
To be able to test generalisation properties of their model, \cite{DBLP:journals/corr/HeCHGLDO15} introduced a version of \emph{Machine of Death} with paraphrased action descriptions which we also utilise in our generalisation experiment.

\subsubsection{Star Court}
\label{sec:gsc}

\emph{Star Court} is a story set in distant future where the player is on trial for a crime they likely did not commit and where they have to defend themselves to prove their innocence.

The game is highly random and as even the number of years the player has to spend in prison --- should they be found guilty --- is random, there are effectively thousands of different endings and possible final reward values.

Both the game transitions and game descriptions vary highly and it is very difficult to determine if a good ending can always be reached without relying on luck\footnote{We were unable to determine whether this actually is the case. Anyway, it is \emph{likely} that it is not possible to consistently reach a good ending with a positive reward.\label{ftn:sc}}.

\subsubsection{Saving John, Cat Simulator 2016, The Red Hair, Transit}
\label{sec:gother}

All four remaining games --- \emph{Saving John}, \emph{Cat Simulator 2016}, \emph{The Red Hair} and \emph{Transit} --- are deterministic both in their transitions and in their descriptions.

Consequently, as discussed in section \ref{sec:game-properties}, these games should be similarly difficult to play.

\subsubsection{Difficulty and optimal rewards}
\label{sec:optimal-rewards}

By the criteria established in section \ref{sec:game-properties}, all games but \emph{Machine of Death} and \emph{Star Court} are \emph{simple}, i.e. both their transitions and descriptions are deterministic and can thus be solved by simple search algorithms.
\\\\
Therefore, the optimal cumulative reward is always reachable in any instance of a simple game as seen in table \ref{tab:game-stats}.

The same applies for each branch of \emph{Machine of Death}, i.e. a deterministic policy exists that always finds the maximum possible reward in all three stories.
However, the value of the game's average optimal reward depends on the hidden transition probabilities and we estimate it at about $21.4$.

In \emph{Star Court}, it is yet unknown if an optimal or a positive reward can always be reached\textsuperscript{\ref{ftn:sc}}.

\subsection{Game simulator properties}

The agent interacts with \emph{game simulators} implemented in pyfiction, objects that serve as interfaces between the agent and the game.

There are slight differences between the underlying games and the output of their simulators for the purposes of the learning task.
\\\\
First of all, the game simulators remove all special game elements such as images or sounds.

Secondly, in IF games, available actions are usually presented in the same way and in particular, in the same order.
Consequently, the game simulator randomly shuffles the game actions in each step.
This is to force the agent to consider the contents of the actions and to make it impossible to only learn to act based on the action order.

Lastly, we impose a limit on the maximum number of steps the agent takes in the game in order to break otherwise infinite cycles.
Its value is set to~$500$ for \emph{Machine of Death} and \emph{Star Court} and to~$100$ for all other games.

\subsection{Evaluation}

To evaluate an agent's performance on a game, we simply let the agent play the game using the action selection mechanism as described in section \ref{sec:action-selection} with $\epsilon$ set to zero, ergo the agent uses the greedy policy $\pi_0(s)$ and is deterministic.

The reward of a single game episode is simply a sum of the rewards obtained in all timesteps of the episode.
\\\\
More precisely, the agent is tested at the end of every $n$th episode of the training process, i.e. right after the episode's weight update  (see algorithm \ref{alg:dqn}).

In deterministic games, the agent is only evaluated once in each $n$th episode as its behaviour is also deterministic.

In \emph{Machine of Death} and \emph{Star Court}, we let the agent play the game five times in each $n$th episode and record the average of the obtained rewards and their standard deviation.

We set $n$ to~$1$ when playing deterministic games individually and to either $8$ or $16$ in all other cases to speed the training process up.

\subsection{Parameters}
\label{sec:exp-params}

See table \ref{tab:params} for a summary of used parameter values as defined in chapter \ref{chp:agent}.

Unless specified otherwise, we use the default values for built-in parameters of both Keras and Tensorflow.

\begin{table}[!htb]
    
    \begin{subtable}[b]{.5\textwidth}
      \centering
        
        \begin{tabular}[b]{lr}
        \toprule
          Layer& Dimension \\
          \midrule
          Embedding & 16 \\
          LSTM & 32 \\
          Dense & 8 \\
          \bottomrule
        \end{tabular}
        \caption{SSAQN layer dimensions}
    \end{subtable}%
    \begin{subtable}[b]{.5\textwidth}
      \centering
        
        \begin{tabular}[b]{lr}
        \toprule
         Parameter & Value \\
         \midrule
          Optimiser & RMSProp \\
          Learning rate & \emph{variable} \\
          Batch size & 256 \\
          $\gamma$ & 0.95 \\
          $\epsilon$ & 1 \\
          $\epsilon$-decay & \emph{variable} \\
          Prioritised fraction & 0.25 \\
          \bottomrule
        \end{tabular}
        \caption{Training algorithm parameters}
    \end{subtable} 
    \caption{Parameters used in the learning tasks.}
    \label{tab:params}
\end{table}

\setcounter{figure}{0}
\section{Individual games}
\label{sec:individual-games}

Firstly, we train the agent on individual games and evaluate its performance on the same data, i.e. train datasets are equal to test datasets.
\\\\
The results are compared to the random agent baseline (labelled \cd{random}) and additionally, for \emph{Saving John} and \emph{Machine of Death}, we are able to compare the results to the DRRN model from \cite{DBLP:journals/corr/HeCHGLDO15}.

Note that for all following experiments, we additionally use the result of this individual game task (labelled \cd{test}) as a baseline.

\newpage
\subsection{Setting}

We run a separate experiment for each game using the learning parameters as seen in table \ref{tab:individual}.

Recall that the evaluation is done using the agent's greedy test policy realised by the SSAQN, i.e. $\epsilon = 0$.

\begin{table}[htb!]
\centering
\begin{tabular}{lrr}
\toprule
                   & Learning rate & $\epsilon$-decay \\
\midrule
Saving John        & 0.00010        & 0.990             \\
Machine of Death   & 0.00001       & 0.999            \\
Cat Simulator 2016 & 0.00010        & 0.990             \\
Star Court         & 0.00001       & 0.999            \\
The Red Hair       & 0.00010        & 0.990             \\
Transit            & 0.00100         & 0.990             \\
\bottomrule
\end{tabular}
\caption{Learning parameters for the individual games task.}
\label{tab:individual}
\end{table}

\captionsetup[subfigure]{font=scriptsize,labelfont=scriptsize}
\subsection{Results}

\begin{figure}[htb!]
\begin{subfigure}{.5\textwidth}
\includegraphics[width=\linewidth]{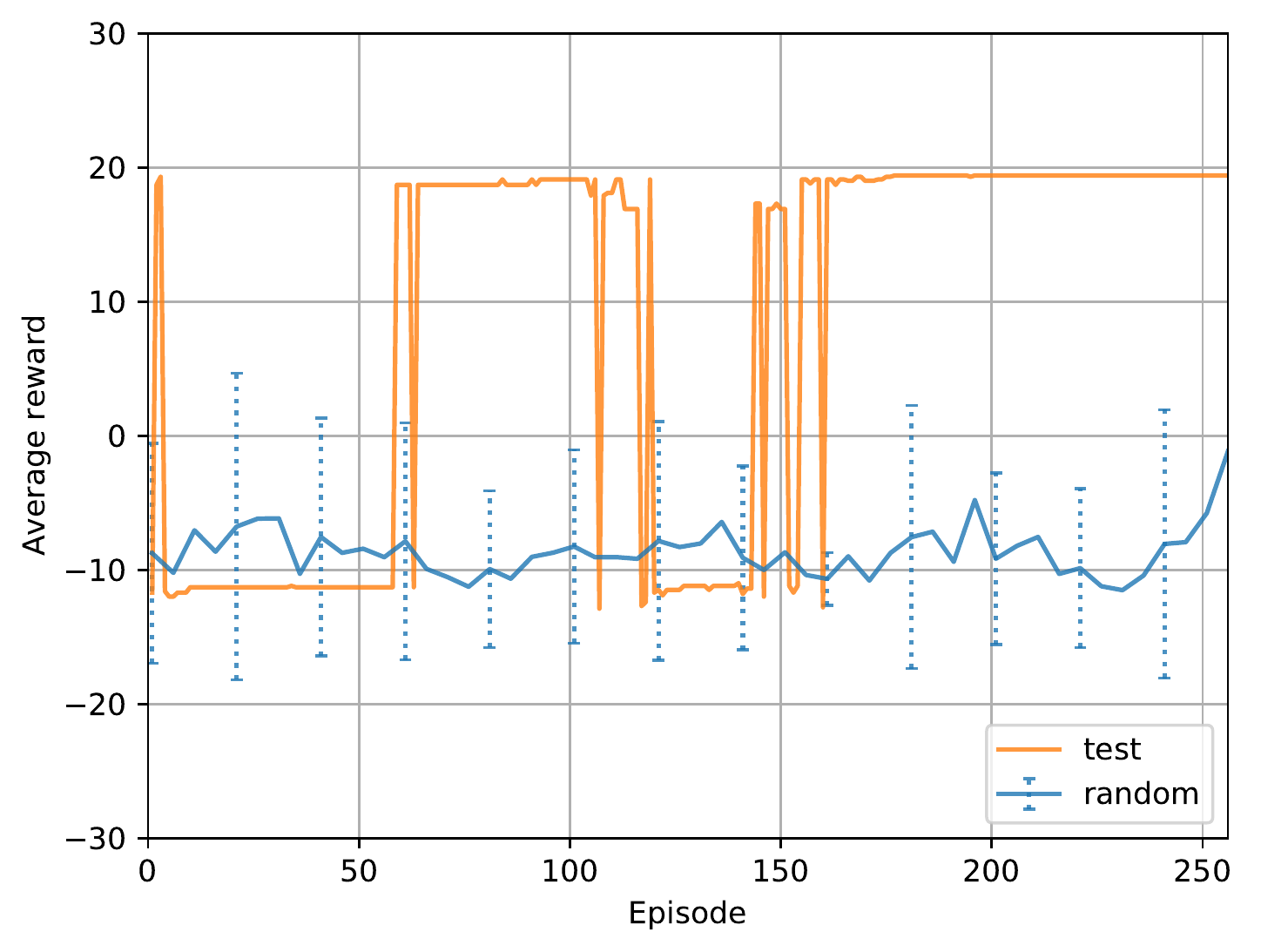}
\caption{Saving John}
\end{subfigure}
\begin{subfigure}{.5\textwidth}
\includegraphics[width=\linewidth]{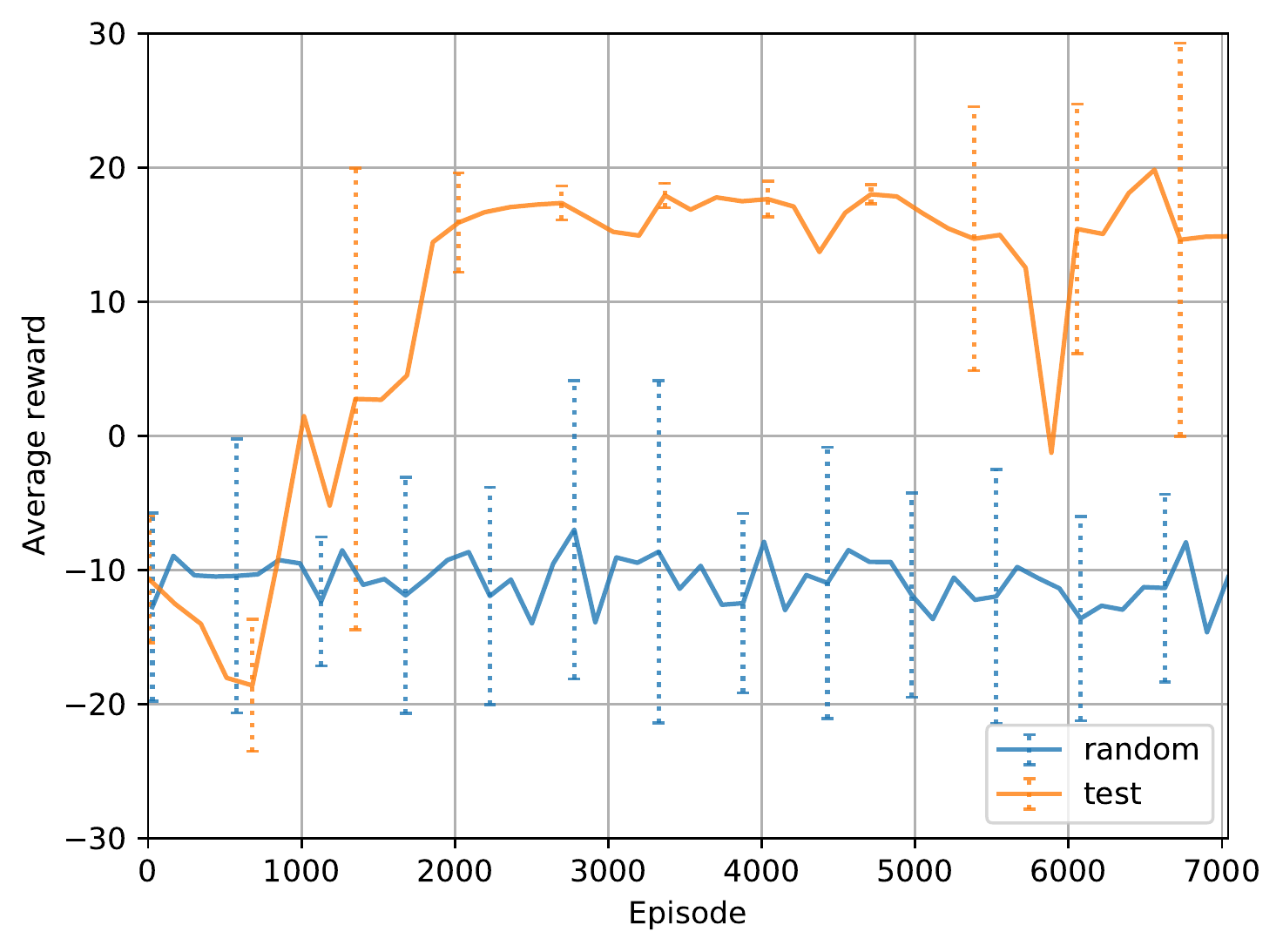}
\caption{Machine of Death}
\label{res:rmod}
\end{subfigure}
\begin{subfigure}{.5\textwidth}
\includegraphics[width=\linewidth]{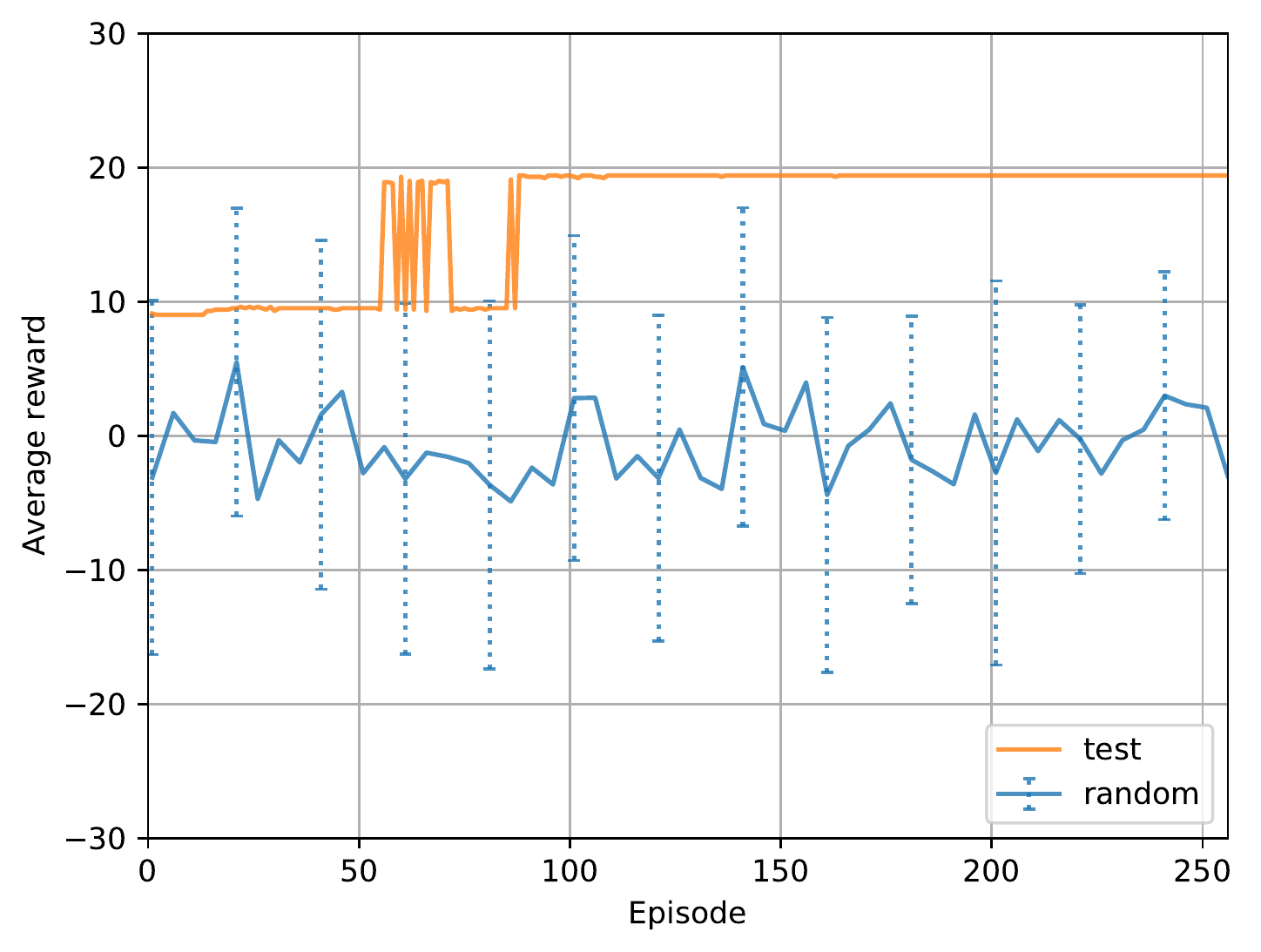}
\caption{Cat Simulator 2016}
\end{subfigure}
\begin{subfigure}{.5\textwidth}
\includegraphics[width=\linewidth]{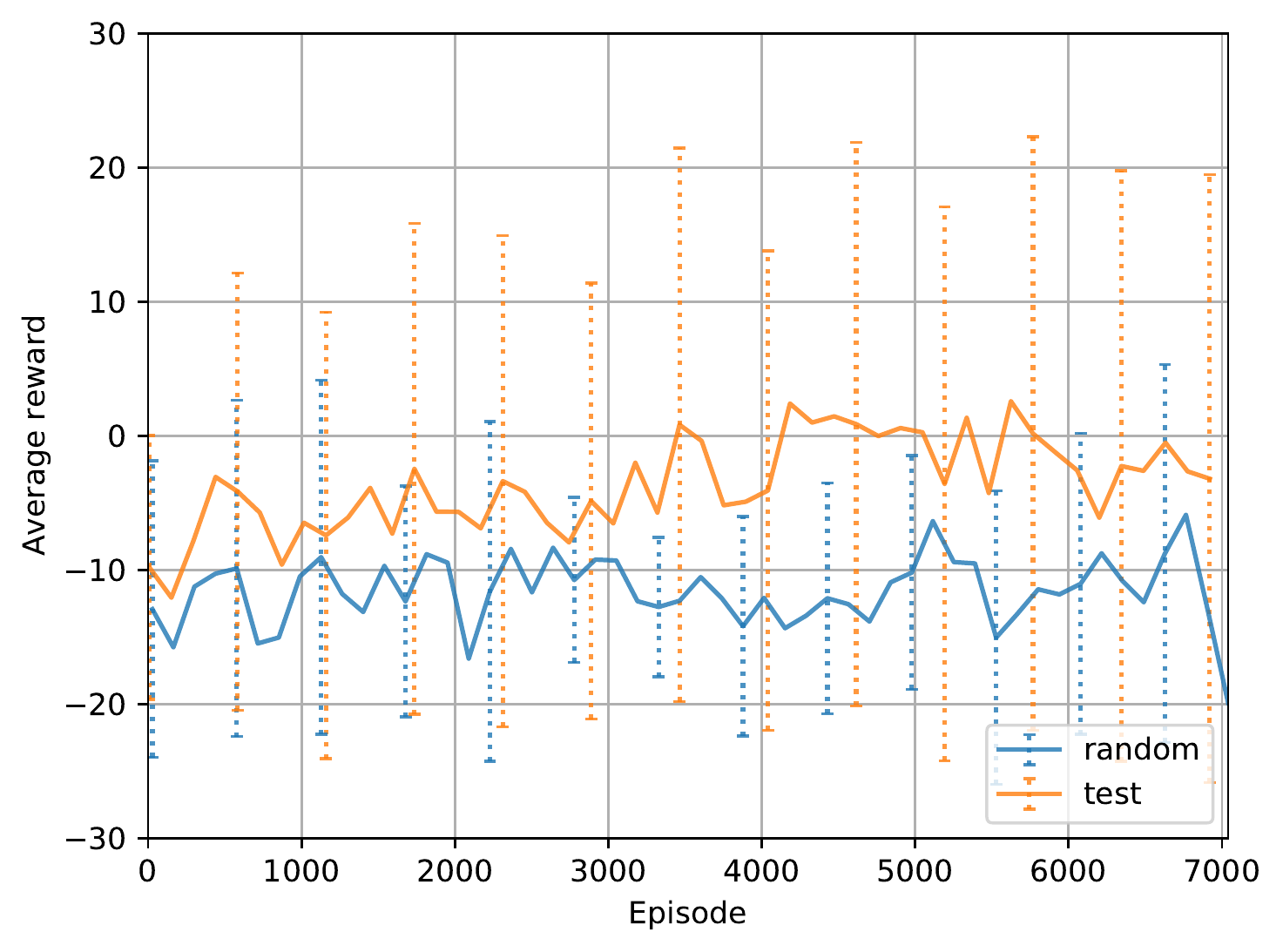}
\caption{Star Court}
\label{res:rsc}
\end{subfigure}
\begin{subfigure}{.5\textwidth}
\includegraphics[width=\linewidth]{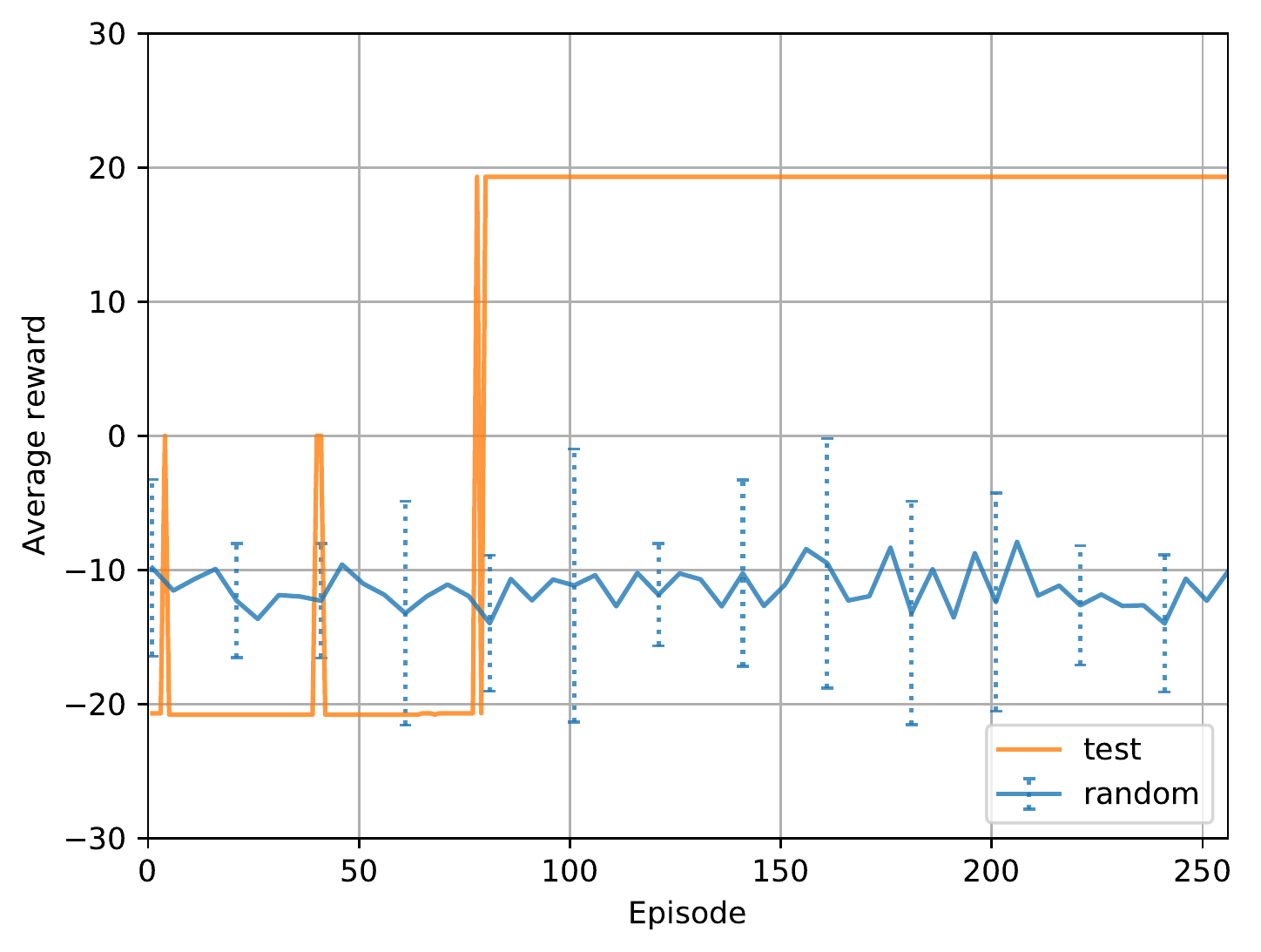}
\caption{The Red Hair}
\end{subfigure}
\begin{subfigure}{.5\textwidth}
\includegraphics[width=\linewidth]{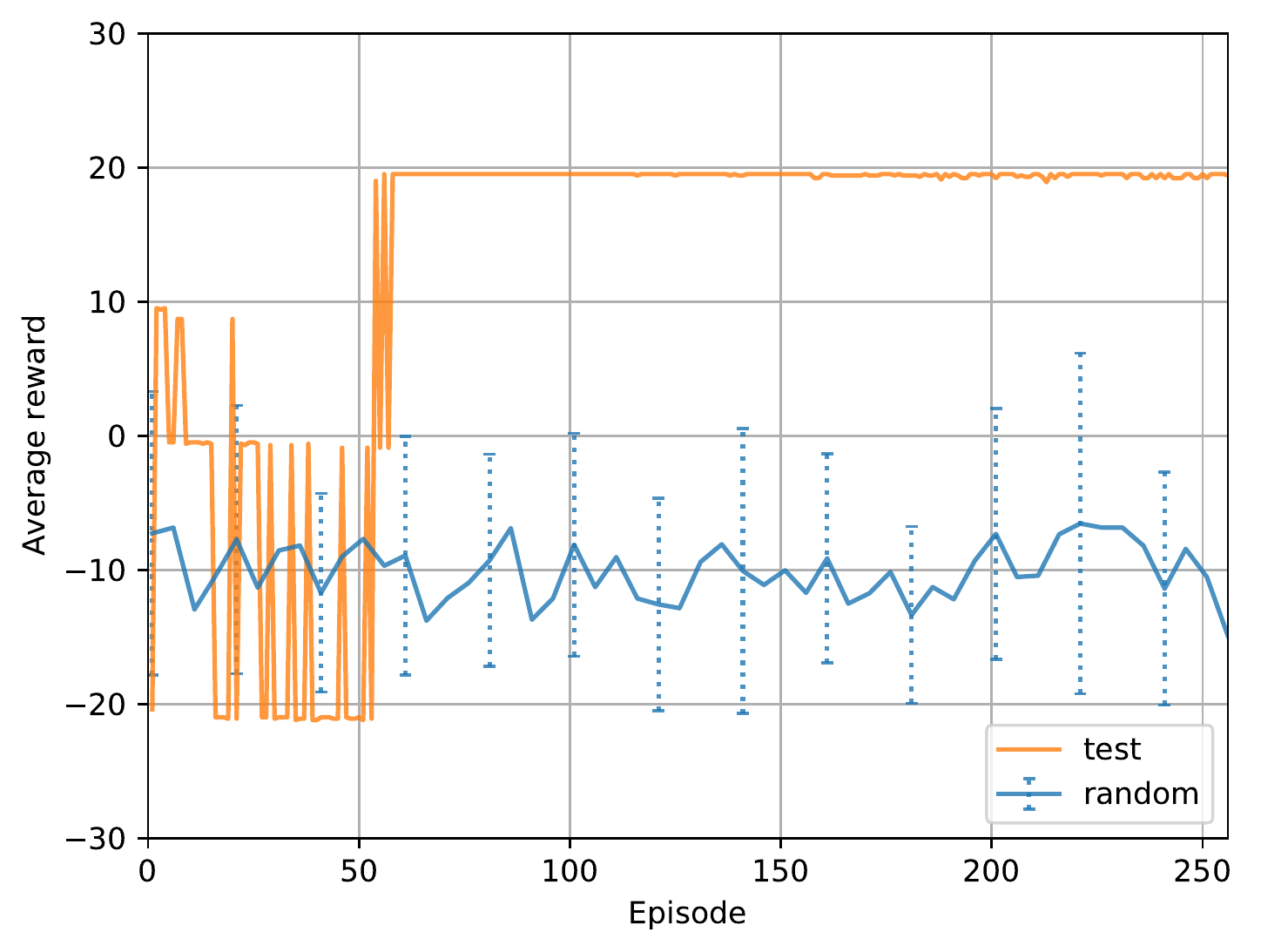}
\caption{Transit}
\end{subfigure}
\caption{Performance of the SSAQN agent in individual games.}
\label{res:random}
\end{figure}

Results are depicted in figure \ref{res:random} on page \pageref{res:random}.
\\\\
In all fully deterministic games, the agent converges to an optimal policy\footnote{Recall that a policy is optimal if it maximises the cumulative reward (see section \ref{sec:policy-updating}). Consequently, the agent converged to the best possible total reward.} in under $250$ episodes.

In \emph{Machine of Death} (figure \ref{res:rmod}), the agent did not converge in the given number of episodes as indicated by the high standard deviations of the average reward. Nevertheless, the resulting policy performs significantly better than the random baseline.

In \emph{Star Court} (figure \ref{res:rsc}), the agent performs better than the random baseline on average but it is difficult to judge the optimality of the resulting policy (see section \ref{sec:optimal-rewards}).

\FloatBarrier
\subsubsection{Comparison to DRRN}

\begin{figure}[htb!]
\begin{subfigure}[t]{.5\textwidth}
\includegraphics[width=\linewidth]{img/exp/random_sj.pdf}
\caption{SSAQN on Saving John}
\label{res:rsj}
\end{subfigure}
\begin{subfigure}[t]{.5\textwidth}
\includegraphics[width=\linewidth]{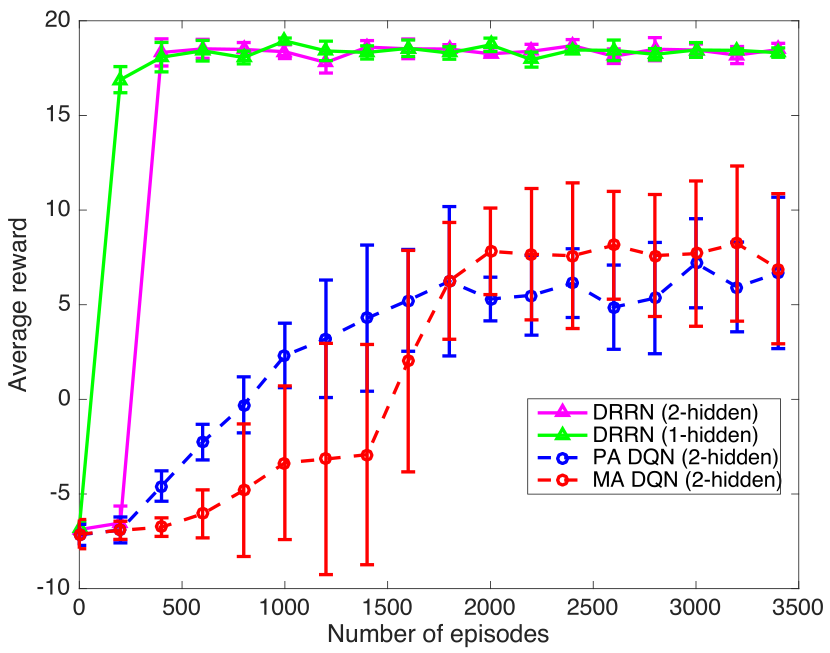}
\caption{DRRN on Saving John}
\end{subfigure}
\begin{subfigure}[t]{.5\textwidth}
\includegraphics[width=\linewidth]{img/exp/random_mod.pdf}
\caption{SSAQN on Machine of Death}
\end{subfigure}
\begin{subfigure}[t]{.5\textwidth}
\includegraphics[width=\linewidth]{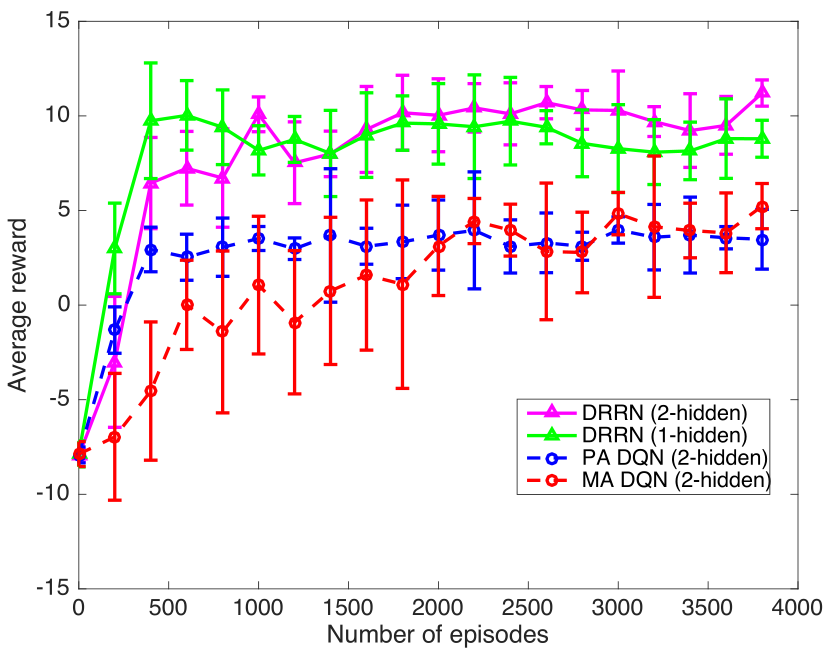}
\caption{DRRN on Machine of Death}
\end{subfigure}
\caption{Comparison of SSAQN and DRRN on Saving John and Machine of Death.}
\label{res:drrn}
\end{figure}

In figure \ref{res:drrn}, we compare the performance of SSAQN and DRRN on \emph{Saving John} and \emph{Machine of Death}.

Note that the speed of convergence is incomparable as both algorithms perform slightly different operations in a single episode.

\FloatBarrier

In \emph{Saving John}, the SSAQN agent converged to the optimal reward of~$19.4$, whereas the DRRN agent converged to a nearly-optimal reward of~$18.7$ (\cite{DBLP:journals/corr/HeCHGLDO15}).

The fluctuations in DRRN's performance are likely caused by its employment of softmax selection.
\\\\
In \emph{Machine of Death}, the DRRN agent converged to a local optimum of~$11.2$. The SSAQN agent did not converge in this number of iterations and the performance deviations are much higher. However, the average reward is noticeably higher than in DRRN, at about $15.4$.

When looking at the raw performance data of SSAQN on \emph{Machine of Death}, we see that the high standard deviations are caused by its volatility in one of the three game branches where the reward oscillates highly.

In the other two game branches, SSAQN converged to the optimal rewards of~$28.5$ and $18.4$ respectively.

\FloatBarrier
\subsection{Discussion}
\label{sec:inddisc}
The agent was able to learn to play all games well barring \emph{Star Court}, in which it is difficult to define the optimal solution and thus judge the agent's performance.

It will be more interesting to see how different tasks influence agent's performance in \emph{Star Court}.
\\\\
In \emph{Machine of Death}, a higher number of iterations or employment of other techniques such as learning rate decay would likely be needed to converge to a local optimum on the third game branch.

Still, the resulting policy was better than that of the DRRN agent on average.

We hypothesise that this is thanks to the SSAQN's ability to extract a higher number of features from the same amount of data, as it is able to capture sentence-level word relations and to distinguish between a number of states and actions represented identically in DRRN.

This difference is consequently more pronounced on \emph{Machine of Death}, where there are significantly more states and actions than in \emph{Saving John} in which SSAQN only performs slightly better than DRRN.

\subsubsection{Learning instabilities}

An interesting property of the learning curves (fig. \ref{res:rsj} being the prime example) of the simple games is their high instability.

We believe these instabilities exist because in a single episode, there is a comparatively low number of decisions to make. Additionally, IF games can often be extremely sensitive and volatile (or, in the IF terms, ``unforgiving'') and a single wrong action can result in falling into a game branch with very low rewards without being to able to correct the mistake.

Taken together, this means that even a very slight change in the policy affecting only a single decision can often significantly impact the final reward, which is what we see in the results.

Compare this to the Atari games platform where in each episode, the agent performs thousands of updates on average. As a result, the impact of a single decision is much less significant than in IF games.

\setcounter{figure}{0}
\section{Generalisation}
\label{sec:generalisation}

Next, we test how the agent performs in terms of generalisation.

The ability to generalise, or, in our context, the ability to perform well in previously unseen games, is a crucial factor determining the power and the usefulness of the model.

\subsection{Setting}
\label{sec:gen-setting}

For this purpose, we train the agent using the leave-one-out method, in which five out of the six games are provided to the agent as training simulators and the agent's performance is simultaneously evaluated on the sixth game, unseen during training.

The vocabulary provided to the agent is a unification of the vocabularies of all six individual games consisting of~$5567$ tokens.

In this section, referring to an experiment by a game name means that this specific game was left out during training phase and only used for testing.
\\\\
We used the same parameters for all games; a learning rate of~$0.00001$ and an~$\epsilon$-decay of~$0.999$.

\subsection{Results}

\begin{figure}[htb!]
\begin{subfigure}{.5\textwidth}
\includegraphics[width=\linewidth]{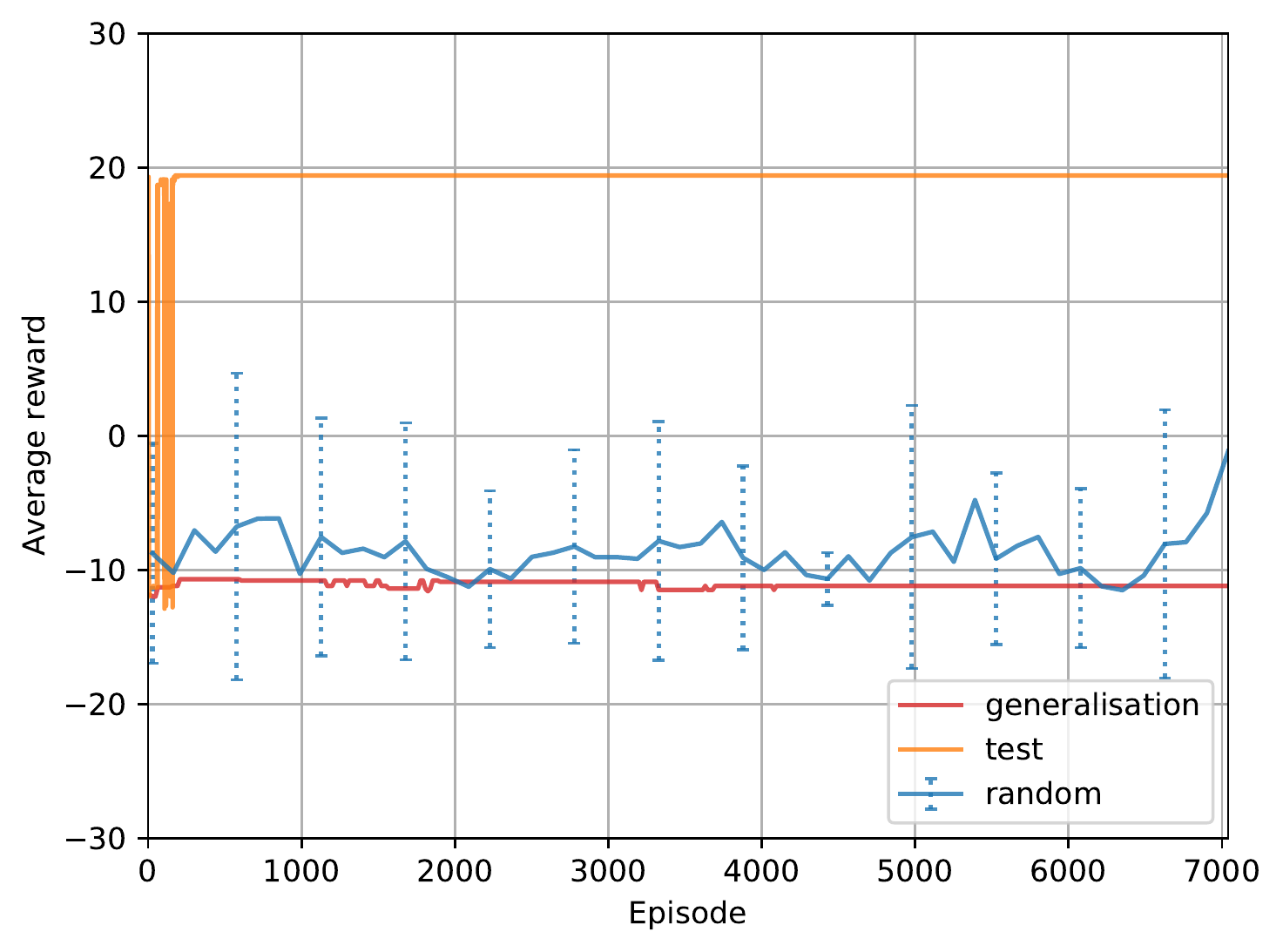}
\caption{Saving John}
\end{subfigure}
\begin{subfigure}{.5\textwidth}
\includegraphics[width=\linewidth]{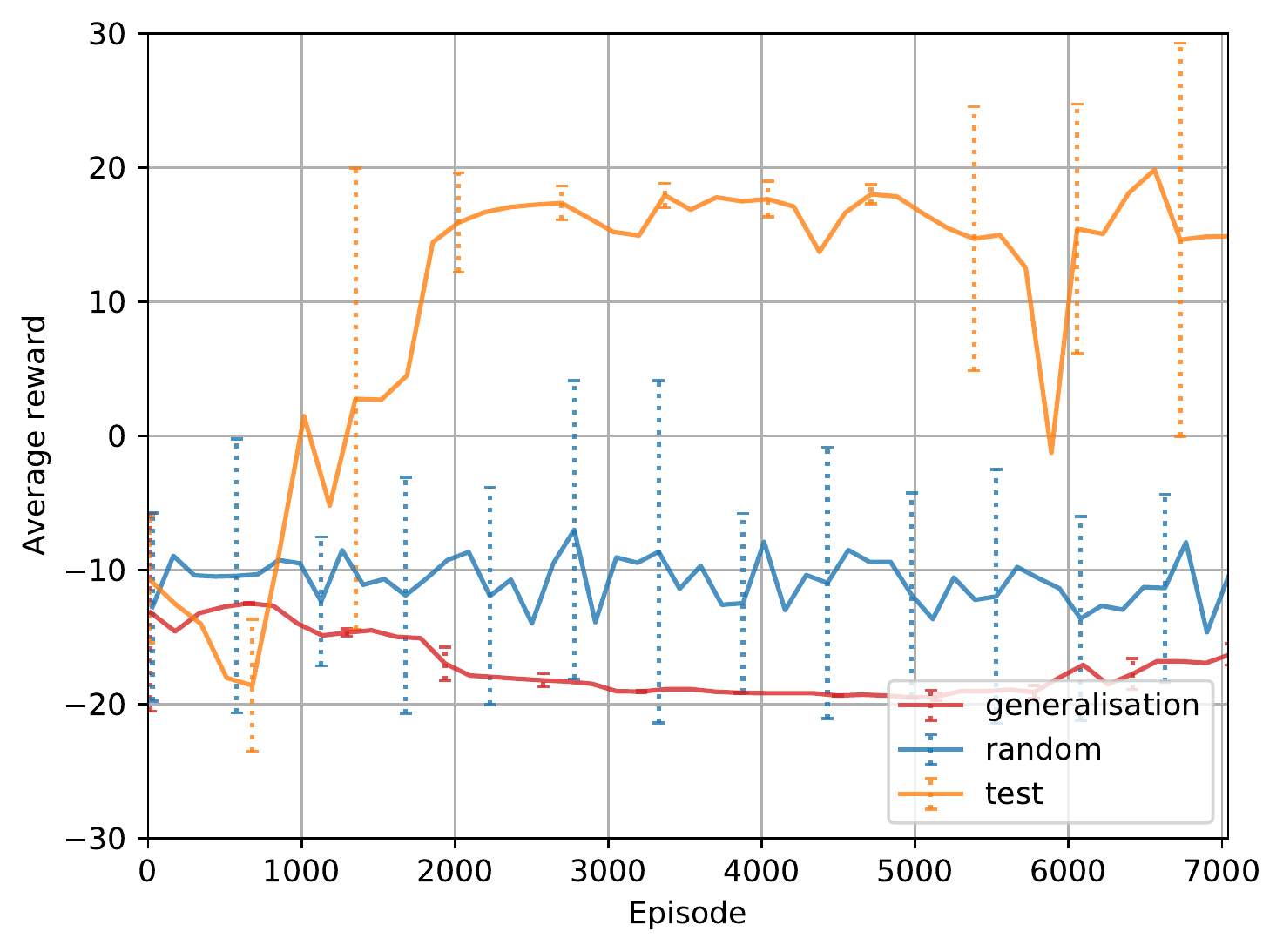}
\caption{Machine of Death}
\end{subfigure}
\begin{subfigure}{.5\textwidth}
\includegraphics[width=\linewidth]{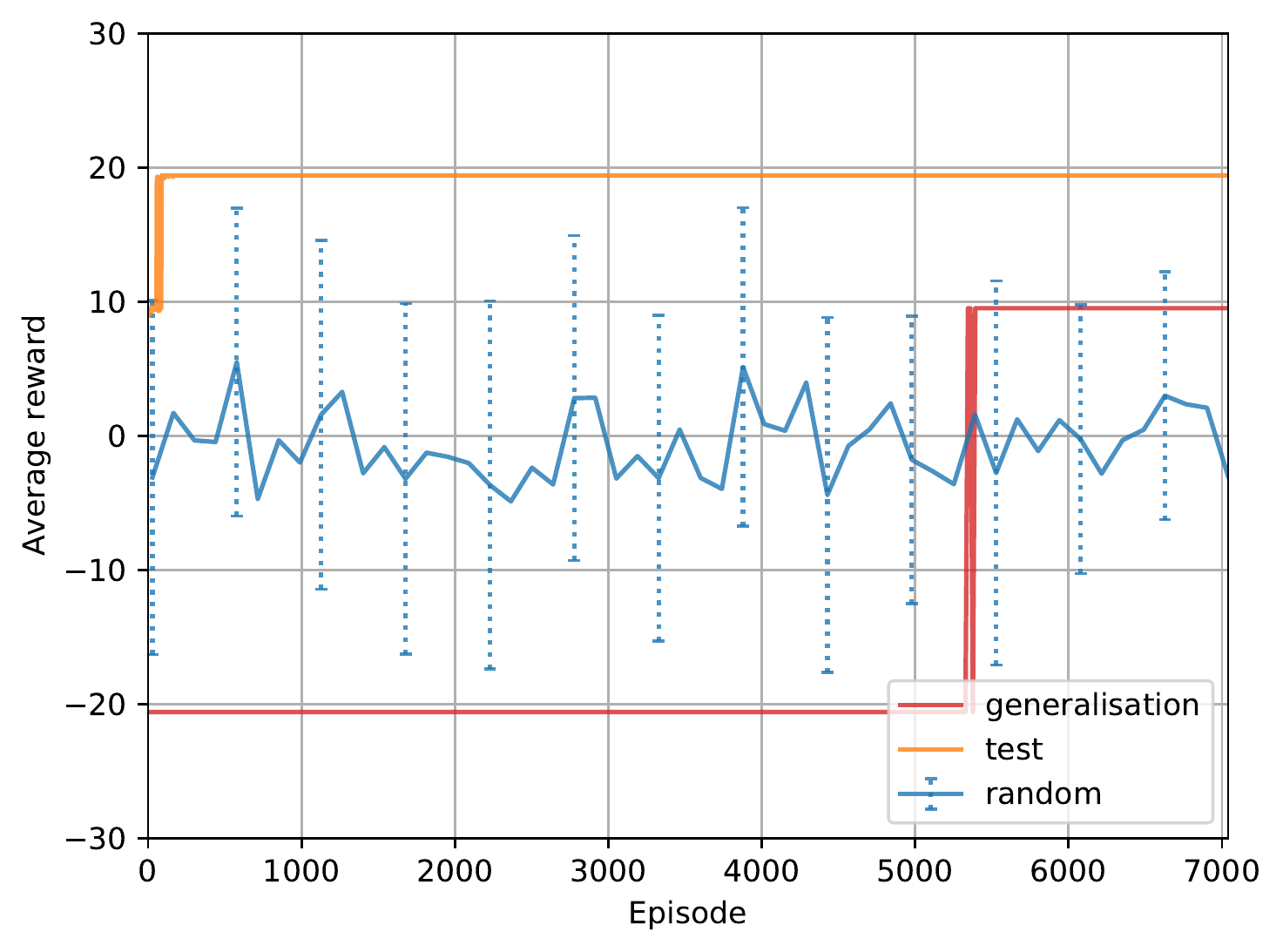}
\caption{Cat Simulator 2016}
\end{subfigure}
\begin{subfigure}{.5\textwidth}
\includegraphics[width=\linewidth]{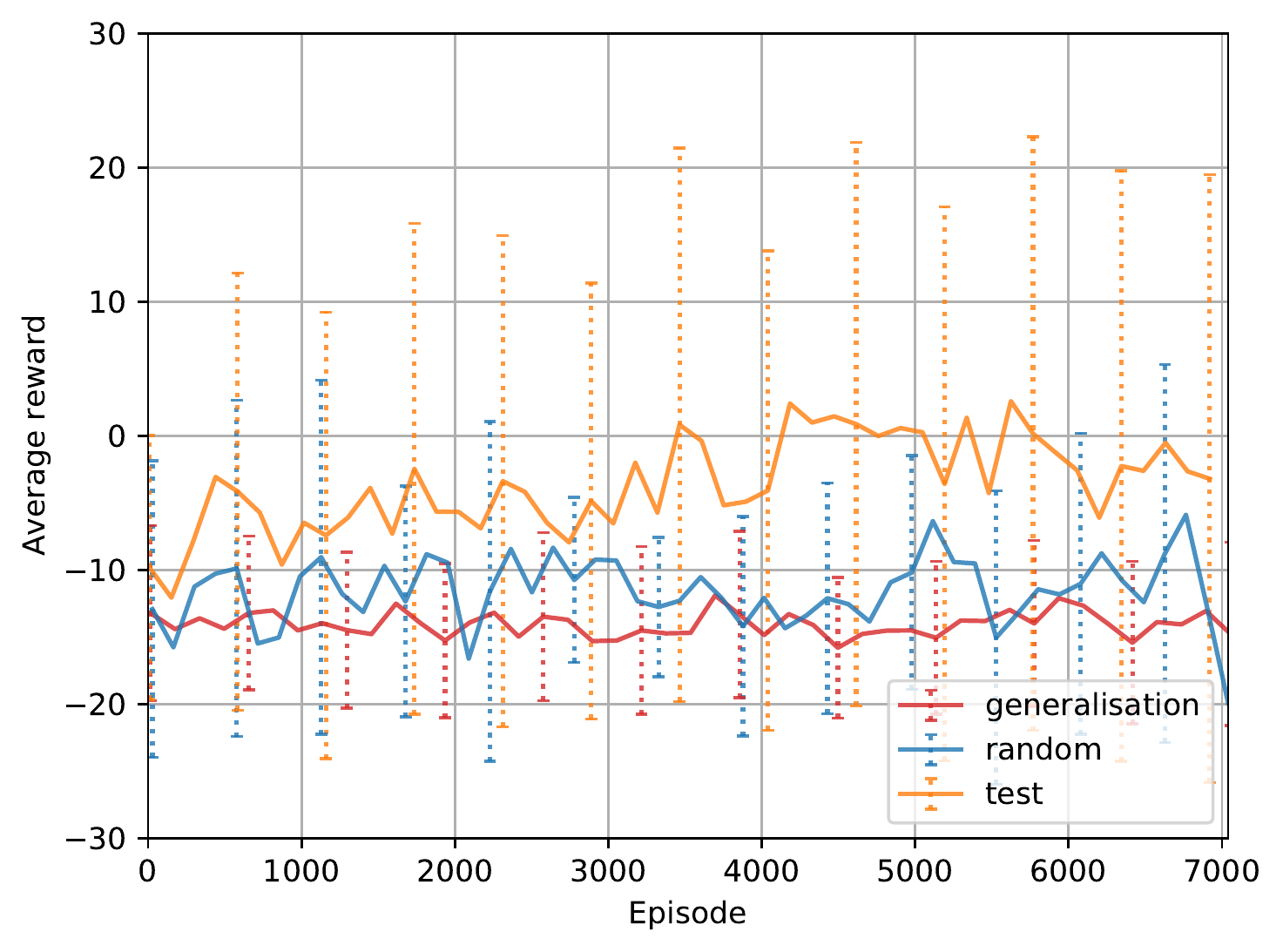}
\caption{Star Court}
\end{subfigure}
\begin{subfigure}{.5\textwidth}
\includegraphics[width=\linewidth]{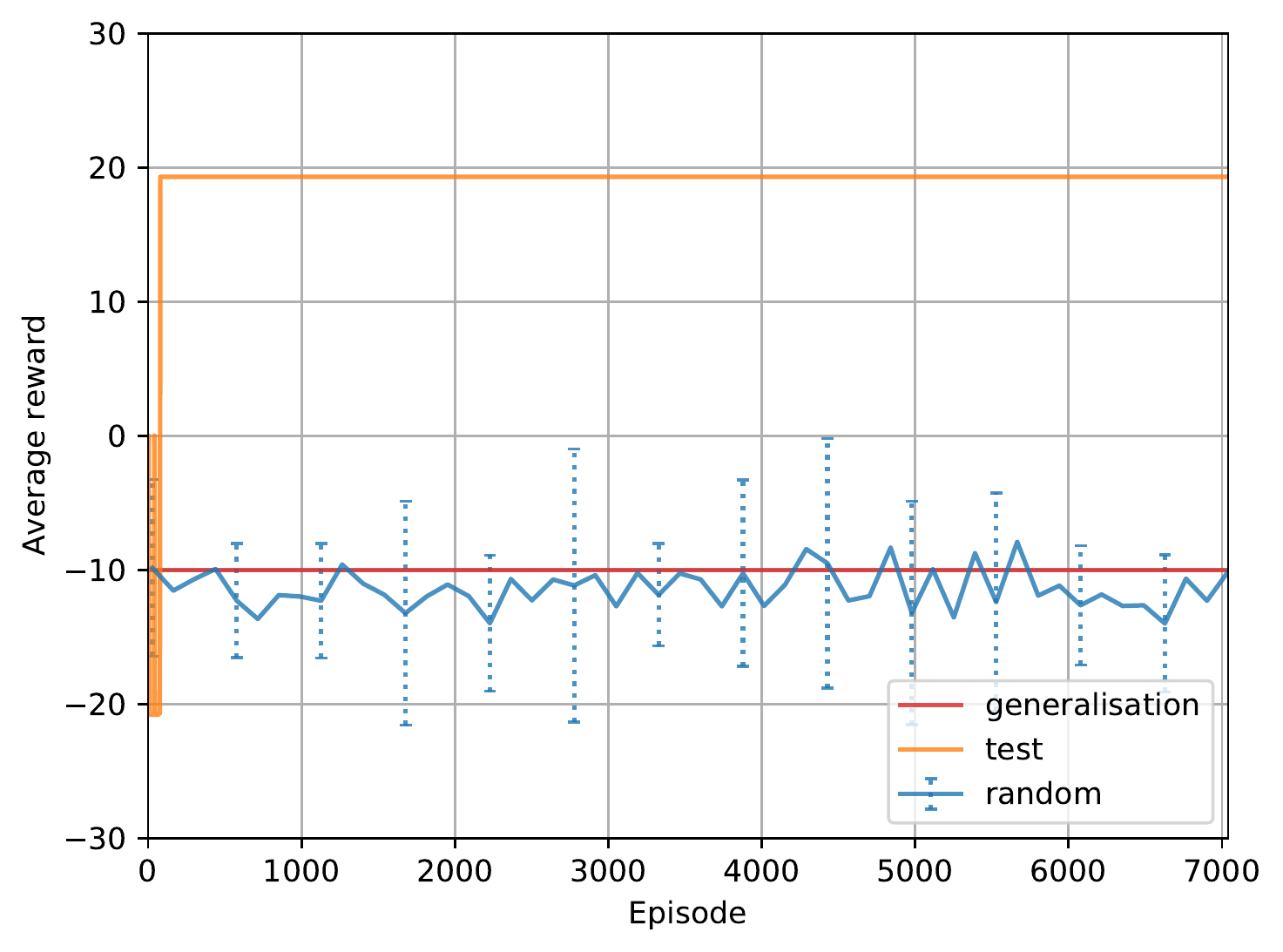}
\caption{The Red Hair}
\end{subfigure}
\begin{subfigure}{.5\textwidth}
\includegraphics[width=\linewidth]{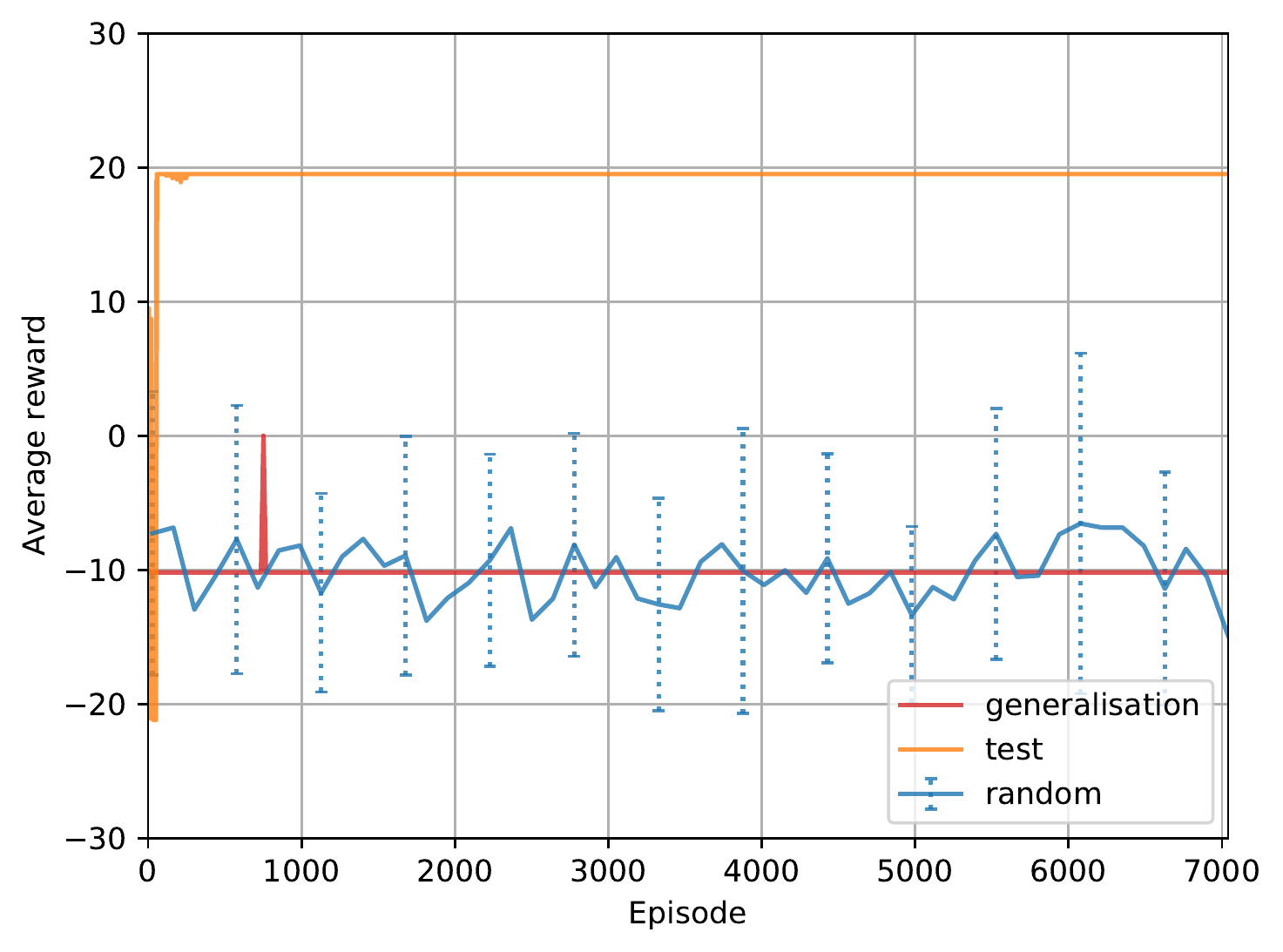}
\caption{Transit}
\end{subfigure}
\caption{Generalisation performance.}
\label{res:generalisation}
\end{figure}

Results are shown in figure \ref{res:generalisation}.

\FloatBarrier
\newpage
\subsection{Discussion}

Unfortunately, the agent cannot generalise on the given dataset.

The result is not surprising, and as we can also see from the flat learning curves, the process of training on different games has hardly any impact on the performance in the unseen game.

This is partly due to the incomplete coverage of the state and action natural language spaces of the train and test domains, i.e. some words of the tested game do not appear at all in the training games.
Or, even if they are present in at least one training game (see table \ref{tab:word-coverage}), their appearance therein might be minimal or insignificant and consequently, without much impact on the specific word embedding.

\begin{table}[!htb]
\centering
\small
\begin{tabular}{lrrrrrr}
\toprule
                                  & SJ & MoD & CS & SC & TRH & TT \\
\midrule
\% tokens present in other games  &  68.4 & 56.0  & 79.4   &  33.7  &  92.3   & 72.7   \\
\bottomrule
\end{tabular}
\caption[Shared vocabulary statistics.]{Shared vocabulary statistics. A token is shared between a game and other games if it is present in at least one other game.}
\label{tab:word-coverage}
\end{table}

Most importantly, though, generalisation is hard in IF games because the individual games do not necessarily share objectives and actions that lead to good rewards in one game are often very bad choices in another.

IF games are works of art created by humans and as such, there is no universal correct decision-making policy valid in different games and different contexts, as it obviously is very common for humans to have different opinions on what action is the right choice in a given context.
\\\\
We hypothesise that even if a set of games shared the natural language state and action spaces and we tested a well-trained model on a previously unseen game, the resulting performance would --- on average --- not be significantly better than that of a random agent.

As discussed in section \ref{sec:inddisc}, another factor that supports this hypothesis is the fact that there are comparatively few decisions to make in a single game run and only failing once is often sufficient to fail completely.

Thus, the impact of individual decisions is very high and it is very likely that, when generalising, at least one ``correct'' decision in the new game will not be selected by the policy learned on different games.

\subsubsection{Paraphrased actions}

\cite{DBLP:journals/corr/HeCHGLDO15} used a modified version of \emph{Machine of Death} for testing generalisation abilities of their model.

In the alternative version, all actions descriptions are paraphrased in natural language and the state descriptions remain unchanged.

The DRRN agent trained on the original actions was shown to generalise well to the paraphrased dataset, reaching a reward of~$10.5$ as compared to the original reward of~$11.2$.
\\\\
We were unable to reproduce this behaviour with our model, or, more precisely, we were unable to reach this performance \emph{consistently}.

In reality, different instances of our model pre-trained on the original actions that reached almost identical performance in the original game, performed very differently in the paraphrased version of the game.
Some were on par with the original model, reaching better generalisation results than DRRN, but some performed significantly worse.

This leads us to believe that the evaluation of such experiments should be done in a continuous manner, similarly to how we evaluate our generalisation experiments, i.e. by continuously monitoring the test performance on the paraphrased dataset while training on the original dataset and thus averaging the results.
\\\\
The performance of our agent while following exactly this approach can be seen in figure \ref{res:paraphrased}:

\begin{figure}[htb!]
\centering
\includegraphics[width=.5\textwidth]{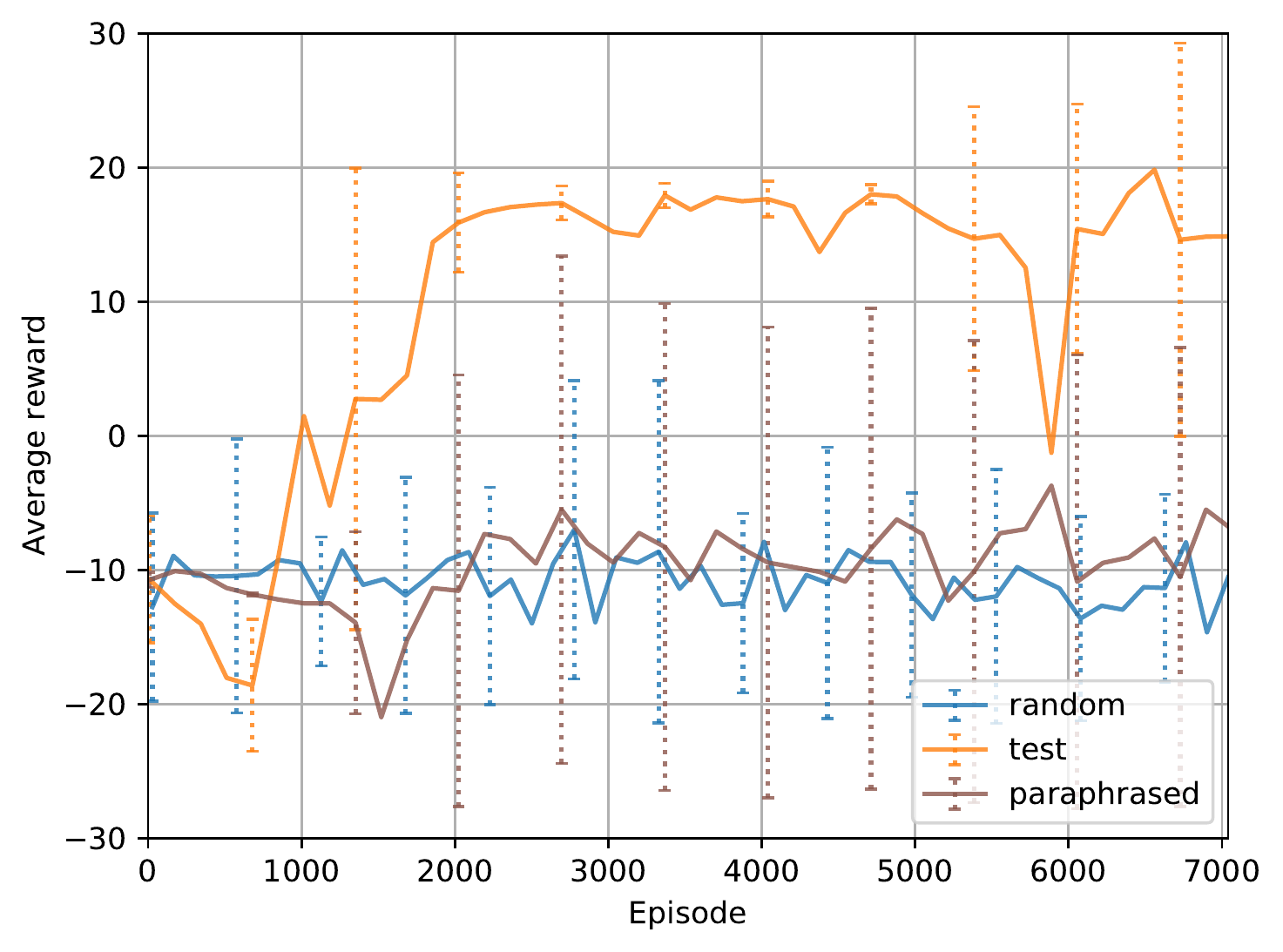}
\caption{Generalisation on paraphrased actions in \emph{Machine of Death}.}
\label{res:paraphrased}
\end{figure}

Unfortunately, the agent was able to generalise somewhat successfully in only one of the three game branches.

Generally speaking, though, it is very important to evaluate generalisation experiments carefully.
Especially in the context of IF games that are often volatile (see section \ref{sec:inddisc}), it is common to simply ``get lucky'' and find an instance of a policy resulting in a high reward despite not actually being able to understand the underlying text.

In fact, we run a random agent on \emph{Machine of Death} and observe that it receives a positive cumulative reward in $8.12 \%$ of games across $10000$ simulation episodes.

\FloatBarrier

\setcounter{figure}{0}
\section{Transfer learning}
\label{sec:transfer}

Following the generalisation experiment, we now test the agent's transfer learning ability, i.e. the ability to make use of the knowledge obtained on a different dataset when being confronted with a new domain.

Even if the agent was not able to generalise, it is still possible that its obtained knowledge will eventually lead to better results when facing a different environment.

\subsection{Setting}

In this experiment, the weights of the model are not initialised to random values.

Instead, the agent is given the model trained on five different games in the generalisation experiment (see section \ref{sec:generalisation}) and we start training it on the sixth, previously unseen game.

We also use the same unified vocabulary as in the generalisation experiment (see section \ref{sec:gen-setting}).
\\\\
We used the same parameters as in the generalisation experiment, i.e. a learning rate of~$0.00001$ and $\epsilon$-decay of~$0.999$.

\subsection{Results}

\begin{figure}[htb!]
\begin{subfigure}{.5\textwidth}
\includegraphics[width=\linewidth]{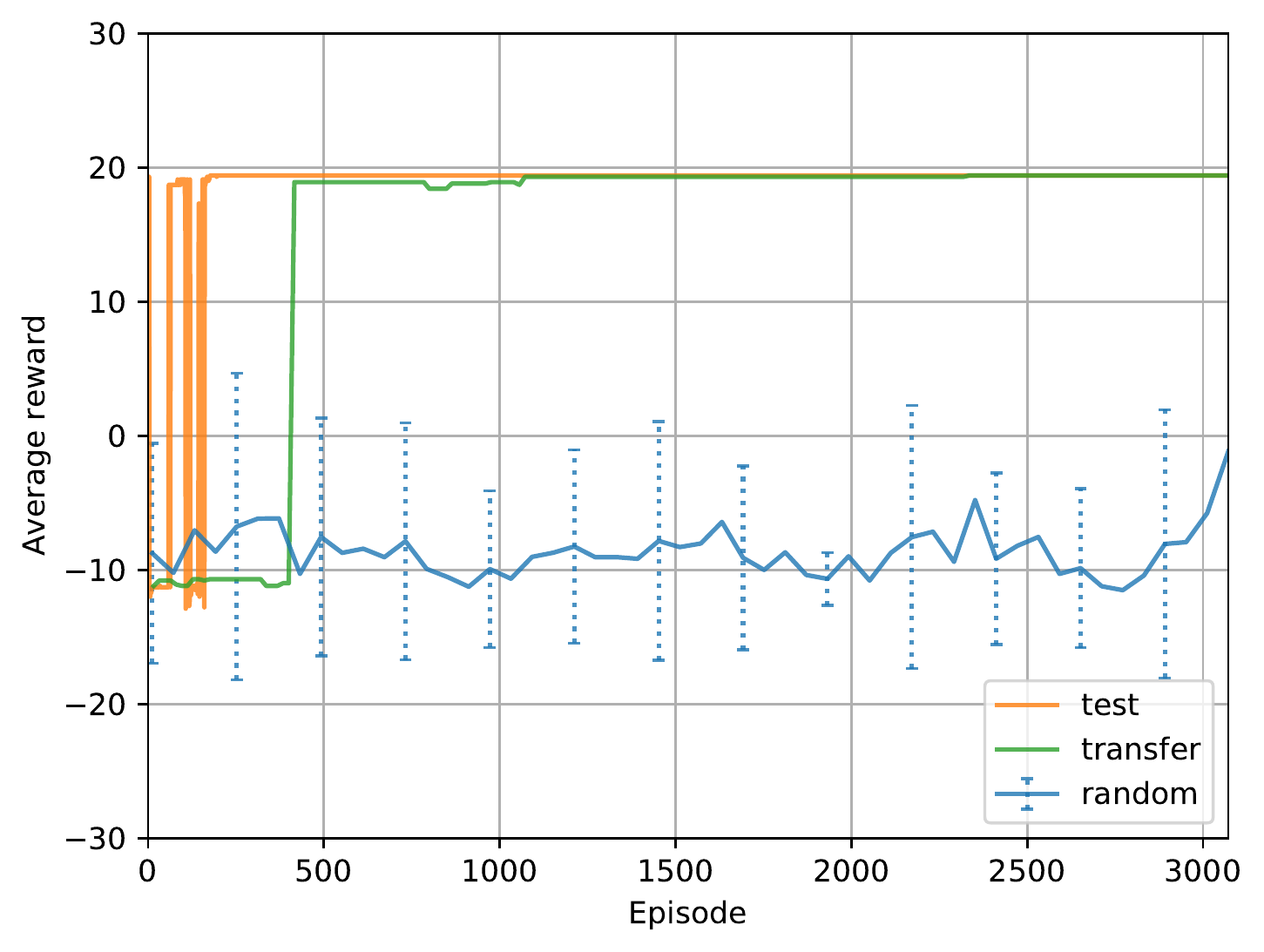}
\caption{Saving John}
\end{subfigure}
\begin{subfigure}{.5\textwidth}
\includegraphics[width=\linewidth]{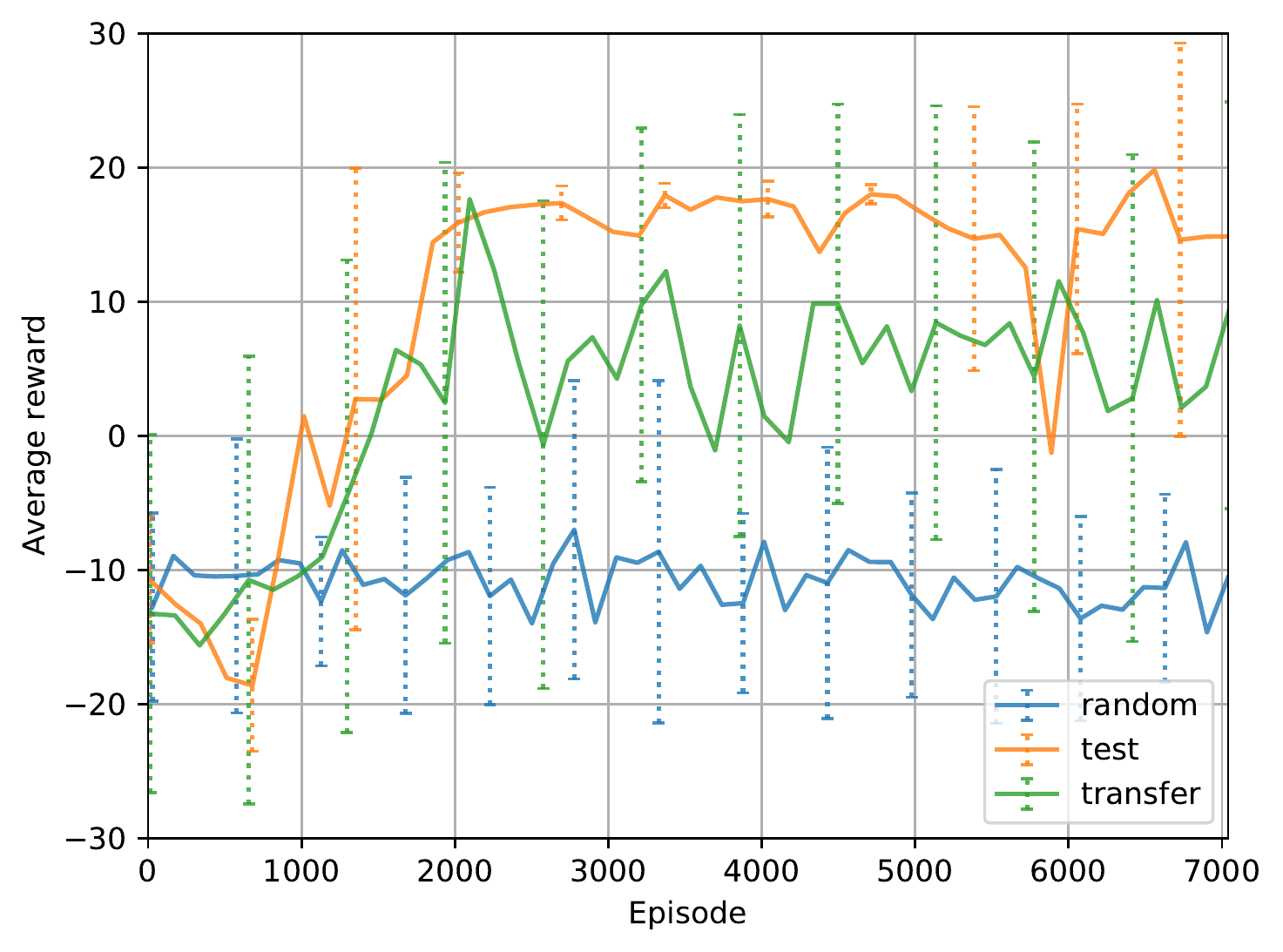}
\caption{Machine of Death}
\end{subfigure}
\begin{subfigure}{.5\textwidth}
\includegraphics[width=\linewidth]{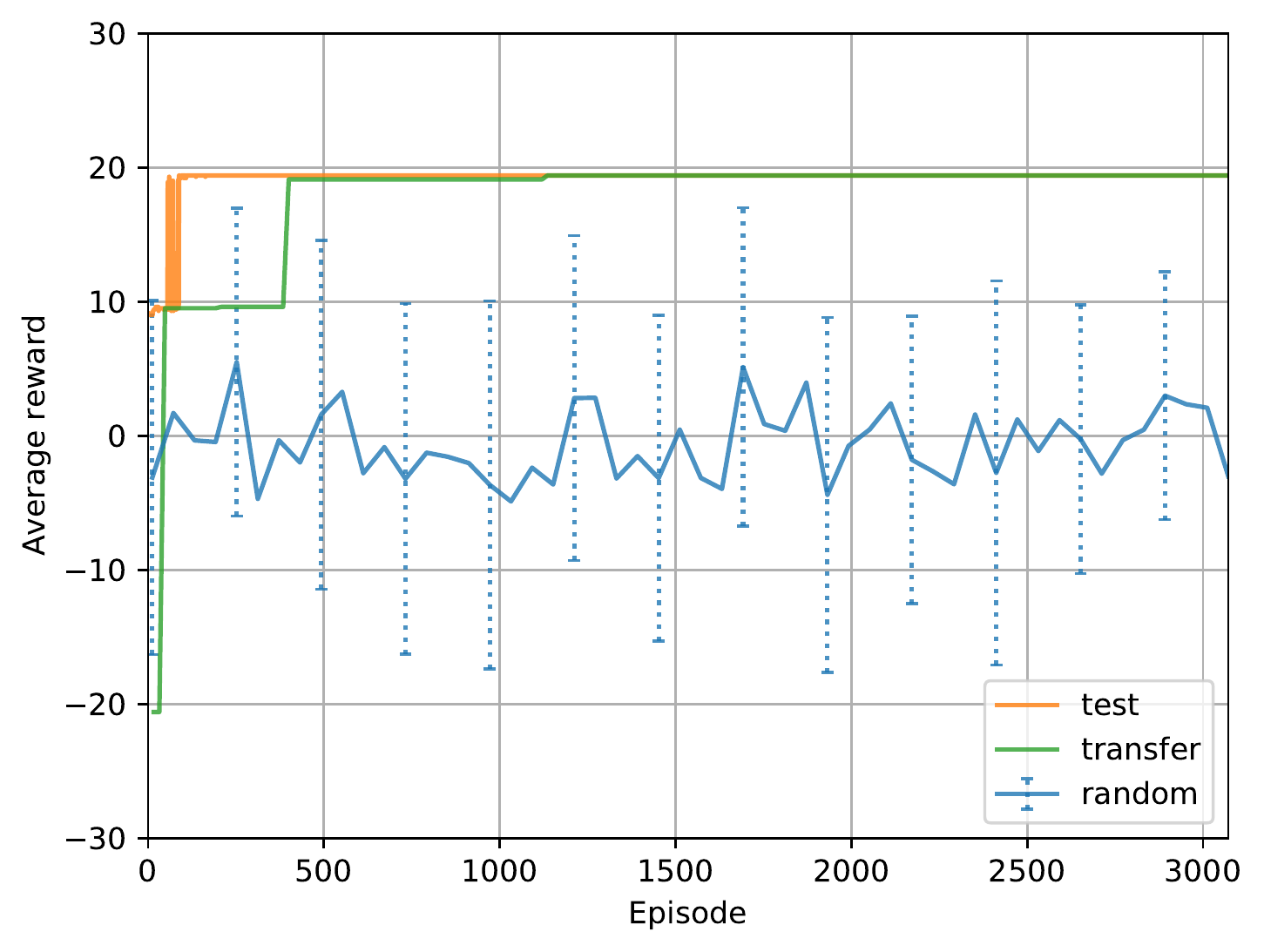}
\caption{Cat Simulator 2016}
\end{subfigure}
\begin{subfigure}{.5\textwidth}
\includegraphics[width=\linewidth]{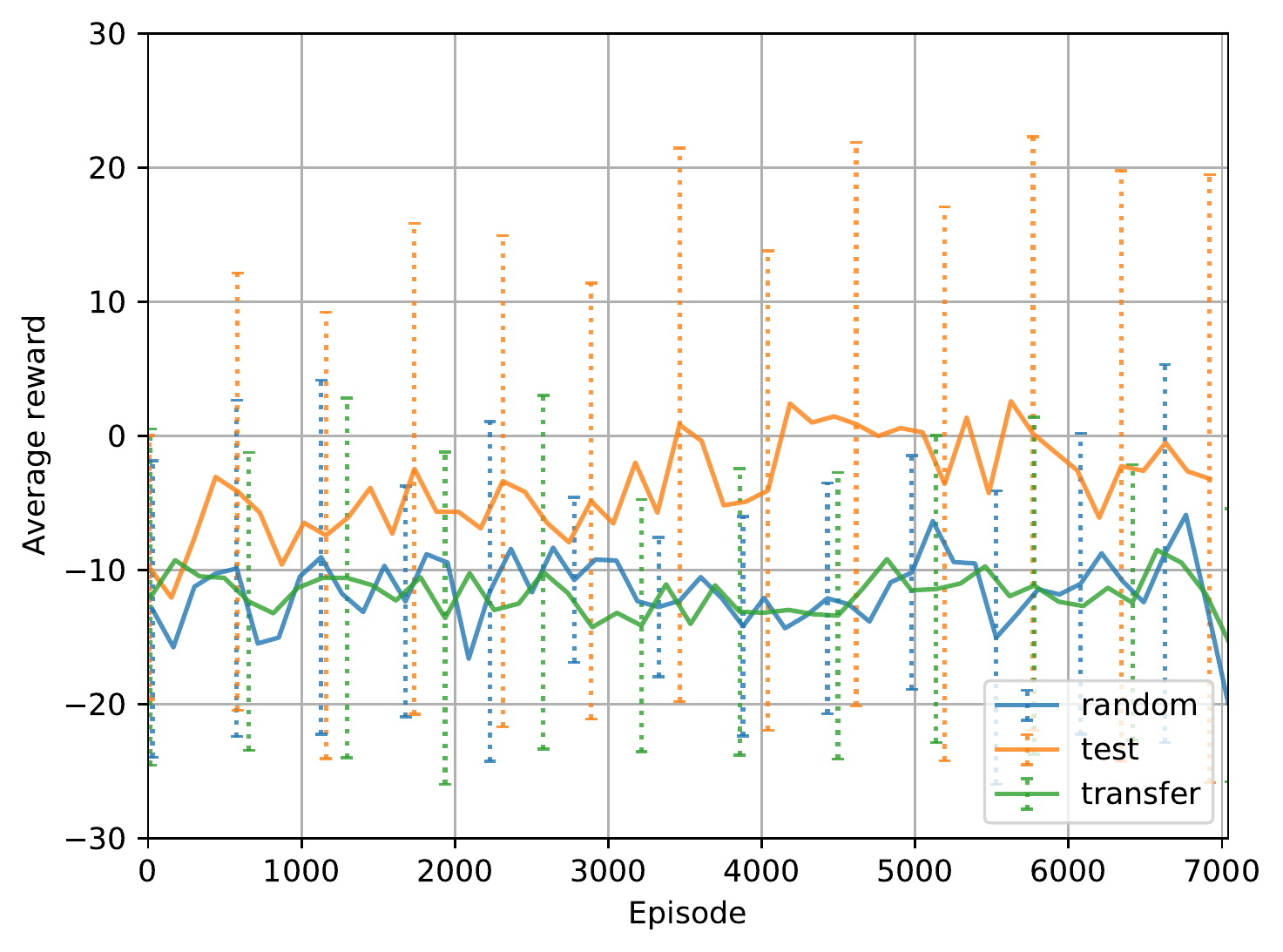}
\caption{Star Court}
\end{subfigure}
\begin{subfigure}{.5\textwidth}
\includegraphics[width=\linewidth]{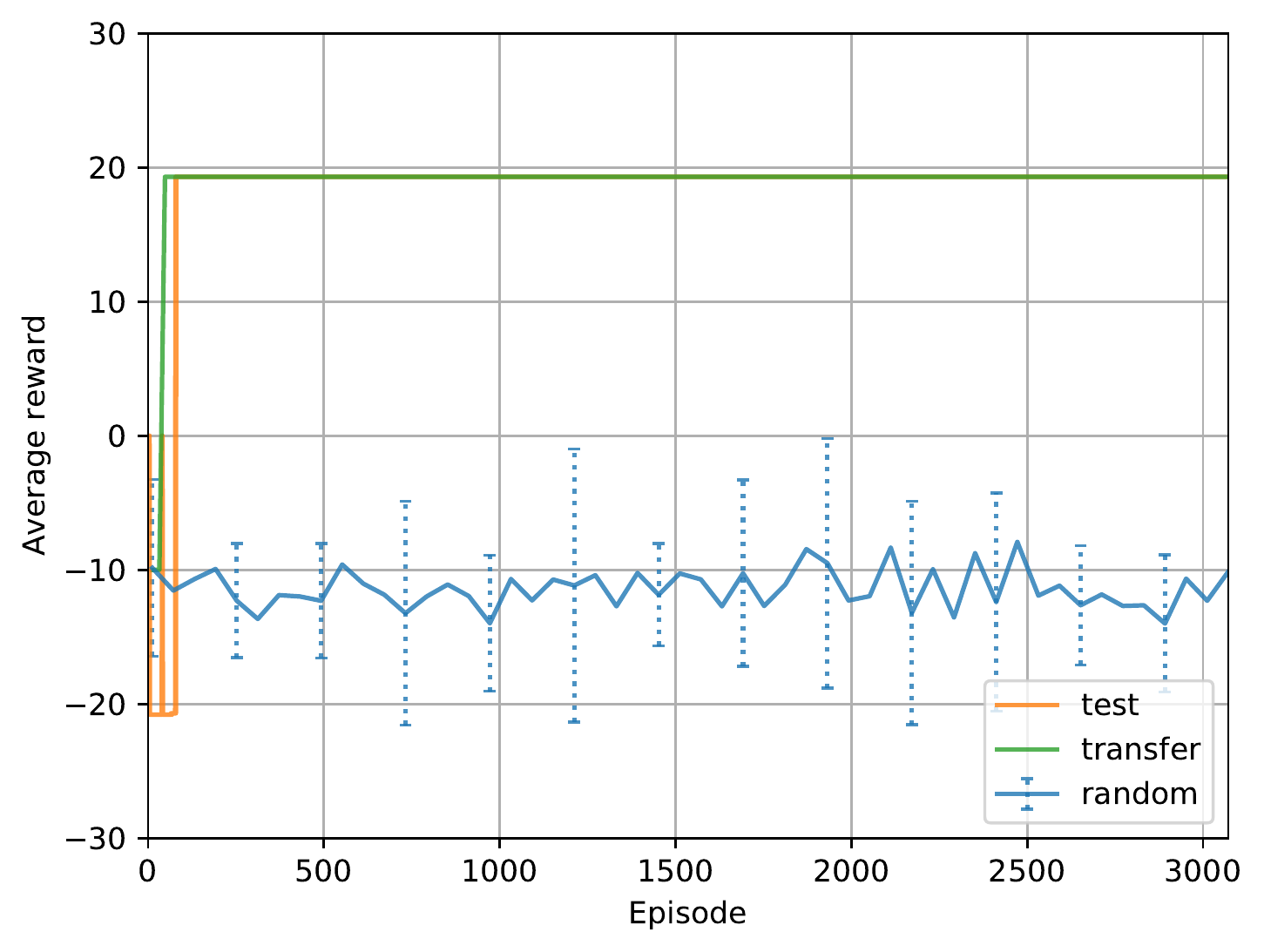}
\caption{The Red Hair}
\label{res:ttrh}
\end{subfigure}
\begin{subfigure}{.5\textwidth}
\includegraphics[width=\linewidth]{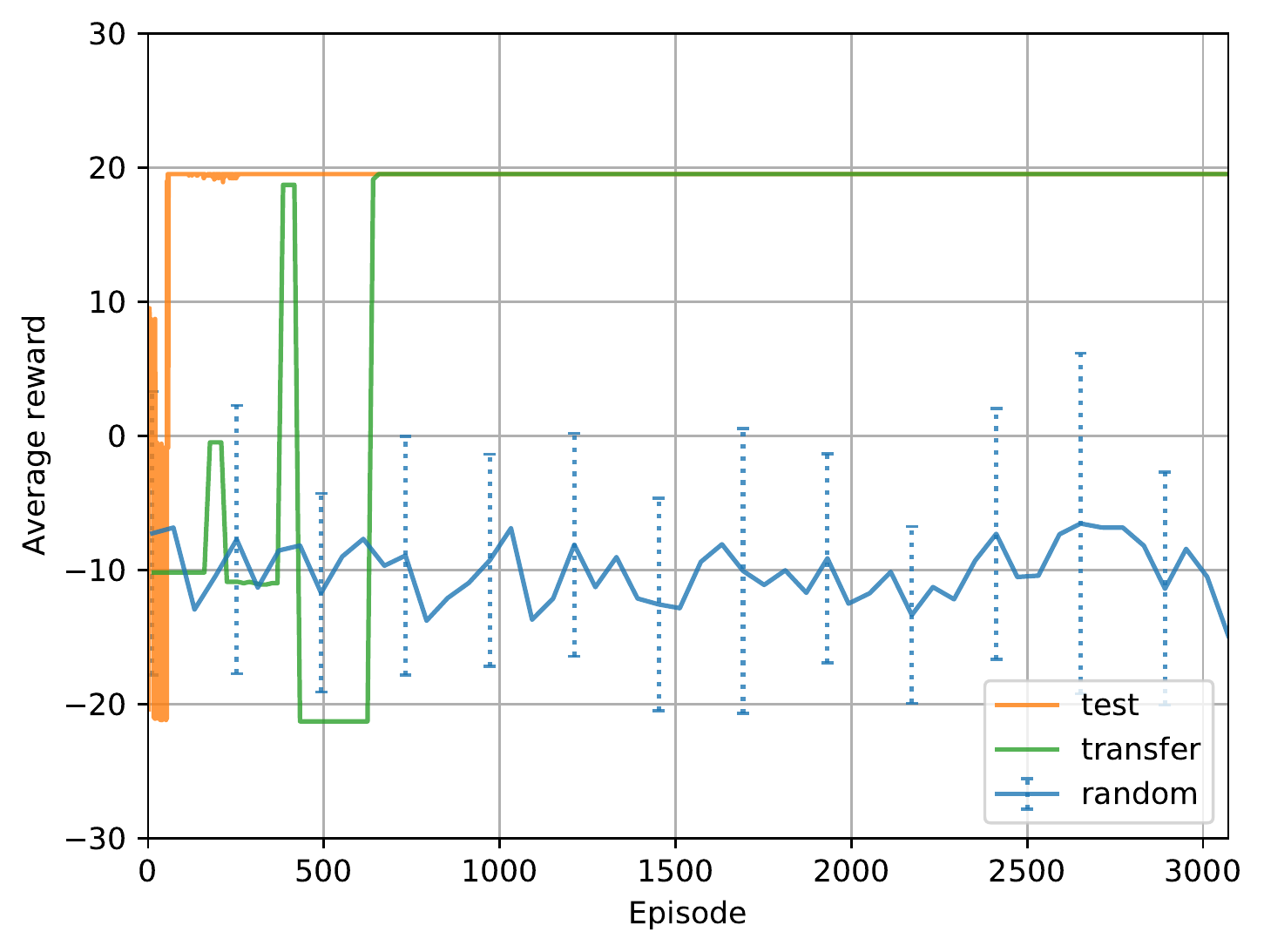}
\caption{Transit}
\end{subfigure}
\caption{Transfer learning performance.}
\label{res:transfer}
\end{figure}

For results, see figure \ref{res:transfer}.

The performance is, in most cases, slightly worse than in the task of playing individual games.







\subsection{Discussion}

For the most part, transfer learning on the simple games leads to a slower learning process with identical final results (the agent still converges to an optimal policy).
This is likely because instead of starting with small random weights, the model is now trained on five different games and, instead of making use of the pre-trained knowledge, the agent first has to ``forget'' and override its information about different games, making the task even more difficult.

The only exception is \emph{The Red Hair} (fig. \ref{res:ttrh}), in which an optimal policy is found immediately in transfer learning.
We hypothesise that this is the case thanks to its almost complete vocabulary coverage as seen in table \ref{tab:word-coverage}.
\\\\
In both \emph{Machine of Death} and \emph{Star Court}, the performance is significantly worse than in the individual task, both due to their sheer size (resulting in the pre-trained model overfitting on the simple games) and the specificity of their vocabularies.
\\\\
To summarise, the agent only displays little ability to transfer previously obtained knowledge to a similar but previously unseen dataset.

However, on a single game with a high percentage of shared tokens, transfer learning does yield better results than individual learning, although further testing on different datasets would be needed to confirm that this is the expected behaviour.

Lastly, while not necessarily benefiting the learning process, transfer learning at the very least does not render the agent unable to learn the simpler games at all.
Instead, the process is only slowed due to overfitting on previous data.

\setcounter{figure}{0}
\section{Playing multiple games at once}
\label{sec:universal}

Finally, we let the agent play all six games simultaneously, meaning both training and testing phases are executed on all games at the same time.
More precisely, a \emph{single instance} of the agent is used for this purpose
\\\\
This is a challenging task, since it requires the agent to integrate and combine the knowledge obtained from playing different text games into its limited model.

The first interesting question is whether the information from all six games can be condensed into such a small model, i.e. whether the agent will be able learn to play the games simultaneously at all.

Now if that would be the case, we would then be interested in the differences between learning curves of individual games when compared to the simple one game learning task.

Additionally, if the agent was not able to learn a game in an individual, per-game setting, will the universal task perhaps enhance its performance on the previously too difficult game?

\subsection{Setting}

Similarly to the generalisation experiment (section \ref{sec:generalisation}), we use a learning rate of~$0.00001$ and $\epsilon$-decay of~$0.999$.

The agent also uses the unified vocabulary of all games (see section \ref{sec:gen-setting}).
\\\\
When following the algorithm \ref{alg:dqn}, the agent samples each game once in each episode when following the exploration policy and storing experience data.

Consequently, the number of experience tuples of different games in the replay memory $\mathcal{D}$ should be proportionate to their respective average number of steps.

This is convenient, as the more complex games should then be sampled more often for learning.

\subsection{Results}

\begin{figure}[htb!]
\centering
\includegraphics[width=.5\textwidth]{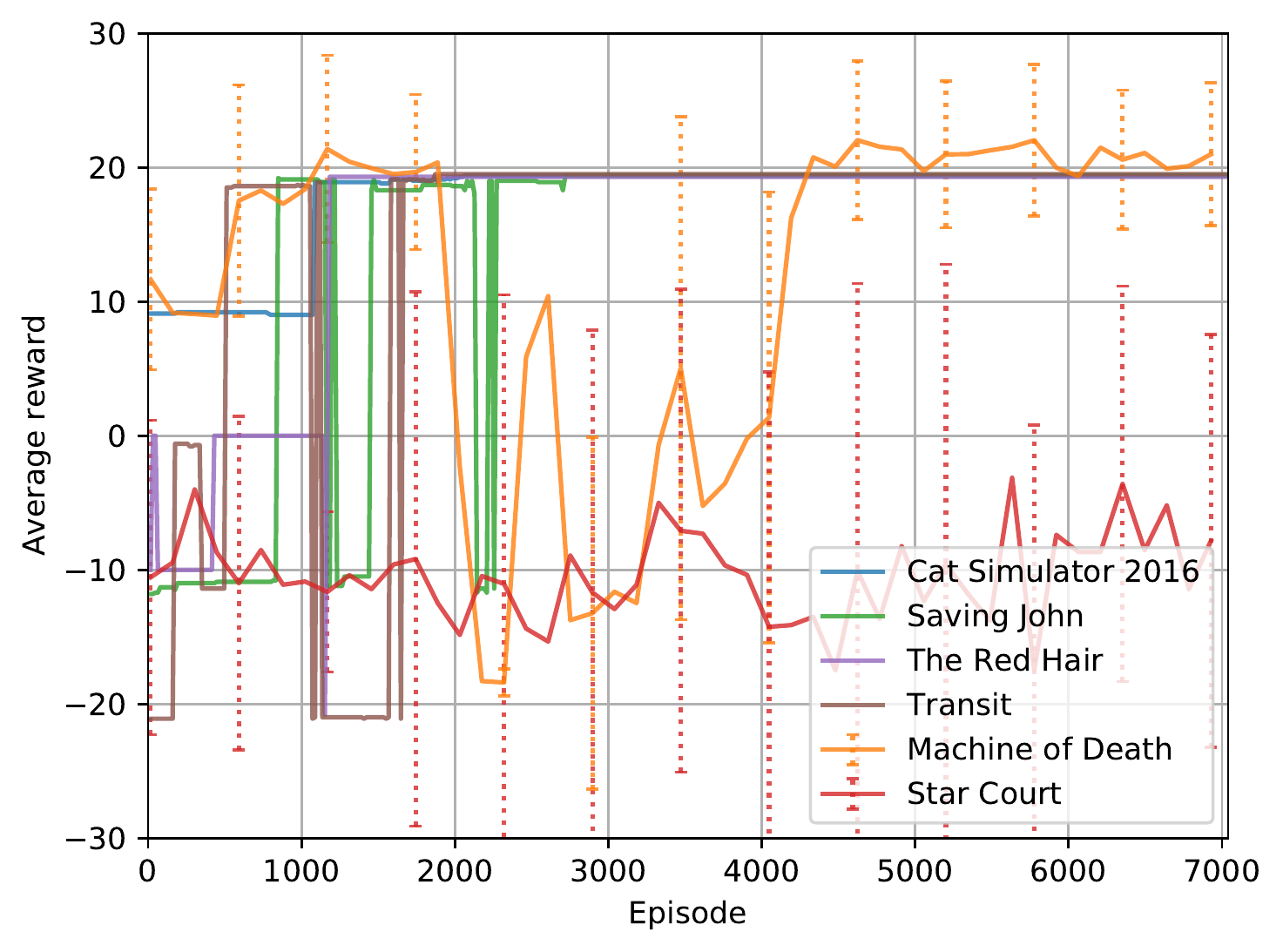}
\caption{Comparison of performance in all games in the multiple-games learning task.}
\label{res:universal-all}
\end{figure}

\begin{figure}[htb!]
\begin{subfigure}{.5\textwidth}
\includegraphics[width=\linewidth]{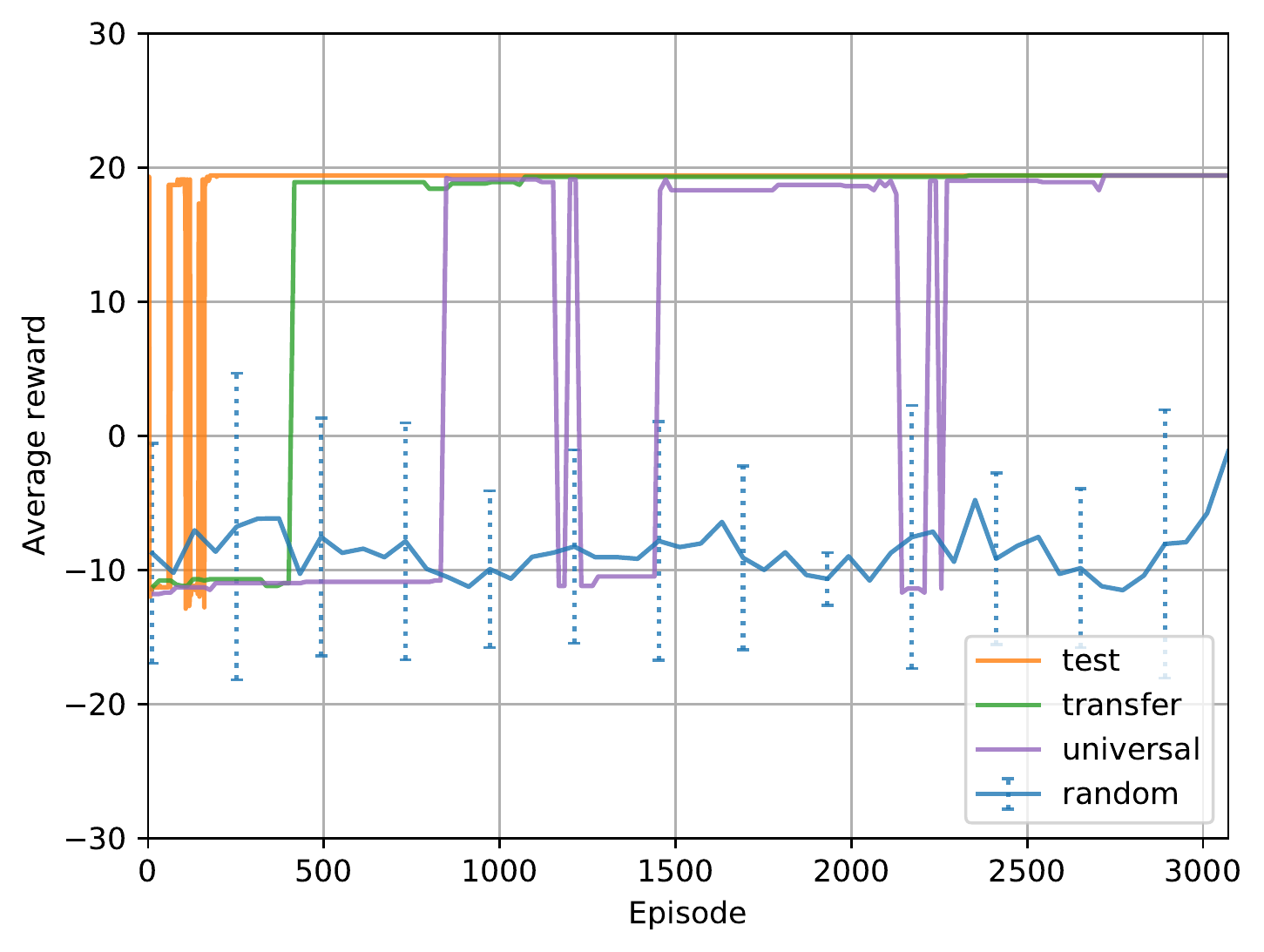}
\caption{Saving John}
\end{subfigure}
\begin{subfigure}{.5\textwidth}
\includegraphics[width=\linewidth]{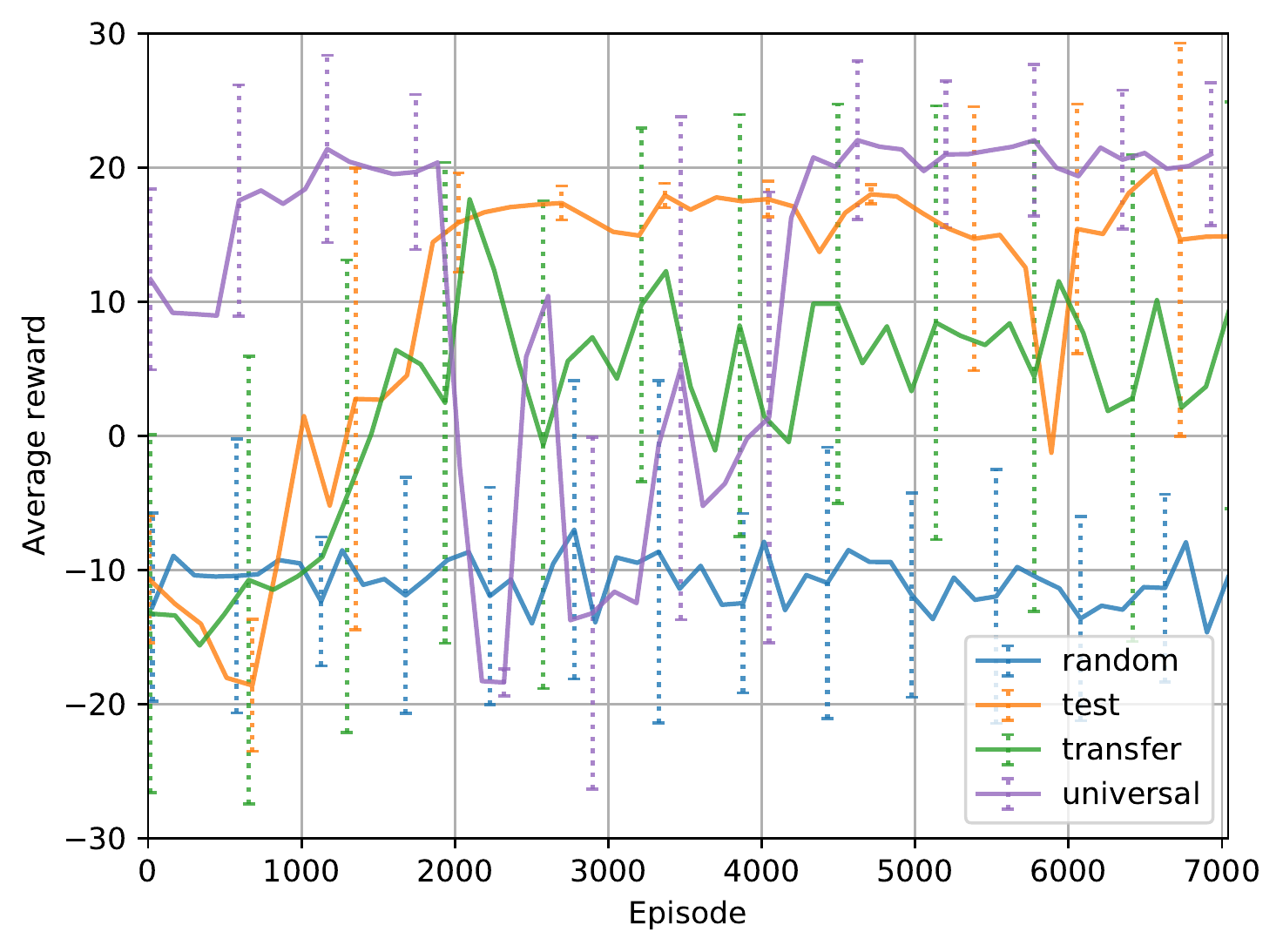}
\caption{Machine of Death}
\label{res:modu}
\end{subfigure}
\begin{subfigure}{.5\textwidth}
\includegraphics[width=\linewidth]{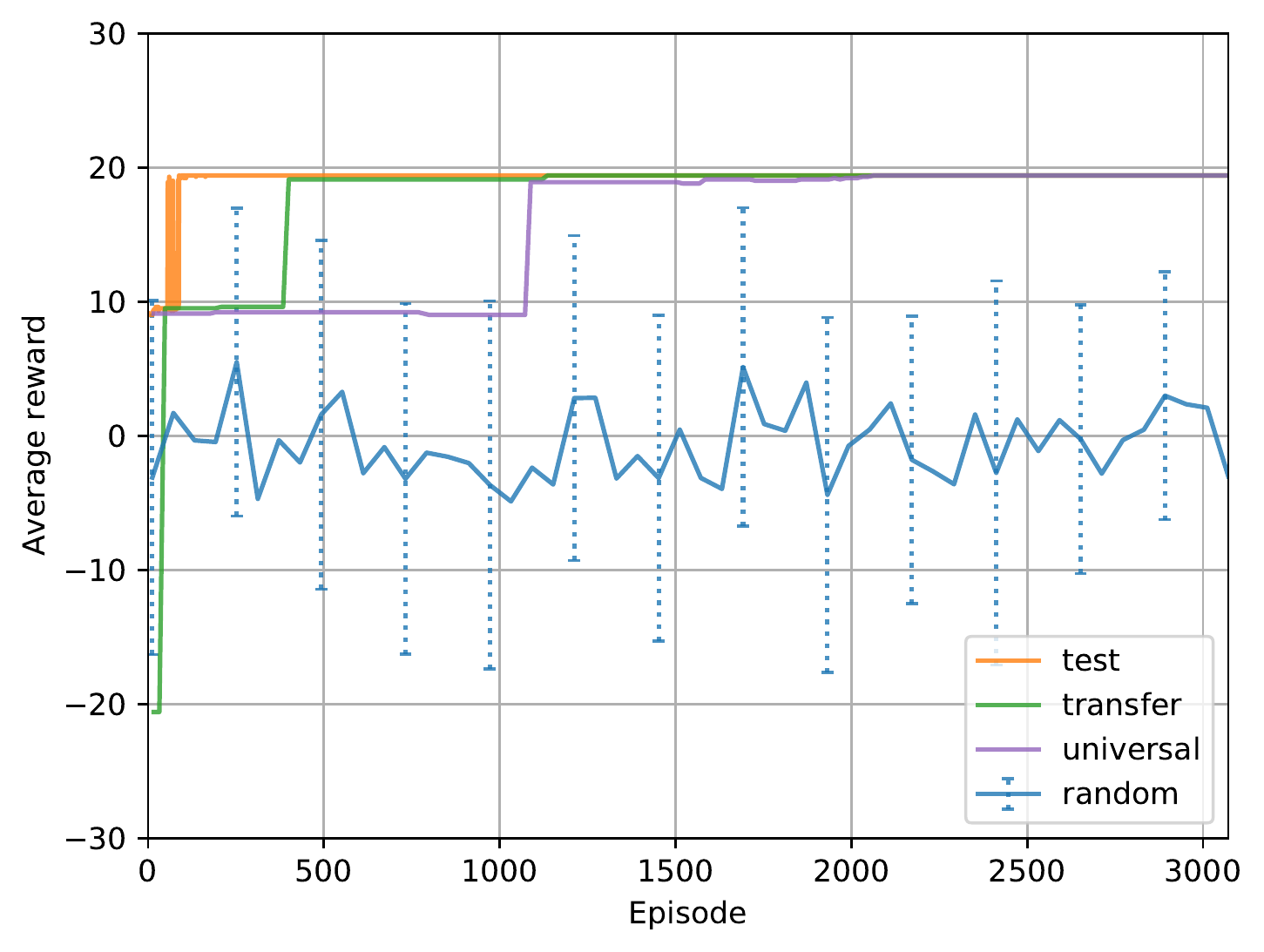}
\caption{Cat Simulator 2016}
\end{subfigure}
\begin{subfigure}{.5\textwidth}
\includegraphics[width=\linewidth]{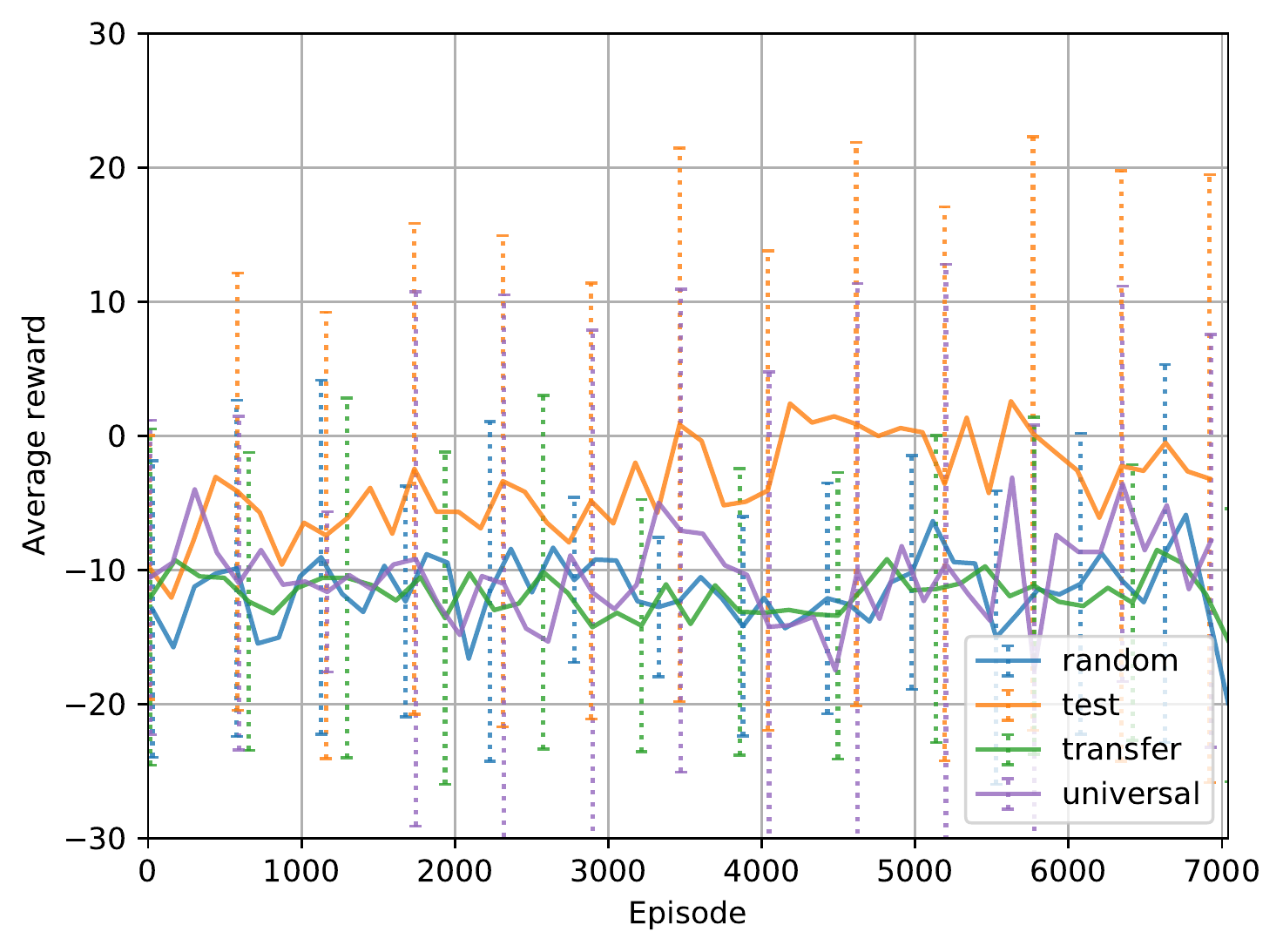}
\caption{Star Court}
\end{subfigure}
\begin{subfigure}{.5\textwidth}
\includegraphics[width=\linewidth]{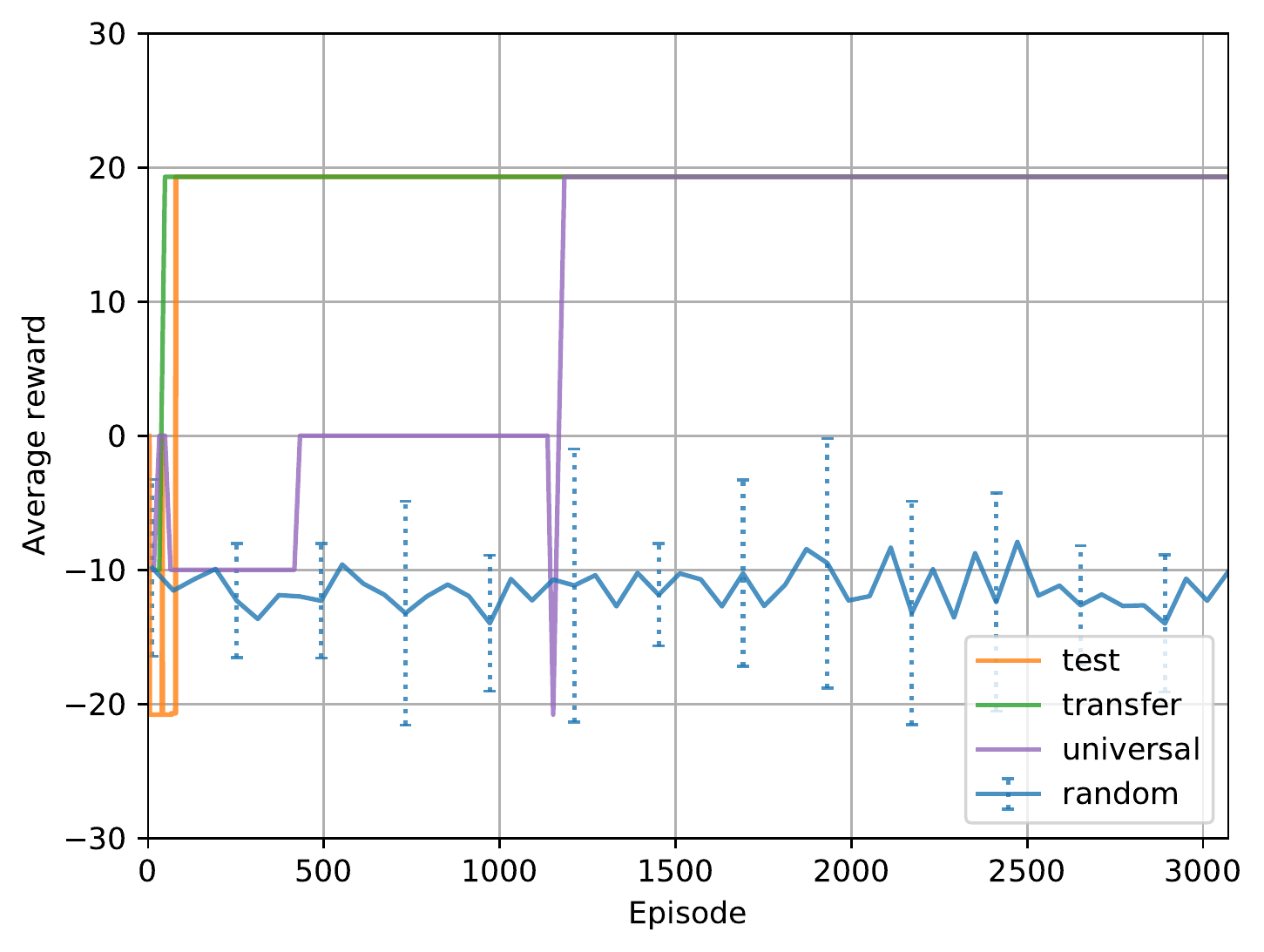}
\caption{The Red Hair}
\end{subfigure}
\begin{subfigure}{.5\textwidth}
\includegraphics[width=\linewidth]{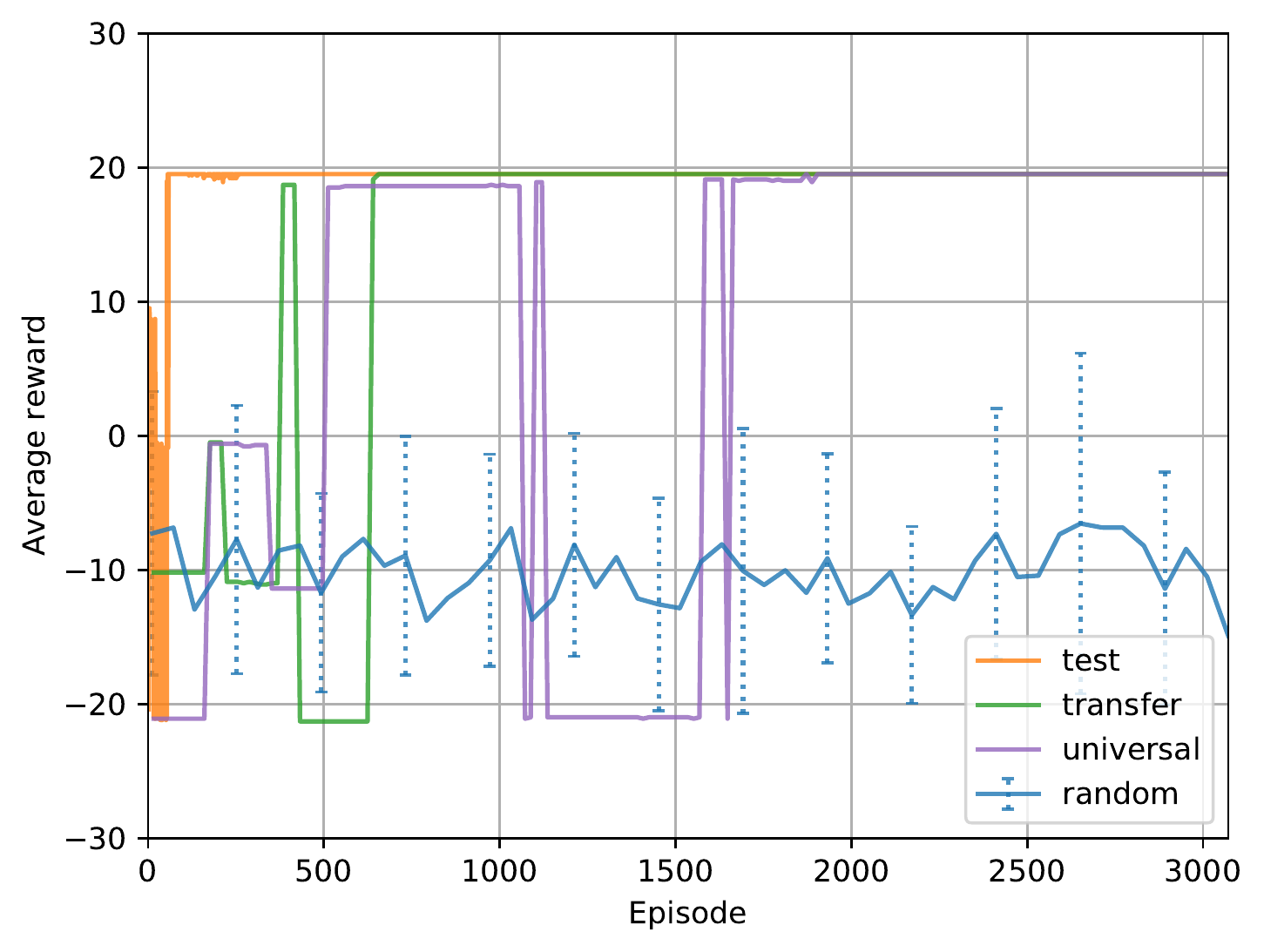}
\caption{Transit}
\end{subfigure}
\caption{Performance on the universal task --- playing multiple games at once.}
\label{res:universal}
\end{figure}

See figure \ref{res:universal-all} for a comparison of all games on this task only and refer to figure \ref{res:universal} for the usual format where we visualise multiple experiments for each game separately.

The \cd{universal} label corresponds to the multiple-games experiment.

\FloatBarrier

\FloatBarrier
\subsection{Discussion}

First of all, the agent was able to learn five games similarly to the individual task (see section \ref{sec:individual-games}).

The convergence rate on the simple games was considerably slower than in the individual and transfer learning settings, suggesting that at first, the agent struggled to compress the knowledge required to play multiple games.

Ultimately, the agent still found a single policy that is optimal in all four simple games, which is a very good result.
\\\\
For \emph{Star Court}, training on multiple games unfortunately did not result in better performance; the opposite was in fact the case.
This is to be expected, though, as not only was the individual \emph{Star Court} task difficult in itself, but we could also see in section \ref{sec:transfer} that learning from other games does not help if the majority of vocabulary is not shared between the game input data.
\\\\
The most interesting result, though, is that the agent was now able to converge on \emph{Machine of Death}, even to a nearly optimal reward (see figure \ref{res:modu}).

The optimal reward is equal to about $21.4$ (see table \ref{tab:game-stats} and section \ref{sec:optimal-rewards}) and the agent converged to a reward of~$21.0$.
For comparison, the best version of the DRRN agent converged to~$11.2$ in the individual-game task.

We hypothesise that the improvement in the multiple-game setting over the individual-game setting exists because the simpler games act as regularisers for the neural network.

It is possible that instead of overfitting on two of the three game branches of \emph{Machine of Death}, as was likely the case in section \ref{sec:individual-games}, the network is now forced to condense its information to the point that information from other games positively influence the agent's decisions in \emph{Machine of Death}.

Moreover, as suggested by the very slow convergence on the four simple games, it is likely that the higher rate of information compression results in better feature distillation and consequently, to a better knowledge transfer from the simpler games to \emph{Machine of Death}.
\\\\
At any rate, similarly to all our experiments, further testing with more data is needed to verify our hypotheses.

We can still safely conclude that the SSAQN agent is capable of playing multiple games at the same time whilst only using a minimal architecture with shared representation layers for states and actions.
More precisely, the universal agent is able to find a policy that is optimal in several different games.

Additionally, the agent performs better on a complex, non-deterministic game in the multiple-games setting than when only learning it individually, even converging to a nearly optimal average reward\footnote{The final average reward is better than that of human players in \emph{Machine of Death}, as \cite{DBLP:journals/corr/HeCHGLDO15} recorded a performance of $16.0$ for experienced players.}.

This suggests that the agent is capable of transfer learning, although it seems that a high level of knowledge compression is required for the agent to display this ability.

\chapter*{Future work}
\addcontentsline{toc}{chapter}{Future work}

We did not systematically search for optimal hyper-parameters for our agent model and it is likely that a better performance could be reached by their tuning.

However, achieving the best possible results on the presented environments is not the primary goal. Instead, the most interesting and relevant topic in the domain of IF games is the agent's ability to generalise and it is very likely that in order for the artificial IF agents to display good generalisation properties, a much larger volume of IF games would be needed.

We thus believe that the major challenge lies in expanding the set of available IF games by adding more environments, resulting in a bigger overlap between the natural language space of the individual IF games as well as between the domains of IF games and natural dialogues.

This would likely lead to both better generalisation and transfer learning capabilities of models similar to SSAQN (\cite{DBLP:journals/corr/BajgarKK16}).
\\\\
Another interesting direction of future research might include learning from IF game manuals or perhaps even from game transcripts of human players, similarly to \cite{DBLP:journals/corr/BranavanSB14} or \cite{DBLP:journals/corr/WilliamsAZ17}.

For a lot of popular IF games, transcripts are available on the \emph{ClubFloyd}\footnote{ClubFloyd information and history: \url{http://www.ifwiki.org/index.php/ClubFloyd}.} website, where players meet regularly to play IF games together. The game traces are archived and publicly available\footnote{Transcripts are published on \url{http://www.allthingsjacq.com/interactive_fiction.html\#clubfloyd}.} and could consequently be used as a~basic source of guidance data.
%
%

\chapter*{Conclusion}
\addcontentsline{toc}{chapter}{Conclusion}

In this thesis, we introduced the task of learning control policies for playing text-based games, in which state and action descriptions are given in natural language.

The text-game learning task is rather challenging and the ability to act optimally in complex text games could potentially have a positive impact on solving interesting real-world tasks such as comprehending and answering in natural dialogues.
\\\\
Similarly to the work presented in \cite{DBLP:journals/corr/NarasimhanKB15} and \cite{DBLP:journals/corr/HeCHGLDO15}, we employed deep reinforcement learning techniques based on the DQN algorithm (\cite{DBLP:journals/nature/MnihKSRVBGRFOPB15}) to build an agent theoretically capable of capturing high-level features of the game descriptions.

Our goal was to design a general architecture with shared representations for both states and actions, serving as a minimalistic proof of concept.

The resulting agent model only employs one recurrent and one dense layer in its underlying neural network that realises feature extraction from text descriptions and it performs similarly to or better than the DRRN model from \cite{DBLP:journals/corr/HeCHGLDO15} on two games presented therein.
\\\\
Importantly, we note that the main challenge in similar language-based sequential decision-making tasks does not lie in maximising performance on a single game, but rather in being able to generalise to previously unseen data.

For this purpose, we present pyfiction, an open-source library enabling universal access to different text games.

The library includes several different games on which we evaluated our agent mainly in terms of its generalisation and transfer learning capabilities.
\\\\
We show that it is very difficult to generalise in text games, especially on a relatively small number of games.
Even more importantly, it is difficult to correctly assess generalisation properties of a model in this context as text games are very sensitive to subtle changes of the agent's policy.

Finally, we test our agent on the universal task of playing multiple text games at once.
We observe that a single instance of the agent is capable of simultaneously playing multiple games optimally and that this task leads to an even better performance on a non-deterministic game when compared to learning it individually, suggesting the agent's ability to transfer knowledge.
\\\\
Additionally, the source code of our agent model is publicly available as a part of pyfiction and we hope that both the model and the text-game interface that pyfiction introduces could serve as a baseline for future research.

\renewcommand{\thefigure}{\oldfig} 


\bibliographystyle{plainnat}    

\renewcommand{\bibname}{Bibliography}

\bibliography{bibliography}

\listoffigures

\listoftables

\chapwithtoc{List of Abbreviations}

\begin{table}[ht]
\label{tab:abbreviatons}
\begin{tabular}{ll}
\toprule
IF & interactive fiction \\
IFDB & Interactive Fiction Database, \url{http://ifdb.tads.org/} \\
CYOA & choose your own adventure \\
\midrule
MDP & Markov decision process \\
AI & artificial intelligence \\
RL & reinforcement learning \\
(A)NN & (artificial) neural network \\
MLP & multilayer perceptron \\
MSE & mean squared error \\
SGD & stochastic gradient descent \\
RNN & recurrent neural network \\
LSTM & long short-term memory (a form of a RNN, \cite{DBLP:journals/neco/HochreiterS97}) \\
BPTT & backpropagation through time \\
NLP & natural language processing \\
BOW & bag of words \\
DQN & Deep Q-Network (\cite{DBLP:journals/nature/MnihKSRVBGRFOPB15}) \\
DRRN & Deep Reinforcement Relevance Network (\cite{DBLP:journals/corr/HeCHGLDO15})\\
SSAQN & Siamese State-Action Q-Network (chapter \ref{chp:agent})\\
\midrule
SJ & Saving John \\
MoD & Machine of Death \\
CS & Cat Simulator 2016\\
SC & Star Court\\
TRH & The Red Hair\\
TT & Transit\\
\bottomrule

\end{tabular}
\end{table}


\begin{appendices}

\chapter{Text games}
\label{app:text-games}

In this appendix, we

(i) present a table on page~\pageref{tab:summary} that summarises the majority of relevant properties of the different games as well as the results of the SSAQN agent in these environments;

(ii) provide complete information about the rewards assigned to different IF games we used for evaluating our agent model.
\\\\
To recapitulate, we utilised the following IF games: \emph{Saving John}\footnote{SJ: \url{http://ifdb.tads.org/viewgame?id=1vzv5y27vixinm22}}, \emph{Machine of Death}\footnote{MoD: \url{http://ifdb.tads.org/viewgame?id=u212jed2a7ljg6hl}}, \emph{Cat Simulator 2016}\footnote{CS: \url{http://ifdb.tads.org/viewgame?id=79f1ic623cvxtpio}}, \emph{Star Court}\footnote{SC: \url{http://ifdb.tads.org/viewgame?id=u1v4q16f7gujdb2g}}, \emph{The Red Hair}\footnote{TRH: \scriptsize{\url{http://textadventures.co.uk/games/view/r0fika63aksao_qkq8n3lq/the-red-hair}}} and  \emph{Transit}\footnote{TT: \url{http://ifdb.tads.org/viewgame?id=stkrrbel8m21b37q}}.

For numerical rewards associated with the different game endings of \emph{Saving John} and \emph{Machine of Death}, see \cite{DBLP:journals/corr/HeCHGLDO15} whose values we use and whose text game simulator is wrapped by pyfiction (see appendix \ref{app:pyfiction}), resulting in an unified interface for all employed games.

\subsection*{Rewards assigned to different endings}

For determining the ending type, we usually only detect and present substrings of the final states.

The rules are evaluated in the order given by the table.

In pyfiction's source code, the reasoning behind the reward value is usually given for the corresponding reward state in each game simulator.

\subsubsection*{Cat Simulator 2016}

The rewards are based on whether the cat was able to get food and whether it found a good place to sleep.

\begin{table}[htb!]
\centering
\footnotesize
\begin{tabular}{lr}
\toprule
\textbf{Start of the ending text}& \textbf{Reward} \\
\midrule
this was a good idea & 0 \\
as good a place as any & -20 \\
mine! & 10 \\
catlike reflexes & -20 \\
finish this & -20 \\
friendship & 20 \\
not this time, water & 10 \\
serendipity & 10 \\
\bottomrule
\end{tabular}
\caption{Cat Simulator 2016 rewards.}
\label{rew:cs}
\end{table}
\FloatBarrier

\newpage
\subsubsection*{Star Court}

The rewards are based on the fact if the player survived; if they did, but got sentenced, the reward scales with the imprisonment time in years. Positive rewards are obtained by either escaping the prison or by ``living happily ever after'', with bonuses to using favours to improve quality of life.

The $X$ variable refers to the length of imprisonment, $X \leq 2000$.

\begin{table}[htb!]
\centering
\footnotesize
\begin{tabular}{lr}
\toprule
\textbf{Ending substring}& \textbf{Reward} \\
\midrule
You get a job as a & 5 \\
Here on the astral plane, your psychic bodies are as physical & -20 \\
Nah. You die as poison consumes your body. And because you failed trial by poison & -30 \\
You're all out of favors! I guess working as & 15 \\
The only thing Pride finds more beautiful than itself is the destruction  & -30 \\
Immediately upon starting the battle, the titanic creature falls asleep & -30 \\
You are torn limb from limb by the many-limbed creature! & -30 \\
You remember you training at Psi City and concentrate & -30 \\
And so you do, spacer, so you do. & 15 \\
 BLAMMO!! You're dead! And what's worse, you're guilty!& -30\\
 The Judge bangs their laser gavel a final time. "Robailiff, you may take the prisoner& $-X / 100$ \\
 You're dead! I guess that means you're guilty! & -30\\
 You are neither guilty nor innocent, as law has been dethroned in the universe.& -20\\
You let Star Court evaporate like a bad memory. You're on the other side  & 10\\
How does Star Court generate this much trash, you think as you burn. & -20\\
You got smoked by a crime ghost. & -20\\
Congratulations, you're innocent! You're also dead. & -20\\
The knife hits you right between the eyes. You are killed immediately, & -30\\
means you're guilty! & -30\\

\bottomrule
\end{tabular}
\caption{Star Court rewards.}
\label{rew:sc}
\end{table}
\FloatBarrier
\newpage
\subsubsection*{The Red Hair}

The rewards are based on whether the player accepts the job and whether they are able to protect the children.

\begin{table}[htb!]
\centering
\footnotesize
\begin{tabular}{lr}
\toprule
 \textbf{Ending substring}& \textbf{Reward} \\
\midrule
you lose&-10 \\
all there is left is a red hair& -20\\
it was the clown statue missing&-20 \\
you stay in the bedroom and eventually the parents come back and thank you&20 \\
\bottomrule
\end{tabular}
\caption{The Red Hair rewards.}
\label{rew:trh}
\end{table}

\subsubsection*{Transit}

Positive rewards are for tackling the right man; negative rewards are assigned to the endings in which the player is arrested or dies.

\begin{table}[htb!]
\centering
\footnotesize
\begin{tabular}{lr}
\toprule
 \textbf{Ending substring}& \textbf{Reward} \\
\midrule
if anyone can help you&10 \\
you buy one more can& -20\\
even though it was just in-passing& 20\\
you make swift use of& -20\\
the guards know&-10 \\
as you predicted&-10 \\
you close your eyes and submit to death& -20 \\
you're in a country& -10\\
through the haze of the drinks& 10\\
while the last parts of your mind untouched& -10 \\
\bottomrule
\end{tabular}
\caption{Transit rewards.}
\label{rew:tt}
\end{table}

\newpage

\begin{landscape}
\begin{table}[ht]

\label{tab:summary}
\small
\centering
\begin{tabular}{llrrrrrr}

\toprule
& & \textbf{Saving John} & \textbf{Machine of Death} & \textbf{Cat Simulator 2016} & \textbf{Star Court} & \textbf{The Red Hair} & \textbf{Transit} \\

\midrule
\multirow{6}{*}{\rotatebox[origin=c]{90}{\makecell{\textsc{Properties}}}}
                                                               
&\# tokens                             &  1119  &   2055  &   364 &  3929  &  155   &  575  \\
&\# states                             & 70   &  $\geq$ 200   &  37  & $\geq$ 420  &  18   &  76  \\
&\# endings                            &  5  &  $\geq$ 14  &  8  & $\geq$ 17   &    7 &  10  \\
&Avg. words/description                &  73.9  & 71.9    & 74.4   &  66.7  &  28.7   & 87.0   \\
&Deterministic transitions             &  Yes  & No    & Yes   & No   &   Yes  & Yes    \\
&Deterministic descriptions            &   Yes & Yes    &  Yes  &  No  &  Yes   & Yes  \\

\midrule
\multirow{6}{*}{\rotatebox[origin=c]{90}{\makecell{\textsc{Final reward}}}}

& (\ref{sec:individual-games}) \ \ Random agent (average)            &      -8.6       &  -10.8  &      -0.6    &     -11.6  &   -11.4  &     -10.1    \\
& (\ref{sec:individual-games}) \ \ Individual game         &        19.4      & 15.4 &    19.4     &    -2.2    &     19.3   &    19.5     \\
& (\ref{sec:generalisation}) \ \ Generalisation          &         -11.2     & -15.1 &     5.7       &   -13.2 &        -10.0      &    -10.2     \\
& (\ref{sec:transfer}) \ \ Transfer learning       &           19.4   &   8.7 &        19.4            &     -13.3       &     19.3 &      19.5   \\
& (\ref{sec:universal}) \ \ Multiple games           &       19.4      &     21.0       &     19.4       &     -8.2       &      19.3 &     19.5    \\
& (\ref{sec:optimal-rewards}) Optimal                  &    19.4     &  $\approx$ 21.4  &        19.4       &     ?       &      19.3    &    19.5 \\

\bottomrule
\end{tabular}
\caption[Game properties and agent performance summary.]{Summary of game statistics and performance comparison of the SSAQN agent on different tasks.}
\label{tab:games-summary}

\end{table}
\end{landscape}

\chapter{pyfiction}
\label{app:pyfiction}

\textbf{pyfiction} is an open-source Python library for simple access to different IF games developed originally for its use in this thesis.
\\\\
In addition to the text games supported by pyfiction and used in this thesis (see appendix \ref{app:text-games}), the SSAQN agent was also implemented as a part of the library.

Consequently, all experiments conducted in chapter \ref{chp:experiments} using the SSAQN agent are available as runnable examples in pyfiction.
\\\\
The complete documentation on how to install pyfiction and run the examples is available on the library's website:

\url{https://github.com/MikulasZelinka/pyfiction}.

For future reference to the code base at the time of writing this thesis, we also released version 0.1.0 of pyfiction, always available under its tag:

\url{https://github.com/MikulasZelinka/pyfiction/releases/tag/v0.1.0}.

\subsection*{SSAQN as implemented in pyfiction}

Our implementation of the SSAQN agent is realised in Keras (\cite{chollet2015keras}) using the Tensorflow backend (\cite{tensorflow2015-whitepaper}).

\begin{figure}[htb!]
\includegraphics[width=\textwidth]{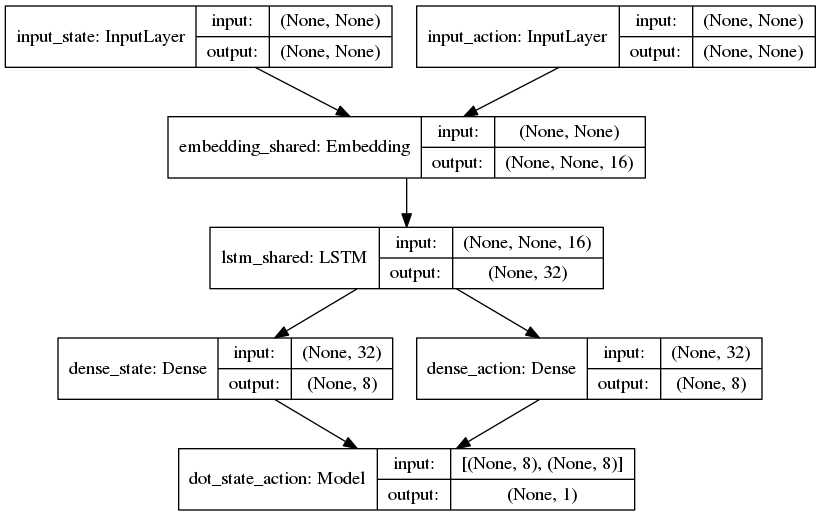}
\caption[Keras implementation of the SSAQN model.]{
Keras implementation of the SSAQN model (see figure \ref{fig:architecture}). as visualised by \cd{plot\_model} in Keras. The first \cd{None} values correspond to the variable batch size. The second \cd{None} values, if present, correspond to the length of the state and action descriptions.}
\label{fig:architecture-keras}
\end{figure}

For purposes of future reference, we provide a visualisation of our implemented model in figure \ref{fig:architecture-keras} as plotted by Keras.

\end{appendices}

\openright

\end{document}